# Multiclass Linear Perceptrons with Multiplicative Margins

Dmitri Rachkovskij[1,2], Evgeny Osipov[1], Olexander Volkov[2], Daswin De Silva[3], Denis Kleyko[4,5]
[1]Department of Computer Science, Electrical and Space Engineering, Luleå University of Technology, 971 87 Luleå, Sweden
[2]Institute of Information Technologies and Systems, 03187 Kyiv, Ukraine
[3]Centre for Data Analytics and Cognition, La Trobe University, Melbourne, Australia
[4]AI, Robotics and Cybersecurity Center and Department of Computer Science, Örebro University, Örebro, 70182, Sweden
[5]Intelligent Systems Lab, Research Institutes of Sweden, Kista, 16440, Sweden

## Abstract

This paper introduces a family of multiclass linear Perceptron classifiers with a multiplicative margin mechanism (MMPerc), as an alternative to standard margin-free and additive margin Perceptrons. The multiplicative formulation enforces classification confidence by requiring the true class score to exceed that of competing classes by a specified fraction of itself, rather than by a fixed additive threshold. This avoids dependence on score magnitudes arising from varied norms of data and class weight vectors.
We propose several architectural and algorithmic variants of MMPerc, derive associated loss functions and mistake bounds for both linearly separable and non-separable data, and analyze key design considerations, including bias, margin threshold selection, and training modes. Extensive experiments on synthetic and real datasets show that MMPerc classifiers typically outperform the standard Perceptron, as well as classic baselines such as Support Vector Machines and Ridge classifiers. Owing to their simplicity, minimalistic design, and computational efficiency, MMPerc classifiers are promising candidates for conventional machine learning tasks, linear evaluation of Deep Neural Networks, integration with Hyperdimensional Computing / Vector Symbolic Architecture representations, and deployment in resource-constrained applications.

# 1 Introduction

**Linear models**. Despite the rise of Deep Neural Networks (DNNs), models linear in their parameters (Searle & Gruber, 2016), (Matloff, 2017), (Christensen, 2019) remain fundamental and widely used in both research and practice. Their advantages include simplicity and interpretability, computational efficiency in training and evaluation, compatibility with convex optimization and regularization, and broad applicability across domains. Two fundamental and closely related categories in supervised machine learning are linear classification and linear regression, which support tasks such as classification, prediction, detection, identification, assessment, analytics, etc.

In many real-world problems, the input-output relationship is linear not only in the model parameters but also in the input data features. When the latter does not hold, one can apply nonlinear transformations to the input space, mapping the original features into a secondary feature space (often referred to as a Hilbert space), where the relationship is linear in the transformed features. This extends the applicability of linear models to problems that are nonlinear in the original feature space.

**The Perceptron**, introduced by Frank Rosenblatt in the 1950s (Rosenblatt, 1957, 1962), was the first trainable neuro-inspired classifier. It established core principles later adopted in machine learning and neural networks, with practical implementations extending to hardware (Hay et al., 1960). Rosenblatt's models also included multilayer variants, such as the alpha Perceptron, which combined fixed randomized nonlinear transformations in the early layers with a trainable linear output layer; these ideas have been revisited and extended in later work.

The corresponding two-stage architecture with a fixed randomized nonlinear feature transformation followed by a linear classifier has become a standard design in machine learning. It supports efficient training and inference, requires only modest computational requirements, and offers flexibility across diverse data types that can be represented as vectors. The linear output layer can be trained using simple iterative procedures (e.g., Perceptron/gradient-based methods) or in closed form (e.g., regularized least squares).

**Linear evaluation / probing in DNNs.** An important evolution of Rosenblatt's multilayer Perceptrons is the modern multilayer Perceptron (MLP), in which all layers are trainable; stacked MLPs constitute a canonical DNN architecture. Although DNNs typically require much larger datasets and longer training than single-layer linear Perceptrons, their learned nonlinear transformations can capture complex patterns in the data and address task-specific challenges. A common strategy for downstream use of DNN representations is transfer learning via linear evaluation (or linear probing): the pre-trained model is frozen and a linear classifier is trained on its learned features (He et al., 2020; T. Chen et al., 2020; Radford et al., 2021). Empirically, this indicates that the learned representations often make classes approximately linearly separable, a phenomenon supported by recent theory (Saunshi et al., 2019), (HaoChen et al., 2021). In practice, linear evaluation is an efficient, scalable alternative to full fine-tuning, especially as model size and complexity grow – highlighting the continued relevance of single-layer linear Perceptrons.

**HDC/VSA.** Hyperdimensional Computing (HDC) (Kanerva, 2009), also known as Vector Symbolic Architectures (VSA) (Gayler, 2004), provides a versatile paradigm that integrates neuro-inspired and symbolic AI, machine learning techniques, and computing frameworks for emerging nanoscale hardware, extending down to ultra-low-power edge devices (Neubert et al., 2019), (Thomas et al., 2021), (Aygun et al., 2023), (Clarkson et al., 2023), (Thomas et al., 2023), (Raviv, 2024), (Reimann, 2025).

A central task in HDC is to transform instances of fundamental data types (scalars, vectors, sequences, and graphs) into *hypervectors* such that their pair-wise similarities reflect and transform the original data relationships in ways relevant to downstream tasks. These transformations are typically randomized and nonlinear. The required dimensionality and format of hypervectors depend on the application. In scenarios involving complex knowledge structures, such as knowledge graphs, hypervectors may require thousands or more dimensions to adequately encode information, whereas TinyML applications benefit from low-dimensional, efficient formats, such as binary representations, for energy efficiency and low latency.

In classification tasks (Kussul et al., 1993, 1994), (Ge & Parhi, 2020), (Vergés et al., 2025), it is essential to deploy efficient classifiers that can operate across varying hypervector formats and dimensionalities. These classifiers should also support online learning, making linear Perceptron-based models particularly well suited for this role.

**Our study.** In linear classification, enforcing a margin between training instances and decision boundaries is known to improve generalization to unseen data. Support Vector Machines (SVMs) are the classic example (Vapnik & Chervonenkis, 1964), (Vapnik & Chervonenkis, 1974), (Cervantes et al., 2020). Large-margin extensions of the Perceptron have also been studied extensively, from early work (Mays, 1964), (Nilsson, 1965b), (Duda & Fossum, 1966) to later developments (Krauth & Mezard, 1987), (Freund & Schapire, 1999), (Crammer & Singer, 2003), (Shalev-Shwartz et al., 2010).

Large-margin Perceptrons have also been explored in HDC (Kussul et al., 2001), (Kussul & Baidyk, 2004), (Rachkovskij, 2007), (Rachkovskij, 2022), (Smets et al., 2023), (Smets, Rachkovskij, Osipov, Van Leekwijck, et al., 2025), though this line of work remains limited. Notably, (Kussul et al., 2001) introduced a type of margin interpretable as relative or *multiplicative*: the true-class score must exceed competing scores by a fixed proportion of its own value, rather than by a fixed additive constant.

Prior work on multiplicative margin Perceptrons lacks comprehensive descriptions of the method and omits key elements such as associated loss functions, convergence analysis and mistake bounds, behavior on linear non-separable data, design considerations and variants. It has also been restricted to binary input vectors and non-negative class weights, constrained to integer (Kussul et al., 2001), (Kussul & Baidyk, 2004), (Rachkovskij, 2007), (Rachkovskij, 2022) or even binary values (Smets, Rachkovskij, Osipov, Van Leekwijck, et al., 2025).

This article addresses this gap by introducing a family of large-margin, multiclass linear Perceptrons that enforce a multiplicative margin and operate on vector representations of arbitrary format. These models are relevant both to general multiclass linear classification and to HDC-based applications, including resource-constrained settings such as IoT, TinyML, Edge AI, and AIoT. We present the Multiplicative Margin Perceptron (MMPerc) framework, including its loss functions, learning rules, design considerations, and mistake bound analysis.

We evaluate MMPerc on synthetic and real datasets, using both real-valued and binarized original data vectors, as well as their transformations into hypervectors of varying dimensionalities. The primary focus is on comparing its classification accuracy with that of the standard Perceptron. As baselines, we also report results for well-known linear classifiers such as SVM and Ridge. Across most settings, MMPerc outperforms the baselines.

Our main **contributions** are as follows:

1. Architecture and algorithms for multiclass linear Perceptron classifiers with a multiplicative margin mechanism.
2. Basic theoretical analysis of multiclass linear Perceptron classifiers, covering cost functions, weight-update rules, and mistake bounds in both linearly separable and non-separable settings.
3. Key design considerations for large margin Perceptron classifiers, including bias handling, margin threshold selection, online and offline training modes, number of training epochs and stopping criteria, and selection of class prototypes for inference.
4. Experimental evaluation on synthetic and real datasets, using real-valued and binary vector representations across varied dimensionalities.
5. Performance comparison of MMPerc against standard Perceptrons, SVM, Ridge classifiers, along with an assessment of the impact of design considerations on the classification accuracy.
6. Identification of future research directions concerning architectural and training strategies for large-margin Perceptrons.

# 2 Background and Basic Notions

## 2.1 Classification and Training Settings

In the multiclass classification setting, the task is to train a classifier to assign a single class label to a given input data instance. The class label $y^*$ is selected from the label set $\{1,2,...,K\} \equiv [K]$, $K \geq 3$. In a linear classification model, input data are represented as vectors $\mathbf{x}$, e.g., of dimensionality $d$ with real-valued components. The classifier is denoted as $h(\mathbf{x})$: $R^d \rightarrow [K]$. Other types of vector components, such as integer, binary, or ternary, are also possible.

The online learning / training setting assumes the existence of a sequence of labeled data instances (examples, points) represented as pairs $({}^{(t)}\mathbf{x}, {}^{(t)}y)$, $t \in \{1,2,...,T\} \equiv [T]$, where ${}^{(t)}y \in [K]$ is the true (ground truth) class label of ${}^{(t)}\mathbf{x}$. Learning is performed over a sequence of rounds / trials. At round $t$, a classifier $h(.)$ is presented with a single instance ${}^{(t)}\mathbf{x}$ and predicts (outputs hypothesis on) its label ${}^{(t)}y^* = h({}^{(t)}\mathbf{x})$. Then, the classifier observes the correct label ${}^{(t)}y$ and uses this information to improve itself by updating its parameters. So, in the online setting, each instance is accessed only once. Training and testing are interleaved. The goal is to minimize the total number of prediction mistakes throughout the process.

In the offline (or batch) setting, the entire training set of $N$ data instances is available in advance, and each data instance may be used multiple times during training. Training is typically conducted epoch-wise, where each epoch processes $N$ data instances selected by sampling, e.g., with or without replacement. This corresponds to stochastic gradient descent (Robbins & Monro, 1951), (Nemirovski et al., 2009), (Rakhlin et al., 2012), (Gower et al., 2019), (Drori & Shamir, 2020), (Gorbunov et al., 2020), or to shuffling (Recht & Ré, 2013), (Mishchenko et al., 2020), (Safran & Shamir, 2020).

The epochs are repeated until a stopping criterion is met, such as reaching a predefined (possibly zero) number of misclassifications on the training set or completing a specified number of epochs. The goal of training is to produce a classifier that performs well on new, unseen data instances. This assumes that the new data instances are drawn from the same distribution as the training set. Therefore, performance on a separate validation dataset can also be used as a stopping criterion.

For online classification (Shalev-Shwartz & Ben-David, 2014), (Mohri et al., 2018), the *realizable* case means that the class labels in the data sequence have been produced by some classifier $h^*$ from the (known) model family $H$. Let us denote $M_C(H) \equiv M(C,H)$ as the maximal number of mistakes an online classifier $C$ might commit on any arbitrary data sequence (without assumptions about the generating process of ${}^{(t)}\mathbf{x}$ or the ordering of those instances) of any length $T$, labeled by any single $h^* \in H$. This represents the worst-case analysis.

If $M_C(H) \leq M_0 < \infty$, then $M_0$ is called the *mistake bound*. The classifier family $H$ is considered *online learnable* if there exists $C$ that has a mistake bound. Note that this implies that after making at most $M_0$ mistakes (in $T_{M_0} \geq M_0$ rounds), $C$ makes no further mistakes, i.e., it converges. This differs from the "usual" notion of convergence (on a training sequence) in that it assumes zero mistakes not only on the observed data but also on any future data (in the realizable case). The issue, however, is that $M_0$ can be very large.

In the typical offline epoch-wise setting, the size of the dataset is $N < M_0$. In the online learnable case, $C$ will converge after no more than $n_{ep} \le M_0$ epochs in the worst case (assuming it makes a single mistake per epoch to avoid earlier convergence). In the online non-learnable case, $C$ does not converge. In practice, it is then stopped after a predefined number of epochs.

**Notation.** We use $d$ for the original feature vector dimensionality and $D$ for the hypervector dimensionality (after transformation); all mentions of "$D$-dimensional" refer to the latter.

## 2.2 The Model: Linear Multiclass Perceptron of Rosenblatt

Consider a multiclass ($K \geq 3$) single-layer Rosenblatt Perceptron that is linear in its parameters. Let $\mathbf{w}_i$ denote the weight vector for class $i \in [K]$. These weight vectors can be arranged into a matrix $\mathbf{W}$ (of size $K \times d$), where the transpose of $\mathbf{w}_i$ forms the $i$-th row of $\mathbf{W}$. To predict the class (label) of a data instance $\mathbf{x}$, the linear multiclass Perceptron classifier $h(\mathbf{W},\mathbf{x})$ operates as follows. First, it computes the vector of class scores (activations) as $\mathbf{s} = \mathbf{W}\mathbf{x}$. The predicted class label is then determined by the index of the highest score:

$$y^* = h(\mathbf{W},\mathbf{x}) = \text{argmax}_{i \in [K]}\, s_i = \text{argmax}_{i \in [K]}\, \langle \mathbf{w}_i, \mathbf{x} \rangle. \tag{2.1}$$

If there are ties among the highest scores, the class returned by (2.1) may be selected either randomly or according to the argmax implementation. Thus, if the true class is among the tied scores, the prediction may or may not be correct.

Given an instance ${}^{(t)}\mathbf{x}$, in the case of a mistake ${}^{(t)}y^* \neq {}^{(t)}y$, the class weight vectors are updated using the Perceptron error-correction rule (Rosenblatt, 1962):

$$\begin{aligned} {}^{(t+1)}\mathbf{w}_y &= {}^{(t)}\mathbf{w}_y + {}^{(t)}\mathbf{x}, \\ {}^{(t+1)}\mathbf{w}_{y^*} &= {}^{(t)}\mathbf{w}_{y^*} - {}^{(t)}\mathbf{x}. \end{aligned} \tag{2.2}$$

If the prediction is correct, i.e., $h({}^{(t)}\mathbf{W},{}^{(t)}\mathbf{x}) = {}^{(t)}y^* = {}^{(t)}y$, the weights remain unchanged:

$${}^{(t+1)}\mathbf{w}_i = {}^{(t)}\mathbf{w}_i \;\forall i \in [K]. \tag{2.3}$$

In the realizable case of linearly separable data, a Perceptron using this learning rule is guaranteed to converge (Novikoff, 1962), (Rosenblatt, 1962), (Duda & Hart, 1973). Many such update rules were originally proposed as ad-hoc heuristics, commonly referred to as "error correction rules," without formal justification. However, they can be interpreted as solutions to specific optimization problems associated with corresponding loss functions.

## 2.3 Perceptron's Loss Function

Machine learning models are commonly trained by formulating an optimization problem with a loss (objective) that quantifies prediction quality (possibly with constraints), and then adjusting parameters using an appropriate optimization method. For online or complex offline problems, iterative schemes such as gradient/subgradient methods are often used (e.g., (Shor, 1985), (Shalev-Shwartz & Ben-David, 2014), (Mohri et al., 2018)).

A natural approach in the online setting is to minimize the cumulative loss over $T$ rounds:

$$\sum_{t=1,T} {}^{(t)}L. \tag{2.4}$$

Here ${}^{(t)}L = L({}^{(t)}y^*, {}^{(t)}y)$ is the loss for the $t$-th data instance. In the offline setting, one minimizes the empirical risk on the training set

$$(1/N) \sum_{n=1,N} {}^{(n)}L, \tag{2.5}$$

as a proxy for the expected risk (generalization error) on unseen testing set.

A natural loss function for classification tasks is the 0-1 loss:

$${}^{(t)}L_{0\text{-}1}({}^{(t)}y^*, {}^{(t)}y) = 1\{{}^{(t)}y^* \neq {}^{(t)}y\}, \tag{2.6}$$

where ${}^{(t)}y^* = h({}^{(t)}\mathbf{W}, {}^{(t)}\mathbf{x})$. For empirical risk minimization (ERM), classifiers are evaluated with fixed $\mathbf{W}$. Direct minimization with the 0-1 loss is difficult because the indicator function $1\{.\}$ is neither continuous nor convex nor differentiable with respect to the weight vectors $\mathbf{w}_i$, $i \in [K]$, which impedes first-order methods (cf. (Shor, 1985), (Shalev-Shwartz & Ben-David, 2014); but see (Nguyen & Sanner, 2013) for approaches addressing this issue).

For a multiclass Perceptron classifier, let us define the classification score margin as:

$$\delta = \langle \mathbf{w}_y, \mathbf{x} \rangle - \max_{i \neq y} \langle \mathbf{w}_i, \mathbf{x} \rangle = \langle \mathbf{w}_y, \mathbf{x} \rangle - \langle \mathbf{w}_c, \mathbf{x} \rangle = s_y - s_c, \quad (2.7)$$

where $c = \mathrm{argmax}_{i \neq y} \langle \mathbf{w}_i, \mathbf{x} \rangle$ is the class-competitor (the class with the highest score among the incorrect classes for $\mathbf{x}$).

Now, a per-instance Perceptron loss could be defined as:

$$L_{\mathrm{Perc}}(\mathbf{W},(\mathbf{x},y)) = [-\delta]_+ = [\max_{i \neq y} \langle \mathbf{w}_i, \mathbf{x} \rangle - \langle \mathbf{w}_y, \mathbf{x} \rangle]_+ = \max_{i \neq y} [\langle \mathbf{w}_i, \mathbf{x} \rangle - \langle \mathbf{w}_y, \mathbf{x} \rangle]_+, \quad (2.8)$$

where $[z]_+ = \max(0,z)$ denotes the positive part function. Thus, $L_{\mathrm{Perc}} = [-\delta]_+ = 0$ when $\langle \mathbf{w}_y, \mathbf{x} \rangle \geq \max_{i \neq y} \langle \mathbf{w}_i, \mathbf{x} \rangle$ and $-\delta \leq 0$; otherwise $-\delta > 0$ and positive loss is incurred: $L_{\mathrm{Perc}} = [-\delta]_+ = -\delta$. In the two-class ERM case, this loss could be inferred from the classic treatment of (Duda & Hart, 1973).

This Perceptron loss function $L_{\mathrm{Perc}}$ is continuous, monotonically non-increasing, and convex. It is differentiable with respect to the model parameters (the Perceptron class weight vectors), except at the hinge point $\delta = 0$, where subgradients (Shor, 1985) can still be computed. Therefore, subgradient descent can be used to update the Perceptron class weight vectors:

$$^{(t+1)}\mathbf{w}_i = {}^{(t)}\mathbf{w}_i - \eta \, \nabla \mathbf{w}_i \, {}^{(t)}L, \quad (2.9)$$

where the vector $\nabla \mathbf{w}_i \, {}^{(t)}L$ is the subgradient of ${}^{(t)}L = {}^{(t)}L({}^{(t)}\mathbf{W}, ({}^{(t)}\mathbf{x},{}^{(t)}y))$ with respect to $\mathbf{w}_i$, and $\eta$ is the learning rate. The subgradient is evaluated at the current weights ${}^{(t)}\mathbf{W}$ and instance $({}^{(t)}\mathbf{x},{}^{(t)}y)$.

Let us consider some issues associated with the use of this $L_{\mathrm{Perc}}$. It is natural to interpret $L_{\mathrm{Perc}} = [-\delta]_+$ as a mistake indicator: $L_{\mathrm{Perc}} > 0$ signals a mistake, while $L_{\mathrm{Perc}} = 0$ corresponds to correct classification. Then, the subgradients are as follows. When $L_{\mathrm{Perc}} > 0$ (indicating mistake), we have $-\delta > 0$, and thus

$$L_{\mathrm{Perc}} = [-\delta]_+ = -\delta = \max_{i \neq y} \langle \mathbf{w}_i, \mathbf{x} \rangle - \langle \mathbf{w}_y, \mathbf{x} \rangle = \langle \mathbf{w}_c, \mathbf{x} \rangle - \langle \mathbf{w}_y, \mathbf{x} \rangle. \quad (2.10)$$

The corresponding subgradients are

$$\nabla_{\mathbf{w}_c} L_{\text{Perc}} = +\,\mathbf{x},$$
$$\nabla_{\mathbf{w}_y} L_{\text{Perc}} = -\,\mathbf{x}, \tag{2.11}$$
$$\nabla_{\mathbf{w}_i} L_{\text{Perc}} = \mathbf{0} \text{ for } i \in [K]\backslash\{y,c\}.$$

When $L_{\text{Perc}} = [-\delta]_+ = 0$ (indicating correct classification), the loss is constant and equal to zero, and we may use $\nabla_{\mathbf{w}_i} L_{\text{Perc}} = \mathbf{0}$, $\forall i \in [K]$. Using these subgradients and setting $\eta = 1$ in (2.9), the resulting weight updates coincide with the Rosenblatt Perceptron error-correcting rule (2.2), (2.3).

However, several issues arise in connection with $L_{\text{Perc}}$:

(1) $L_{\text{Perc}}$ is not an upper bound for the 0-1 loss $L_{0\text{-}1}$ and, therefore, it is not a *surrogate loss* for $L_{0\text{-}1}$ in the sense of (Shalev-Shwartz & Ben-David, 2014). Consequently, minimizing $\sum_{t=1,T} {}^{(t)}L_{\text{Perc}}$ does not necessarily imply a decrease in the number of classification mistakes $\sum_{t=1,T} {}^{(t)}L_{0\text{-}1}$. For example, it may occur that ${}^{(t)}L_{\text{Perc}}$ get arbitrarily small values $\varepsilon > 0$ while the corresponding ${}^{(t)}L_{0\text{-}1}$ remain maximal, i.e., ${}^{(t)}L_{0\text{-}1} = 1$. A particularly illustrative failure of using $L_{\text{Perc}}$ directly as a practical training objective arises when the weights are initialized to zero. In this case, for any $\mathbf{x}$ we have $\langle \mathbf{w}_i, \mathbf{x}\rangle = 0$ for all $i \in [K]$, so $\delta = 0$ and hence $[-\delta]_+ = L_{\text{Perc}} = 0$. With zero loss, there is no update signal, the Perceptron class weights remain at zero, and training stalls before it begins.

(2) Another issue arises when the true-class score ties one or more competing scores, i.e., when $\langle \mathbf{w}_y, \mathbf{x}\rangle = \max_{i\neq y} \langle \mathbf{w}_i, \mathbf{x}\rangle$ for some $i$. Here again $\delta = 0$ and $L_{\text{Perc}} = 0$, so no update is triggered. In contrast, the Perceptron can still update at $\delta = 0$ and $L_{\text{Perc}} = 0$ whenever the class prediction given by (2.1) is a mistake, thereby enabling Perceptron training in practice.

One fix that recovers the Perceptron error-correction rule is to avoid treating $L_{\text{Perc}} = 0$ as certifying correctness and instead rely on the prediction given by (2.1). Then, the subgradients of $L_{\text{Perc}}$ are still used to derive the update rule. When a mistake occurs at $\delta = 0$, where $L_{\text{Perc}}$ is nondifferentiable; any admissible subgradient vector at that point may be applied, specifically, the mistake-case subgradient (2.11) is used. Notably, in the ERM setting, (Duda & Hart, 1973) defined the cumulative loss as the sum of ${}^{(n)}L_{\text{Perc}}$ taken only over those training instances $n$ on which a mistake occurs.

Despite this fix, which allows using $L_{\text{Perc}}$ to recover the Perceptron's error-correcting rule (2.2)-(2.3), $L_{\text{Perc}}$ still does not upper bound the 0–1 loss. Certain generalization guarantees for ERM with surrogate losses rely on the surrogate upper-bounding the target loss; when this condition is not met, those guarantees need not apply. Moreover, for an upper-bounding surrogate, small (or zero) surrogate loss immediately implies small (or zero) target loss; $L_{\text{Perc}}$ lacks this property.

Let us consider some additional issues associated with Perceptron training:

(3) As noted in section 2.2, in the realizable case of linearly separable data, a Perceptron trained with the error-correcting rule converges. However, the resulting class boundaries at convergence may lie arbitrary close to training instances while still separating the classes. This occurs because correct classification requires only that $\delta \geq 0$, regardless of its magnitude. Enforcing a positive margin threshold ($\delta \geq \beta > 0$) increases the separation between the decision boundary and the training instances and can improve generalization by reducing testing error. This principle underlies large-margin classifiers such as SVM and related models.

(4) A related concern is noise sensitivity when the margin threshold is zero: when class scores are close, small fluctuations can arbitrarily flip the predicted class, yielding either no update or inconsistent updates for the same input vector and weights during training. This issue was recognized in early work (Mays, 1964), (Duda & Hart, 1973) and addressed in some later studies, e.g., (Khardon & Wachman, 2007).

## 2.4 Large-Margin Perceptrons

The issues outlined above have been addressed within the framework of large-margin Perceptrons. In the setting we focus on, a positive threshold is applied to the classification score margin. During training, this threshold replaces the Perceptron's original zero-margin criterion for deciding whether a prediction is correct. By requiring higher confidence in correct predictions, this approach encourages decision boundaries to lie farther from the training instances, potentially improving generalization.

For the margin type we call an *additive margin*, the two-class case (which uses a single weight vector; see (Mays, 1964), (Tsampouka & Shawe-Taylor, 2005)) extends naturally to the multiclass setting as follows. A mistake occurs when $\delta < \beta$, with $\beta > 0$, i.e., $\delta = \langle \mathbf{w}_y, \mathbf{x} \rangle - \max_{i \neq y} \langle \mathbf{w}_i, \mathbf{x} \rangle < \beta$, which can equivalently be written as $(\langle \mathbf{w}_y, \mathbf{x} \rangle - \beta)$ $< \max_{i \neq y} \langle \mathbf{w}_i, \mathbf{x} \rangle$. We refer to this as an additive margin because the threshold value $\beta$ is subtracted. Setting $\beta = 0$ recovers the standard Perceptron, except in the case $\delta = 0$.

For the margin type we call a *multiplicative margin*, a mistake occurs when $(\langle \mathbf{w}_y, \mathbf{x} \rangle - \max_{i \neq y} \langle \mathbf{w}_i, \mathbf{x} \rangle) / \langle \mathbf{w}_y, \mathbf{x} \rangle < \alpha$, $0 \leq \alpha < 1$ (Kussul et al., 2001). This condition can be rewritten as $\delta < \alpha \langle \mathbf{w}_y, \mathbf{x} \rangle$ or $(1 - \alpha) \langle \mathbf{w}_y, \mathbf{x} \rangle < \max_{i \neq y} \langle \mathbf{w}_i, \mathbf{x} \rangle$. In this formulation, the margin $\delta$ is compared to a fraction $\alpha$ of the true class score $\langle \mathbf{w}_y, \mathbf{x} \rangle$; we, therefore, refer to it as a multiplicative margin. Setting $\alpha = 0$ again recovers the standard Perceptron, except in the case $\delta = 0$. Note that the ratio form above requires $\langle \mathbf{w}_y, \mathbf{x} \rangle \neq 0$, see relevant discussions in sections 3.1.2.1 and 3.4.2.

The first explicit mention of a Perceptron with a multiplicative margin in English appeared in (Kussul et al., 2001), see also (Kussul & Baidyk, 2004), (Rachkovskij, 2007). A more detailed overview of related work, including Rosenblatt's Perceptrons, multiclass linear Perceptrons, large-margin Perceptrons, fixed input nonlinear transformations, and HDC classifiers, is provided in Supplementary Note 1.

# 3 Method

In this section, we present loss functions and learning rules for multiclass large margin Perceptrons (section 3.1), mistake bound analysis (section 3.2 and 3.3), and discussion (section 3.4).

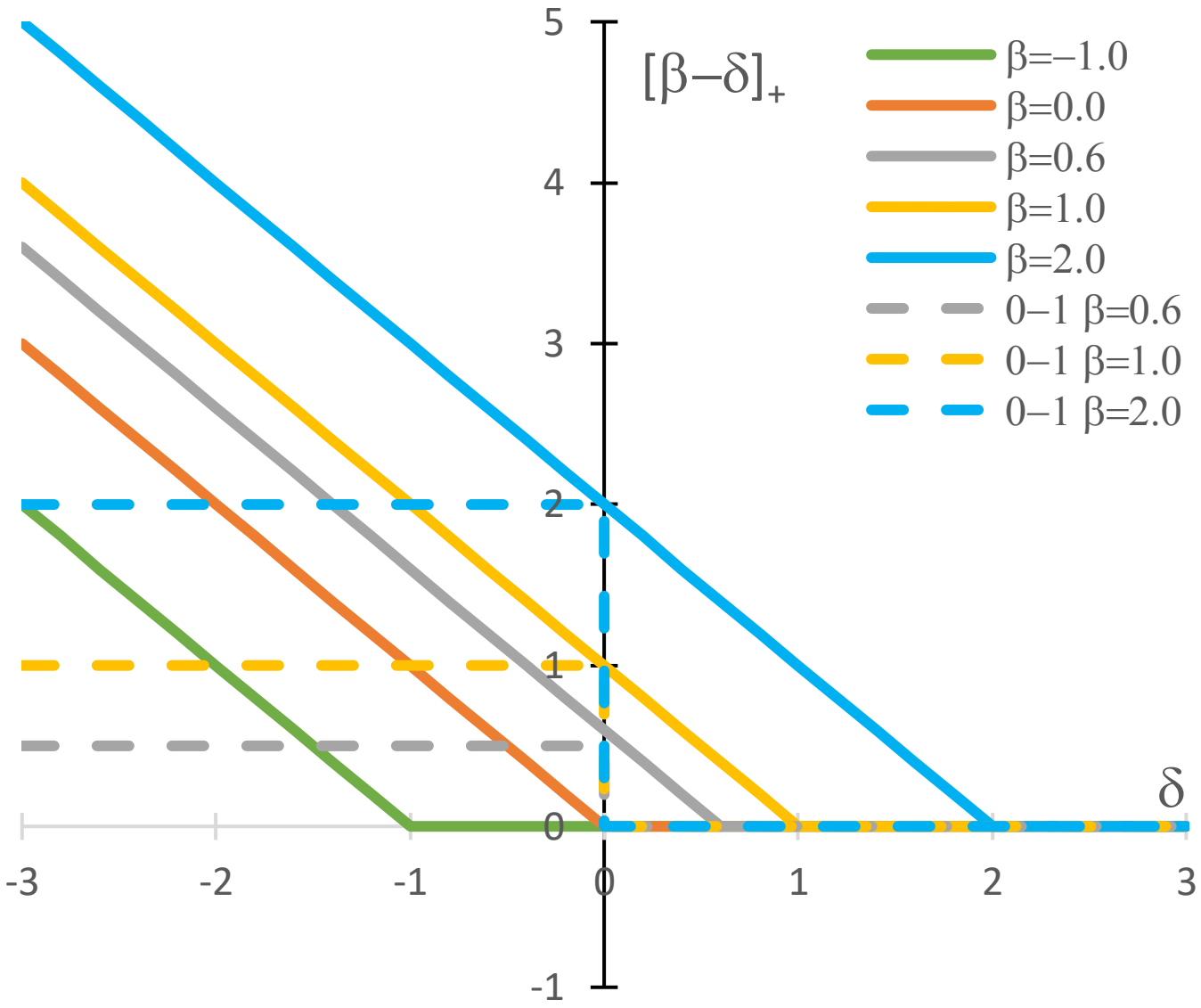

Figure 1: The $\beta$-hinge loss (solid lines) upper bounds the 0-1 loss scaled by $\beta$ (dashed lines).

## 3.1 Loss Functions and Learning Rules for Multiclass Large-Margin Perceptrons

While our primary focus is on multiclass linear large-margin Perceptrons with a multiplicative margin, we also present the multiclass additive margin formulation for completeness, since it is typically treated in the literature only in the two-class setting and to highlight specific aspects of our formulation.

### 3.1.1 Additive Margin Formulation

For the large margin multiclass Perceptron with an additive margin (AMPerc), we define the loss function as the *multiclass* β-*hinge loss* in the sense of (Crammer & Singer, 2001):

$$L_{\beta\text{-hinge}} \equiv [\beta - \delta]_+ = [\beta - (s_y - \max_{i \neq y} s_i)]_+ = [\max_{i \neq y} s_i - (s_y - \beta)]_+ = \\ [\max_{i \neq y} \langle \mathbf{w}_i, \mathbf{x} \rangle - (\langle \mathbf{w}_y, \mathbf{x} \rangle - \beta)]_+. \tag{3.1}$$

For $\beta = 0$, we recover the multiclass Perceptron loss: $L_{0\text{-hinge}} = L_{\text{Perc}}$ (with its drawbacks noted in section 2.3). For $\beta = 1$, we obtain the 1-hinge loss, which corresponds to the multiclass hinge loss function of (Crammer & Singer, 2001): $L_{1\text{-hinge}} = [1 - \delta]_+$. For any $\beta > 0$, $L_{\beta\text{-hinge}}$ upper bounds the scaled 0-1 loss: $L_{\beta\text{-hinge}} \geq \beta\, L_{0\text{-}1}$. Thus, this loss is not only convex (as well as monotonic and continuous, like $L_{\text{Perc}}$), but also qualifies as a convex *surrogate loss* (Shalev-Shwartz & Ben-David, 2014) for the scaled 0-1 loss $\beta\, L_{0\text{-}1}$, see Figure 1.

Now consider the use of $L_{\beta\text{-hinge}}$ for mistake detection during training and for subgradient calculation in deriving the AMPerc update rule. The loss $L_{\beta\text{-hinge}}$ is zero when $\beta - \delta \leq 0$, i.e., $\delta \geq \beta$, and positive when $\delta < \beta$. These correspond to correct and incorrect classification in AMPerc, respectively, as introduced in section 2.3.

Note that if $\mathbf{W} = \mathbf{0}$ and therefore $\delta = 0$, then $L_{\beta\text{-hinge}} = \beta > 0$, which signals a mistake. In contrast, as discussed earlier, $L_{\text{Perc}} = L_{0\text{-hinge}} = 0$ when $\mathbf{W} = \mathbf{0}$, providing no learning signal and effectively preventing training from starting. Furthermore, when $\mathbf{W} \neq \mathbf{0}$ but $\delta = 0$, $L_{\beta\text{-hinge}}$ again signals the need for an update.

To derive the AMPerc learning rule, subgradients are calculated as follows. In the mistake case ($\beta - \delta > 0$):

$$0 < L_{\beta\text{-hinge}} = \beta + \max_{i \neq y} \langle \mathbf{w}_i, \mathbf{x} \rangle - \langle \mathbf{w}_y, \mathbf{x} \rangle = \beta + \langle \mathbf{w}_c, \mathbf{x} \rangle - \langle \mathbf{w}_y, \mathbf{x} \rangle. \quad (3.2)$$

Thus, the resulting subgradients are the same as in (2.11).

In the case of correct classification ($\beta - \delta \leq 0$), the loss is constant and zero: $L_{\beta\text{-hinge}} = 0$. Then $\nabla \mathbf{w}_i\, L = 0$ for all $i \in [K]$. At the hinge point $\delta = \beta$, the subgradient is chosen to be the zero vector. Thus, in AMPerc, weights are updated only when a mistake occurs, modifying the weight vectors of the two classes $y$ and $c$ exactly as in the standard Perceptron, see (2.2)-(2.3). During training, AMPerc detects mistakes using the condition $L_{\beta\text{-hinge}} > 0$. During evaluation, however, it relies on the 0–1 loss, with $y^*$ from (2.1), as the mistake indicator.

### 3.1.2 Multiplicative Margin Formulation

For MMPerc, we define the loss function as the *multiclass* $\alpha s$*-hinge loss*:

$$L_{\alpha s\text{-hinge}} = [\alpha\, s_y - \delta]_+ = [\alpha\, s_y - (s_y - \max_{i \neq y} s_i)]_+ = [\max_{i \neq y} s_i - s_y\, (1 - \alpha)]_+ = \\ = [\max_{i \neq y} \langle \mathbf{w}_i, \mathbf{x} \rangle - \langle \mathbf{w}_y, \mathbf{x} \rangle\, (1 - \alpha)]_+, \text{ where } 0 < \alpha < 1. \quad (3.3)$$

For $\alpha = 0$, we recover the multiclass Perceptron loss: $L_{0s\text{-hinge}} = L_{\text{Perc}}$.

The criterion for mistake detection during training is $\delta < \alpha s_y = \alpha\langle \mathbf{w}_y, \mathbf{x} \rangle$, which corresponds to $L_{\alpha s\text{-hinge}} > 0$, whereas $\delta \geq \alpha s_y = \alpha\langle \mathbf{w}_y, \mathbf{x} \rangle$ and $L_{\alpha s\text{-hinge}} = 0$ indicate correct classification. This formulation directly addresses the issues of the standard Perceptron discussed in section 2.3.

In (Kussul et al., 2001; Kussul & Baidyk, 2004; Rachkovskij, 2007, 2022, 2024), weight updates in the case of a mistake followed the standard Perceptron rule (2.2)-(2.3). In evaluation mode, the trained MMPerc predicts labels exactly as the standard Perceptron (2.1). Now, let us derive the learning rule directly from the subgradients of $L_{\alpha s\text{-hinge}}$. In the case of a mistake, $0 < L_{\alpha s\text{-hinge}} = \langle \mathbf{w}_c, \mathbf{x} \rangle - \langle \mathbf{w}_y, \mathbf{x} \rangle\, (1 - \alpha)$. Then, the subgradients are:

$$\nabla \mathbf{w}_y\, L = -\,\mathbf{x}\, (1 - \alpha), \\ \nabla \mathbf{w}_c\, L = +\,\mathbf{x}, \quad (3.4) \\ \nabla \mathbf{w}_i\, L = 0\ \forall i \in [K] \setminus \{y, c\}.$$

Thus, the update rule becomes:

$$^{(t+1)}\mathbf{w}_y = {}^{(t)}\mathbf{w}_y + {}^{(t)}\mathbf{x}\,(1-\alpha);$$
$$^{(t+1)}\mathbf{w}_c = {}^{(t)}\mathbf{w}_c - {}^{(t)}\mathbf{x}; \qquad (3.5)$$
$$^{(t+1)}\mathbf{w}_i = {}^{(t)}\mathbf{w}_i \ \forall i \in [K]\backslash\{y,c\} \text{ (and } \forall i \in [K] \text{ in the case of no mistake).}$$

We refer to this MMPerc variant as the large margin multiclass Perceptron with the asymmetric learning rule (MMPerc-Asm).

#### *3.1.2.1 The asymmetric case with absolute true-class score*

When the components of input vectors **x** and class weights **w** are restricted to be non-negative (Kussul et al., 2001; Kussul & Baidyk, 2004), see also (Magri, 2015), then we have $s_y \geq 0$, and thus $s_y - \alpha\, s_y \leq s_y$. However, for arbitrary **x** and **w**, it is possible that $s_y < 0$. In this case, $-\alpha\, s_y = \alpha\, |s_y|$ and $s_y - \alpha\, s_y = s_y + \alpha\, |s_y|$, which increases rather than decreases the true-class score. Moreover, if $s_y < 0$, then misclassification cannot occur for any $\delta \geq 0$. To address this issue (but see also section 3.4.2), we propose to modify the true class score $s_y$ during training as $s_y - \alpha\, |s_y|$, ensuring $s_y - \alpha\, |s_y| \leq s_y$. The corresponding loss becomes

$$L_{\alpha s\text{-hinge-abs}} = L_{\alpha|s|\text{-hinge}} =$$
$$[\,\alpha\,|s_y| - \delta]_+ = [\,\alpha\,|s_y| - (s_y - \max_{i\neq y} s_i)\,]_+ = [\max_{i\neq y} s_i - (s_y - \alpha\,|s_y|)\,]_+ = \qquad (3.6)$$
$$[\langle \mathbf{w}_c,\mathbf{x}\rangle - \langle \mathbf{w}_y,\mathbf{x}\rangle + \alpha\,|\langle \mathbf{w}_y,\mathbf{x}\rangle|]_+ = [\langle \mathbf{w}_c, \mathbf{x}\rangle - \langle \mathbf{w}_y, \mathbf{x}\rangle + \alpha\,\langle \mathbf{w}_y, \mathbf{x}\rangle \operatorname{sign}\langle \mathbf{w}_y, \mathbf{x}\rangle]_+,$$

where sign(.) denotes the standard sign function.

A positive incurred loss $L_{\alpha|s|\text{-hinge}} > 0$ indicates a mistake, while $L_{\alpha|s|\text{-hinge}} \leq 0$ corresponds to correct classification. If the update rule is the standard Perceptron one, we refer to this variant as the multiclass Perceptron with an absolute multiplicative margin (MMPerc-Abs).

Now consider the version with an absolute multiplicative margin and an asymmetric learning rule (MMPerc-AbsAsm). In case of a mistake, the incurred loss is positive, and we have:

$$0 < L_{\alpha\text{s-hinge-abs}} = \langle \mathbf{w}_c, \mathbf{x}\rangle - (\langle \mathbf{w}_y, \mathbf{x}\rangle - \alpha\,|\langle \mathbf{w}_y, \mathbf{x}\rangle|) =$$
$$\langle \mathbf{w}_c, \mathbf{x}\rangle - \langle \mathbf{w}_y, \mathbf{x}\rangle + \alpha\,\langle \mathbf{w}_y, \mathbf{x}\rangle \operatorname{sign}\langle \mathbf{w}_y, \mathbf{x}\rangle = \qquad (3.7)$$
$$\langle \mathbf{w}_c, \mathbf{x}\rangle - \langle \mathbf{w}_y, \mathbf{x}\rangle\,(1 - \alpha \operatorname{sign}\langle \mathbf{w}_y, \mathbf{x}\rangle).$$

The subgradients are:

$$\nabla \mathbf{w}_y L = -(\mathbf{x} - \mathbf{x}\ \alpha \text{ sign } \langle \mathbf{w}_y, \mathbf{x}\rangle) = -\mathbf{x}\ (1 - \alpha \text{ sign } \langle \mathbf{w}_y, \mathbf{x}\rangle),$$
$$\nabla \mathbf{w}_c L = +\mathbf{x}, \qquad (3.8)$$
$$\nabla \mathbf{w}_i L = 0\ \forall i \in [K] \backslash \{y,c\}.$$

Accordingly, the weights are updated as

$^{(t+1)}\mathbf{w}_y = \mathbf{w}_y^{(t)} + (\mathbf{x} - \alpha\ \mathbf{x} \text{ sign } a_y) = \mathbf{w}_y^{(t)} + \mathbf{x}\ (1 - \alpha \text{ sign } \langle \mathbf{w}_y, \mathbf{x}\rangle)$, i.e.,

$$^{(t+1)}\mathbf{w}_y = {}^{(t)}\mathbf{w}_y + {}^{(t)}\mathbf{x}\ (1 - \alpha), \text{ if sign } \langle \mathbf{w}_y, \mathbf{x}\rangle = +1;$$
$$^{(t+1)}\mathbf{w}_y = {}^{(t)}\mathbf{w}_y + {}^{(t)}\mathbf{x}\ (1 + \alpha), \text{ if sign } \langle \mathbf{w}_y, \mathbf{x}\rangle = -1;$$
$$^{(t+1)}\mathbf{w}_y = {}^{(t)}\mathbf{w}_y + {}^{(t)}\mathbf{x}, \text{ if sign } \langle \mathbf{w}_y, \mathbf{x}\rangle = 0; \qquad (3.9)$$
$$^{(t+1)}\mathbf{w}_c = {}^{(t)}\mathbf{w}_c - {}^{(t)}\mathbf{x};$$
$$^{(t+1)}\mathbf{w}_i = {}^{(t)}\mathbf{w}_i\ \forall i \notin \{y,c\}.$$

In the following sections, we analyze several variants of MMPerc. For completeness, we also include the AMPerc variant. Section 3.2 presents mistake bounds (as defined in section 2.1) for the realizable case of linearly separable classes, under the classic online learning setting introduced in section 2.1. The linearly non-separable case is discussed in section 3.3.

### 3.2 Mistake Bounds for the Linearly Separable Case

To derive mistake upper bounds for different large-margin Perceptrons, we build upon the classic analysis of the margin-free two-class Perceptron (Novikoff, 1962 and its extensions to the multiclass setting (Crammer & Singer, 2003), (Beygelzimer et al., 2019).

Let us denote by $\mathbf{V}_{\backslash\{y,c\}}$ the matrix $\mathbf{V}$ without rows $y$ and $c$. The Frobenius norm of a matrix $\mathbf{V}$ is denoted by $\|\mathbf{V}\|$, while the Euclidean norm of a vector $\mathbf{v}$ is $\|\mathbf{v}\|$. The Frobenius dot product of two matrices $\mathbf{V}$, $\mathbf{W}$ is denoted by $\langle \mathbf{V}, \mathbf{W}\rangle$.

Consider $T$ data instances:

$$\{{}^{(t)}\mathbf{x}, {}^{(t)}y\},\ t \in [T], \text{ with } \|{}^{(t)}\mathbf{x}\| \le R, \qquad (3.10)$$

and ${}^{(t)}y \in [K]\ \forall t \in [T]$. The data instances are assumed to be linearly separable with an additive margin $\gamma > 0$, meaning there exist weight vectors $\mathbf{u}_1, \mathbf{u}_2, \ldots, \mathbf{u}_K$ (which can be viewed as rows of a matrix $\mathbf{U}$) such that:

$$\langle {}^{(t)}\mathbf{x}, \mathbf{u}_{y(t)} \rangle - \langle \mathbf{x}^{(t)}, \mathbf{u}_i \rangle \geq \gamma,\ \forall i \in [K] \backslash \{{}^{(t)}y\},\ \forall t \in [T]. \tag{3.11}$$

Note that (3.11) is equivalent to ${}^{(t)}L_{\gamma\text{-hinge}}(\mathbf{U}, ({}^{(t)}\mathbf{x}, {}^{(t)}y)) = [\gamma - {}^{(t)}\delta]_+ = 0\ \forall t \in [T]$.
Let us further assume

$$||\mathbf{U}|| = 1. \tag{3.12}$$

3.2.1 Theorem 1 (Mistake bounds for large margin Perceptrons in the linearly separable case):

Using the definitions and notations introduced above, the number of mistakes made by the various large-margin multiclass Perceptron variants is upper-bounded by the following values $M_0$:

(1) Multiclass Perceptron with an additive margin $\beta$ (AMPerc): $M_0 = 2(R^2+\beta)/\gamma^2$.

(2) Multiclass Perceptron with a multiplicative margin $\alpha\, s_y$ and the standard Perceptron learning rule (2.2)-(2.3) (MMPerc): $M_0 = R^2\,(2 - \alpha) / (\gamma^2 - \alpha\, R^2)$, where $\gamma^2 - \alpha\, R^2 > 0$.

(3) Multiclass Perceptron with a multiplicative margin $\alpha\ |s_y|$ and the standard Perceptron learning rule (2.2)-(2.3) (MMPerc-Abs): $M_0 = R^2\,(2 - \alpha) / (\gamma^2 - \alpha\, R^2)$, where $\gamma^2 - \alpha\, R^2 > 0$.

(4) Multiclass Perceptron with a multiplicative margin $\alpha\, s_y$ and the asymmetric learning rule (3.5) (MMPerc-Asm): $M_0 = (1 + (1 - \alpha)^2)\, R^2 / (\gamma - \alpha\, R)^2$, where $\gamma - \alpha\, R > 0$.

(5) Multiclass Perceptron with an "absolute" multiplicative margin $\alpha\ |s_y|$ and the asymmetric learning rule (3.9) (MMPerc-AbsAsm): $M_0 = (1 + (1 + \alpha)^2)\, R^2 / (\gamma - \alpha\, R)^2$, where $\gamma - \alpha\, R > 0$.

**Proof.** See Supplementary Note 2.1.

By setting the margin threshold parameters $\alpha$ or $\beta$ to zero in the above formulas for multi-level Perceptrons, we recover the mistake bound of the standard multiclass Perceptron:

$$M_0 = 2R^2/\gamma^2, \tag{3.13}$$

as established in prior work (Crammer & Singer, 2003), (Beygelzimer et al., 2019). As discussed in section 2.1, the existence of such a bound in the linear separable case also implies convergence of training in the offline setting.

From the mistake bounds (2)-(3) for MMPerc and MMPerc-Abs, we observe that a finite and positive upper bound exists only if $\gamma^2 - \alpha R^2 > 0$, i.e., $\gamma^2 / R^2 > \alpha$. Similarly, from the mistake bounds (4)-(5) for MMPerc-Asm and MMPerc-AbsAsm, the condition becomes $\gamma - \alpha R > 0$, i.e., $\gamma / R > \alpha$. Since $\gamma$ is generally unknown, selecting an appropriate value of $\alpha$ is necessary to ensure the existence of a valid upper bound. Additionally, noting that $\gamma^2 - \alpha R^2 \leq \gamma^2$, and $\gamma - \alpha R \leq \gamma$, we see that introducing a margin increases the upper bound on the number of mistakes for large-margin Perceptrons compared to the standard Perceptron. On the other hand, if the upper bound becomes non-positive according to these formulas, alternative convergence criteria may be considered, such as those based on step-wise convergence (Tsampouka & Shawe-Taylor, 2005). Furthermore, an increase in the number of mistakes does not necessarily imply slower convergence, whether measured in epochs or even in the number of trials with mistakes. It may lead to solutions with larger margins, as discussed in section 3.4.2.

### 3.3 Mistake Bounds for the Linearly Non-Separable Case

In the non-realizable case of linearly non-separable data (section 2.1), we analyze the number of mistakes made by multiclass Perceptrons with margin, relative to the performance of any fixed linear classifier ("competitor"), possibly the best in hindsight. We adopt the analysis framework developed for the margin-less multiclass Perceptron in the online setting, as discussed in, e.g., (Beygelzimer et al., 2017) (full-information setting), (Fink et al., 2006; Kakade et al., 2008). See also the two-class case in (Mohri & Rostamizadeh, 2013), (Shalev-Shwartz & Ben-David, 2014), (Mohri et al., 2018).

The performance of the competitor linear classifier with weight matrix $\mathbf{U}$ is measured in terms of the multiclass $b$-hinge loss, defined as:

$L_{b\text{-hinge}}(\mathbf{U}, (\mathbf{x},y)) \equiv L_b(\mathbf{U}, (\mathbf{x},y)) = [b - \delta(\mathbf{U},\mathbf{x})]_+ = [b - (\langle \mathbf{u}_y, \mathbf{x}\rangle - \max_{i \neq y} \langle \mathbf{u}_i, \mathbf{x}\rangle)]_+$
$= [b - (\langle \mathbf{u}_y, \mathbf{x}\rangle - \langle \mathbf{u}_c, \mathbf{x}\rangle)]_+$, where $c = \operatorname{argmax}_{i \neq y} \langle \mathbf{u}_i, \mathbf{x}\rangle$.

3.3.1 Theorem 2 (Mistake bounds for large margin Perceptrons in the linearly non-separable case):

Given any sequence of $T$ data instances $\{{}^{(t)}\mathbf{x}, {}^{(t)}y\}$, $t \in [T]$, where $\|{}^{(t)}\mathbf{x}\| \le R$ (3.10), and ${}^{(t)}y \in [K]$ $\forall t \in [T]$, and any (competitor) multiclass linear classifier with weight vectors $\mathbf{u}_1, \mathbf{u}_2, \ldots, \mathbf{u}_C$ (rows of matrix $\mathbf{U}$), the number of mistakes made by the various large-margin multiclass Perceptron variants is upper-bounded by $M_0$ as follows.

Let $L_b \equiv L_b(\mathbf{U}) \equiv \sum_{m=1,M} L_b\,(\mathbf{U}, ({}^{(m)}\mathbf{x}, {}^{(m)}y))$ be the cumulative $b$-hinge loss of the competitor on the instances where a large-margin multiclass Perceptron variant makes a mistake. Then:

(1) Multiclass Perceptron with additive margin β (AMPerc):

$M_0 = (1/b^2)\ \{b\,L_b + 2\,\|\mathbf{U}\|^2\,(R^2 + \beta) + [2\,b\,L_b\,\|\mathbf{U}\|^2\,(R^2 + \beta)]^{1/2}\}$.

(2) Multiclass Perceptron with multiplicative margin $\alpha s_y$ and the Perceptron learning rule (2.2)-(2.3) (MMPerc):

$M_0 = \{b\,L_b + (2–\alpha)\,R^2\|\mathbf{U}\|^2 + [bL_b\,(2–\alpha)\,R^2\,\|\mathbf{U}\|^2 + \alpha R^2\,\|\mathbf{U}\|^2\,L_b^2]^{1/2}\}/\,(b^2 – \alpha R^2\,\|\mathbf{U}\|^2)$.

(3) Multiclass Perceptron with multiplicative margin $\alpha|s_y|$ and the Perceptron learning rule (2.2)-(2.3) (MMPerc-Abs):

$M_0 = \{b\,L_b + (2–\alpha)\,R^2\|\mathbf{U}\|^2 + [bL_b\,(2–\alpha)\,R^2\,\|\mathbf{U}\|^2 + \alpha R^2\,\|\mathbf{U}\|^2\,L_b^2]^{1/2}\}/\,(b^2 – \alpha R^2\,\|\mathbf{U}\|^2)$.

(4) Multiclass Perceptron with multiplicative margin $\alpha s_y$ and the asymmetric learning rule (3.5) (MMPerc-Asm):

$M_0 = \{(b – \alpha\|\mathbf{U}\|R)\,L + (2 – 2\alpha + \alpha^2)\,R^2\,\|\mathbf{U}\|^2 + [(b – \alpha\|\mathbf{U}\|R)\,L\,(2 – 2\alpha + \alpha^2)\,R^2\,\|\mathbf{U}\|^2]^{1/2}\}\,/\,(b – \alpha\|\mathbf{U}\|R)^2$.

(5) Multiclass Perceptron with "absolute" multiplicative margin $\alpha|s_y|$ and the asymmetric learning rule (3.9) (MMPerc-AbsAsm):

$M_0 = \{(b – \alpha\|\mathbf{U}\|R)\,L + (2 + 2\alpha + \alpha^2)\,R^2\,\|\mathbf{U}\|^2 + [(b – \alpha\|\mathbf{U}\|R)\,L\,(2 + 2\alpha + \alpha^2)\,R^2\,\|\mathbf{U}\|^2]^{1/2}\}\,/\,(b – \alpha\|\mathbf{U}\|R)^2$.

**Proof.** See Supplementary Note 2.2.

The mistake bounds for the linearly non-separable case differ from those of the separable case in that the former are not fixed constants but grow with the number of training instances $T$. Setting the margin threshold parameters α or β to zero in the above formulas for large-margin Perceptrons, we obtain the mistake bound:

$M_0 = \{b\,L_b + 2\,R^2\,||\mathbf{U}||^2 + (b\,L_b\,2\,R^2\,||\mathbf{U}||^2)^{1/2}\} / b^2$.

Now, setting $b$=1 so that $L_1$ is the cumulative 1-hinge loss of the competitor, we recover the known mistake bound of the standard multiclass Perceptron in the linearly non-separable case: $M_0 = L_1 + 2\,||\mathbf{U}||^2\,R^2 + (2\,L_1\,||\mathbf{U}||^2\,R^2)^{1/2}$, as established in (Fink et al., 2006), (Kakade et al., 2008), (Beygelzimer et al., 2017).

In the linearly separable case, where the cumulative 1-hinge loss of the competitor is zero ($L_1 = 0$), this simplifies to $M_0 = 2\,||\mathbf{U}||^2\,R^2$. The condition $L_{1\text{-hinge}}(\mathbf{U},(\mathbf{x},y)) = [1-(\langle \mathbf{u}_y, \mathbf{x}\rangle - \langle \mathbf{u}_c, \mathbf{x}\rangle)]_+ = 0$ any input $\mathbf{x}$ implies that $(\langle \mathbf{u}_y, \mathbf{x}\rangle - \langle \mathbf{u}_c, \mathbf{x}\rangle) \geq 1$. By normalizing the competitor's weights as $\mathbf{U}/||\mathbf{U}||$ we obtain linear separability with additive margin $\gamma = 1/||\mathbf{U}||$. Thus, we recover the classical mistake bound for the linearly separable case (3.13): $M_0 = 2\,R^2 / \gamma^2$.

An extension to other loss functions of the competitor can be considered, such as any power of the multiclass hinge loss between 1 and 2, via Hölder's inequality (Beygelzimer et al., 2017), or to a family of admissible convex losses (Mohri & Rostamizadeh, 2013), (Mohri et al., 2018).

## 3.4 Discussion

### 3.4.1 On Generalization Error

Generalization bounds for algorithms typically require the data instances to be i.i.d. (independently and identically distributed) samples from some underlying distribution. One approach, proposed in (Cesa-Bianchi et al., 2004), establishes bounds on the expected loss on future instances by selecting the classifier obtained at round $t$ that minimizes a *penalized empirical risk*. The loss function may be the number of mistakes, thereby yielding generalization error bounds. This framework enables the incorporation of the mistake bounds derived in previous sections to obtain generalization bounds for the corresponding multiclass Perceptron variants.

More broadly, the study of generalization behavior and bounds remains a promising direction, particularly in the modern context of *benign overfitting* in overparameterized models, as explored, e.g., in (Bartlett & Long, 2021; Gastpar et al., 2023). Overparameterization is a common scenario for linear models when they are applied to high-dimensional data representations such as hypervectors in HDC.

### 3.4.2 Additive vs Multiplicative Margin

**Learning rate.** Categorization of Perceptron algorithms based on the behavior of their "*effective*" learning rate and margin threshold during training has been discussed in (Tsampouka & Shawe-Taylor, 2005). In the Perceptron variants we consider, a constant nominal learning rate $\eta = 1$ is used. However, if we view the learning rate in terms of the relative increment in the norm of the weights vectors, this *effective* learning rate decreases over the course of training. This occurs because the norm of the class weight vectors grows as training progresses, while the norms of the data instance vectors remain fixed. As a result, the relative impact of each individual update diminishes over time.

For multiclass Perceptrons, we have $||^{(M+1)}\mathbf{W}|| \geq \gamma M$. When a mistake occurs on $\mathbf{x}$, $^{(M+1)}\mathbf{W}$ is updated with $\Delta\mathbf{W}$, which consists of terms $\eta\mathbf{x}$ in the rows $y$ and $y^*$, regardless of $M$. The relative norm change $||\Delta\mathbf{W}|| \;/\; ||^{(M+1)}\mathbf{W}|| \leq (||\mathbf{x}||\sqrt{2}/\gamma)\ (\eta/M)$. Here, $\eta/M$ can be interpreted as an effective learning rate, which decreases linearly with the number of mistakes $M$. This behavior can be viewed as a natural form of learning-rate decay: as training progresses and the norms of the class weight vectors grow, each individual update has a smaller relative effect. This, in turn, provides a self-regulating mechanism that contributes to stabilization and convergence of the algorithm, reducing overshooting and allowing the decision boundary to stabilize.

**Margin thresholds.** In AMPerc, misclassification occurs when $\delta(^{(t)}\mathbf{W}, {}^{(t)}\mathbf{x}) = \langle^{(t)}\mathbf{x}, {}^{(t)}\mathbf{w}_y\rangle - \langle^{(t)}\mathbf{x}, {}^{(t)}\mathbf{w}_c\rangle < \beta$. Since weight values grow during training, the margin $\delta$ also tend to increase for the same $\mathbf{x}$, while the threshold $\beta$ remains constant. This affects the classification outcome. To quantify it, let us introduce the effective margin $\delta \;/\; ||^{(M+1)}\mathbf{W}||$, so that scaling of $\mathbf{W}$ does not change the effective margin value. The misclassification condition preserving the same classification decision as before becomes $(\langle\mathbf{x}, {}^{(M+1)}\mathbf{w}_y\rangle - \langle\mathbf{x}, {}^{(M+1)}\mathbf{w}_c\rangle) \;/\; ||^{(M+1)}\mathbf{W}|| < \beta \;/\; ||^{(M+1)}\mathbf{W}||$. Since $||^{(M+1)}\mathbf{W}|| \geq \gamma M$, this implies $\beta \;/\; ||^{(M+1)}\mathbf{W}|| \leq (1/\gamma)\ (\beta/M)$. Thus, the effective margin threshold $\beta_{\text{eff}} = \beta \;/\; ||^{(M+1)}\mathbf{W}||$ decreases with the number of mistakes $M$ made during training.

This self-regulating mechanism is analogous to the decrease of the effective learning rate discussed above. It supports the stabilization and convergence (in linearly separable cases) of the algorithm by preventing overshooting and allowing the decision boundaries to stabilize. However, the decrease of $\beta_{eff}$ contradicts the original motivation for introducing a margin in the Perceptron framework (section 2.3), especially when $M$ becomes large.

Let us consider the value $\beta_{eff} / \gamma$ for AMPerc in the separable case (see also (Krauth & Mezard, 1987; Li et al., 2002; Tsampouka & Shawe-Taylor, 2005)). The value $\beta_{eff} = \beta / ||^{(M+1)}\mathbf{W}||$ is minimal when $||^{(M+1)}\mathbf{W}||$ is maximal. This maximum occurs at $M+1 = M_0$, where $M_0$ is the mistake bound. Using earlier derivations for AMPerc, we obtain $||^{(M+1)}\mathbf{W}|| \leq ||^{(M_0)}\mathbf{W}|| = (2(R^2+\beta)\ M_0)^{1/2} = 2(R^2+\beta)/\gamma$. Therefore, $\beta_{eff} \geq \beta / ||^{(M_0)}\mathbf{W}|| = \beta\gamma / 2(R^2+\beta)$ and $\beta_{eff} / \gamma \geq \beta / 2(R^2+\beta)$, with equality in the worst case.

Further, we observe that $\beta / 2(R^2+\beta) \leq 1/2$, with equality at $\beta >> R^2$. Thus, the number of mistakes required by AMPerc to reach $\beta_{eff} = \gamma/2$ may be much larger than the mistake bound of the standard Perceptron. In contrast, misclassification in MMPerc occurs whenever $\delta(^{(t)}\mathbf{W}, {}^{(t)}\mathbf{x}) \equiv (\langle \mathbf{x}, \mathbf{w}_y \rangle - \langle \mathbf{x}, \mathbf{w}_c \rangle) < \alpha \langle {}^{(t)}\mathbf{x}, {}^{(t)}\mathbf{w}_y \rangle$ (or $|\langle {}^{(t)}\mathbf{x}, {}^{(t)}\mathbf{w}_y \rangle|$). Hence, both the classification outcome and the update/no-update decision depend only on the relative values of the class scores, not on the scaling of $\mathbf{W}$.

Therefore, as the score magnitudes increase during training due to the growing magnitude of the elements of $\mathbf{W}$, the same relative (multiplicative) margin $\alpha$ continues to be enforced. This means that, unlike AMPerc, the number of mistakes in MMPerc does not decrease merely as a result of weight norm growing but depends on how the weights adjust during training to satisfy the specified multiplicative margin threshold condition for the data instances.

A limitation of the MMPerc-Abs margin formulation arises when $\langle \mathbf{x}, \mathbf{w}_y \rangle \geq 0$ and $\langle \mathbf{x}, \mathbf{w}_c \rangle < 0$. In this case, no mistake is ever triggered, regardless of the value of $\alpha$, even as $\alpha$ approaches 1.0. When $\langle \mathbf{x}, \mathbf{w}_y \rangle = 0$, the margin threshold remains zero for any $\alpha$, since $\alpha\langle \mathbf{x}, \mathbf{w}_y \rangle = 0$.

# 4 Experimental Investigation

To evaluate our multiclass linear Perceptron with a multiplicative margin, we first outline the key design considerations and modifications in section 4.1. The datasets and baseline classifiers used in the experiments are introduced in section 4.2. Experimental results are presented in sections 4.3 and 4.4.

### 4.1 Key Design Considerations and Modifications

This section outlines essential design considerations for large-margin Perceptron classifiers, which must be addressed to enable their practical use in experiments.

#### 4.1.1 Bias

Although a bias term is inherent in linear models, it is typically omitted in HDC classifiers. Possible reason is that hypervectors are often implicitly normalized. For example, dense binary hypervectors contain about half 1s, so their squared norm has mean $D/2$ and variance $D/4$ under a binomial model. The resulting Pearson coefficient of variation, $\sigma/\mu = (D/4)^{1/2} / D/2 = 1/\sqrt{D}$, implies concentration around the mean, effectively normalizing them. Moreover, binary hypervectors often maintain a fixed fraction of 1s.

Perceptrons are naturally suited for such bias-free settings, where the decision boundary passes through the origin. However, for more general data, including a bias becomes essential. A standard approach is to augment each data vector with a constant bias component (Mays, 1964), (Duda & Hart, 1973), (Tsampouka & Shawe-Taylor, 2005), allowing the model to operate on homogenous vectors instead of maintaining and updating a separate additive bias term. While this component is typically set to 1, our empirical evaluation found that this choice performed poorly. Instead, setting it proportional to $\sqrt{D}$ notably improved classification accuracy. We attribute this to the fact that unnormalized data vectors have norms up to $R = \Theta(\sqrt{D})$, so substantial shifts of the decision boundary require bias-term updates of order $R^2$. Concretely, the bias-term increment after an update is $(w_{\text{bias}}+R)R - w_{\text{bias}}R = R^2$, compared to 1 when the bias component equals 1.

We have found very few studies that discuss such alternative bias scaling. (Cristianini & Shawe-Taylor, 2000; Li et al., 2002) mention adding $R^2$ when updating the bias term, while (Khardon & Wachman, 2007) use the average squared norm of the data vectors to *initialize* it. Although scaling the bias component $R$ by a constant factor could perform better, we fixed it to $R$ to avoid additional tuning overhead. Note that, for binary vectors, a bias-term update of $R^2$ requires augmenting them with $R^2$ extra 1-components (since $R^2 \leq D$), potentially doubling the dimensionality. In practice, handling the bias component with non-binary arithmetic may therefore be more efficient.

Overall, our experiments confirm that incorporating a bias term can be critical for improving the accuracy of multiclass Perceptrons on certain datasets.

#### 4.1.2 Margin Threshold Selection and Training Settings

The margin threshold $\alpha$ is treated as a hyperparameter and is selected through cross-validation from a predefined set of candidate values. Depending on the dataset, this choice can substantially affect classification accuracy.

A single pass through the training data corresponds to the online learning setting. In contrast, the offline setting presents the training data multiple times, epoch-wise, with evaluation on a separate testing set. Offline training typically yields higher accuracy and is the standard for reporting baseline results of other classifiers; thus, we adopt it. The offline setting further allows the order of training instances to vary between epochs (section 2.1); in our experiments, we apply random shuffling at each epoch. Note that if the online data sequence is constructed by concatenating the per-epoch sequences from offline training, the two procedures produce identical classifiers.

In both online and offline settings, classifier updates need not occur after each mistake but may be accumulated over a batch of instances. However, we use a batch size of 1, consistent with the online setting, and because larger batches did not consistently improve performance in preliminary tests. These tests also showed no benefit from the asymmetric learning rule. Since prior studies consistently employed the symmetric variant, we adopt the symmetric MMPerc-Abs for all reported experiments.

#### 4.1.3 Number of Training Epochs and Stopping Criteria

A basic stopping criterion is to halt training after a fixed number of epochs. Another approach is early stopping once the target training accuracy is reached. Using validation accuracy for stopping is more principled, but it reduces the effective training set and complicates mapping the optimal number of epochs back to the full training set.

In our setup, training stops after a predefined maximum number of epochs, or earlier if training accuracy reaches a specified value, typically 1.0 or close to it, indicating convergence. These values are set for each experiment. For some datasets, convergence is reached within the first few epochs, especially when the margin threshold is small or zero.

#### 4.1.4 Selection of Class Weights for Classifier Evaluation

After training, one must select which set of class weights to use for evaluation. A straightforward approach is to take the final weights from the last training epoch (Kussul et al., 2001; Kussul & Baidyk, 2004; Rachkovskij, 2007, 2022, 2024). However, this can be problematic when determining the optimal number of the training epochs across datasets. Therefore, we instead select the weights from the epoch with highest training accuracy, similar to the pocket algorithm (Gallant, 1990) in the online setting, or to approaches used in recent HDC studies (Smets et al., 2023). If training converges, these weights coincide with the final ones. While this method may not always yield the best testing accuracy (as determined across all epochs), it remains a reasonable approach.

### 4.2 Datasets and Classifiers

#### 4.2.1 Synthetic Datasets from DataGen

Synthetic data provide controlled conditions for experimentation and help build intuition about classifier behaviour – particularly how classification performance is influenced by data dimensionality, number of classes, and training set size. To generate synthetic data, we used DataGen (Rachkovskij & Kussul, 1998). While DataGen offers extensive control over data parameters (e.g., noise types and levels), in this study we used only its basic configuration.

Each data instance is a $d$-dimensional real-valued vector with components in [0,1]. For each class, a single class center is sampled uniformly. Training instances are then generated by sampling each component from a Gaussian distribution centered at the corresponding component of the class center, with a fixed standard deviation. Each class instance is constrained to be closer (in Euclidean distance) to its own class center than to any other. Testing instances are generated in the same way, but from a uniform distribution. No noise is added to the features or class labels in either training or testing sets, producing class regions that are separable by piece-wise linear decision boundaries.

### 4.2.2 Real Datasets

We used the same four publicly available datasets as in (Smets et al., 2023) (see Table 1):

(1) UCIHAR (Anguita et al., 2013) (Dua & Graff, 2019) contains data from acceleration and velocity recordings to recognize the activity of subjects.
https://archive.ics.uci.edu/dataset/240/human+activity+recognition+using+smartphones

(2) CTG (Dua & Graff, 2019) contains cardiotocography data to classify fetal states.
https://archive.ics.uci.edu/dataset/193/cardiotocography

(3) ISOLET (Dua & Graff, 2019) contains speech signal data of each alphabet letter to be classified. https://archive.ics.uci.edu/dataset/54/isolet

(4) HAND (Rahimi, Benatti, et al., 2016) contains electromyography signal data for hand gesture recognition. Five datasets are available: HAND1 through HAND5.
https://iis-people.ee.ethz.ch/~arahimi/papers/ICRC16.pdf

Each dataset was split into 80% training and 20% testing sets. For UCIHAR and ISOLET, these splits were available from the original sources, while for CTG and HAND, we performed the splits ourselves. Additionally, we created five partitions of each training set for 5-fold cross-validation.

Table 1: Real datasets

| dataset | train/test split | train size | test size | classes $K$ | features $d$ |
|---|---|---|---|---|---|
| CTG | No | 1701 | 425 | 3 | 21 |
| UCIHAR | Yes | 7352 | 2947 | 6 | 561 |
| ISOLET | Yes | 6238 | 1559 | 26 | 617 |
| HAND1 | No | 119065 | 29766 | 5 | 4 |
| HAND2 | No | 111308 | 27827 | 5 | 4 |
| HAND3 | No | 110059 | 27514 | 5 | 4 |
| HAND4 | No | 107468 | 26867 | 5 | 4 |
| HAND5 | No | 78504 | 19626 | 5 | 4 |
| HAND(1..5) | No | 526404 | 131600 | 5 | 4 |

### 4.2.3 Data Standardization and Transformation

No standardization was applied to the synthetic DataGen datasets. For real datasets, input data vectors were standardized using z-scoring. Training and testing sets were standardized independently, both during cross-validation and final evaluation.

To evaluate and compare classification accuracy across vector data of varying dimensionalities, we transformed real-valued input vectors into hypervectors. Of the three main approaches for such transformations: compositional, random projection-based, and receptive field-based (Rachkovskij et al., 2013), (Kleyko et al., 2022), we adopted the random projection (RP) approach due to its simplicity and straightforward implementation. Specifically, input vectors were projected to the target dimensionality using multiplication with a random Gaussian matrix. For nonlinear transformation, we applied the simplest scheme: RP followed by component-wise binarization using zero thresholding (Charikar, 2002), see also (Rachkovskij, 2015), (Dasgupta et al., 2018), (Thomas et al., 2023). No further standardization was applied to the resulting hypervectors, whether real-valued or binary.

#### 4.2.4 Classifiers

Our primary focus is to compare the classification accuracy of the standard multiclass Perceptron (Perc) and the MMPerc-Abs variant (hereafter MMPerc for brevity), across different sets of margin threshold values $\alpha$, as described in the respective experiments. To control model capacity, the other two baselines are also restricted to classic linear classifiers that, like Perc and MMPerc, maintain one weight vector per class, but use a one-vs-all setup. These are SVM (MATLAB's built-in implementation with default hyperparameters) and Ridge (scikit-learn's RidgeClassifier with the regularization hyperparameter $\lambda \in \{10^{-8} \cup \text{logspace}(-3, 3, 10)\}$).

### 4.3 Experiments with Synthetic Datasets

In the synthetic dataset experiments, we used $d \in \{2,10\}$ and number of classes $C \in \{3,10\}$. The number of training instances per class varied as $S \in \{5,50,500,5000\}$, with $S_{\text{test}}$ = 500 testing instances per class. Training instances were sampled from Gaussians with standard deviation 0.1. When input vectors were transformed to higher-dimensional hypervectors, the target dimensionality was set to $D$ =1000. Unless otherwise stated, MMPerc classifiers were trained either until convergence (i.e., training accuracy 1.0) or for a maximum of 1000 epochs. For selected configurations, classification accuracies on the testing set are reported in tables for the evaluated classifiers: SVM, Ridge, Perc, and MMPerc (for specific $\alpha$-values indicated in the table headers). The corresponding class regions and decision boundaries are illustrated in associated figures.

### 4.3.1 Results on Synthetic DataGen Datasets

The two-dimensional case ($d = 2$) enables direct visualization of data instances, class regions, and classifier decision boundaries. In the figures below, distinct class regions are shown in different colors. The left panels show decision boundaries produced by SVM, Ridge, Perc, and the best-performing MMPerc variant. The right panels show MMPerc decision boundaries for various $\alpha$-values, with darker shades corresponding to larger $\alpha$.

Results for $d = 2$, $C \in \{3,10\}$, and $S \in \{5,50,500\}$ are summarized in Table 2, Figure 2, and Figure 3. We observe that Perc always perfectly separates the training set; however, its decision boundaries are placed arbitrarily, providing only the minimal separation required. This is particularly evident for small trainings sets (low $S$). In contrast, MMPerc exhibits smooth adjustments of its decision boundaries as $\alpha$ varies, which directly affects classification accuracy. For certain $\alpha$-values, MMPerc consistently achieves higher accuracy than Perc. Meanwhile, SVM and Ridge classifiers occasionally fail to perfectly separate the training set, due to limitations inherent in the one-vs-all classification strategy. Their testing sets accuracies are also lower.

We additionally evaluated other synthetic dataset configurations (Supplementary Note 3.1), including: $d = 10$ and $C \in \{3,10\}$; RP-based expansion from $d \in \{2,10\}$ to $D = 1000$; and assessments of bias and binarization effects.

### 4.3.2 Discussion of Experimental Results

Across the synthetic DataGen datasets, MMPerc almost always outperformed Perc for some value of the margin threshold $\alpha$. Visualizations of decision boundaries further support this, showing that at suitable $\alpha$-values, MMPerc boundaries better align with true class regions. Thus, selecting an appropriate α is critical for realizing MMPerc's benefits. As expected, including a bias term is also important and should be considered carefully when applying the model to new datasets. The accuracy of SVM and Ridge classifiers is generally lower than that of MMPerc, with one exception observed for SVM in the configuration with $C$=3, $d$=10, transformed to $D$ = 1000 via RP, and $S$=5.

In summary, the experiments on synthetic data confirm that the multiclass Perceptron with a multiplicated margin is a compelling alternative to the standard Perceptron, as well as to SVM and Ridge classifiers under one-vs-all setup. With an appropriate choice of the margin threshold hyperparameter and a suitable bias term value (for unnormalized data), MMPerc can yield higher classification accuracy. However, the extent to which these design considerations and conclusions transfer to real datasets remains an open question. To address this, we therefore extend our evaluation of MMPerc to several real datasets, focusing on the impact of margin threshold selection and the use of a bias term.

Table 2: Classification accuracies of SVM, Ridge, Perc, MMPerc (for α-values indicated in the top row). The corresponding class regions and decision boundaries are shown in Figure 2 and Figure 3. $C \in \{3,10\}$, $d$=2.

| *d* | *C* | *S* | SVM | Ridge | Perc | 0.1 | 0.2 | 0.3 | 0.4 | 0.5 | 0.6 | 0.7 | 0.8 | 0.9 | 0.95 |
|---|---|---|---|---|---|---|---|---|---|---|---|---|---|---|---|
| 2 | 3 | 5 | 0.8913 | 0.8660 | 0.7873 | 0.8307 | 0.9013 | 0.9260 | 0.8740 | 0.8927 | **0.9487** | 0.9320 | 0.9320 | 0.9260 | 0.9273 |
| 2 | 3 | 50 | 0.8729 | 0.8487 | 0.9273 | 0.9167 | 0.9327 | 0.9367 | 0.9427 | 0.9347 | **0.9613** | 0.9493 | 0.9407 | 0.9293 | 0.9246 |
| 2 | 3 | 500 | 0.8620 | 0.8347 | 0.9580 | 0.9900 | **0.9933** | 0.9927 | 0.9893 | 0.9807 | 0.9747 | 0.9667 | 0.9560 | 0.9473 | 0.9447 |

| *d* | *C* | *S* | SVM | Ridge | Perc | 0.02 | 0.04 | 0.06 | 0.08 | 0.1 | 0.12 | 0.14 | 0.16 | 0.18 | 0.2 |
|---|---|---|---|---|---|---|---|---|---|---|---|---|---|---|---|
| 2 | 10 | 5 | 0.5326 | 0.5194 | 0.7884 | 0.7874 | 0.7964 | **0.8726** | 0.8396 | 0.7834 | 0.8372 | 0.8378 | 0.7914 | 0.7956 | 0.8108 |
| 2 | 10 | 50 | 0.5306 | 0.5596 | 0.9128 | 0.9216 | 0.9354 | 0.9298 | **0.9532** | 0.9518 | 0.9178 | 0.9344 | 0.9196 | 0.9088 | 0.8934 |
| 2 | 10 | 500 | 0.6266 | 0.5600 | 0.9312 | **0.9662** | 0.9562 | 0.9632 | 0.9614 | 0.9596 | 0.9606 | 0.9534 | 0.9444 | 0.9326 | 0.9156 |

| *d* | *C* | *S* | SVM | Ridge | Perc | 0.1 | 0.2 | 0.3 | 0.4 | 0.5 | 0.6 | 0.7 | 0.8 | 0.9 | 0.95 |
|---|---|---|---|---|---|---|---|---|---|---|---|---|---|---|---|
| 10 | 3 | 5 | 0.8133 | 0.7087 | 0.5207 | 0.5207 | 0.5207 | 0.7287 | 0.7553 | 0.7327 | 0.8067 | 0.8053 | 0.8140 | 0.8027 | **0.8160** |
| 10 | 3 | 50 | 0.8247 | 0.7973 | 0.6567 | 0.6733 | 0.6720 | 0.8380 | 0.8467 | 0.8387 | **0.8647** | 0.8353 | 0.8247 | 0.7860 | 0.7727 |
| 10 | 3 | 500 | 0.8107 | 0.7980 | 0.8260 | 0.8567 | 0.8407 | 0.8380 | 0.8193 | 0.8660 | 0.8760 | **0.9240** | 0.8547 | 0.8260 | 0.8127 |
| 10 | 3 | 5000 | 0.7987 | 0.7960 | 0.8667 | 0.8753 | 0.8787 | 0.8800 | 0.8753 | **0.8953** | 0.8860 | 0.8700 | 0.8500 | 0.8307 | 0.8227 |

| *d* | *C* | *S* | SVM | Ridge | Perc | 0.1 | 0.2 | 0.3 | 0.4 | 0.5 | 0.6 | 0.7 | 0.8 | 0.9 | 0.95 |
|---|---|---|---|---|---|---|---|---|---|---|---|---|---|---|---|
| 10 | 10 | 5 | 0.6770 | 0.5462 | 0.5864 | 0.6136 | 0.5438 | 0.6224 | 0.6258 | **0.7292** | 0.7154 | 0.6992 | 0.6798 | 0.6776 | 0.6374 |
| 10 | 10 | 50 | 0.7344 | 0.5858 | 0.6530 | 0.7394 | 0.6942 | 0.7494 | **0.7660** | 0.7306 | 0.7558 | 0.7628 | 0.7574 | 0.7054 | 0.6660 |
| 10 | 10 | 500 | 0.7266 | 0.5864 | 0.7216 | 0.775 | 0.7858 | 0.773 | 0.7376 | 0.7860 | **0.8070** | 0.7878 | 0.7646 | 0.7012 | 0.6596 |
| 10 | 10 | 5000 | 0.7438 | 0.5972 | 0.8078 | 0.7912 | 0.7814 | 0.8226 | **0.8400** | 0.8208 | 0.8128 | 0.807 | 0.7774 | 0.7154 | 0.6742 |

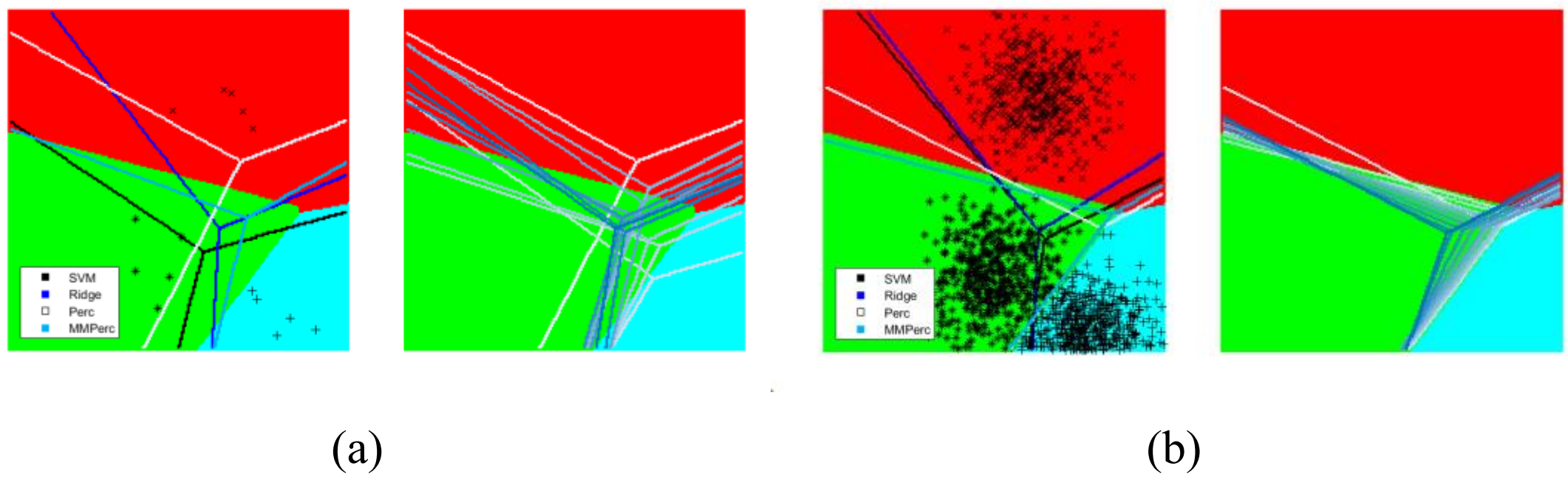

(a) (b)

Figure 2: Class regions (red, green, and blue for the three classes) and decision boundaries of the evaluated classifiers. Left: boundaries produced by SVM, Ridge, Perc, and the best MMPerc. Right: MMPerc boundaries for varying α-values, with darker colors indicating larger α. The corresponding α-values and accuracies are reported in Table 2. $C$=3, $d$=2. (a) $S$=5; (b) $S$=500. α ∈[0.95].

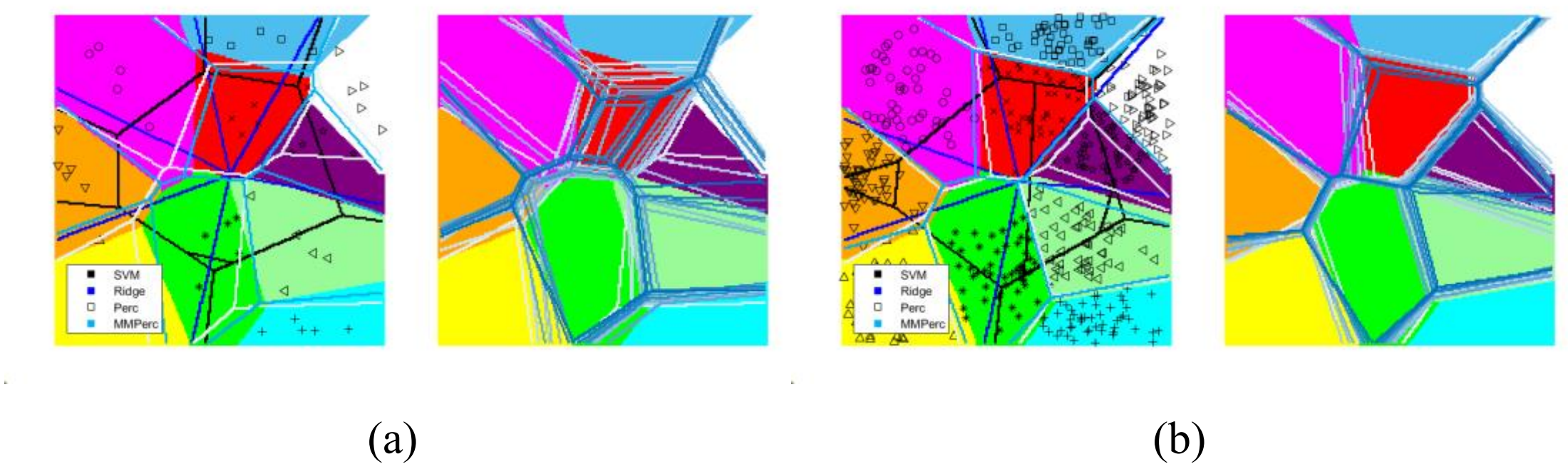

(a) (b)

Figure 3: $C$=10, $d$=2. (a) $S$=5; (b) $S$=50; (c) $S$=500. α ∈[0.2]. Notations and layout are as in Figure 2.

### 4.4 Experiments with Real Datasets

We conducted experiments on both original and binarized versions of real datasets, along with their transformations into real-valued and binary hypervectors (section 4.2.3). Hypervector dimensionalities were chosen from $D$ ∈ {64, 128, 200, 500, 1000, 2000, 4000, 10000}. For MMPerc, α was tuned by cross-validation from one of the following candidate ranges:

α ∈ {0.00 0.01 0.02 0.03 0.04 0.05 0.06 0.07 0.08 0.09 0.1}≡ [0.1];

α ∈ {0.00 0.02 0.05 0.07 0.10 0.12 0.15 0.17 0.20 0.22 0.25 0.27 0.30}≡ [0.3];

α ∈ {0.0 0.1 0.2 0.3 0.4 0.5 0.6 0.7 0.8 0.9} ≡ [0.9].

For Ridge, the regularization hyperparameter $\lambda$ was selected from the range $\lambda \in \{10^{-8} \cup \text{logspace}(-1, 4, 10)\}$. Both MMPerc and Ridge used 5-fold cross-validation; SVM was evaluated with its default hyperparameters.

Training stopped after at most 100 epochs or earlier if the training accuracy reached 0.9999 per epoch, following (Rachkovskij, 2022) and recent studies of HDC classifiers with binary connections (Smets, Rachkovskij, Osipov, Van Leekwijck, et al., 2025; Smets, Rachkovskij, Osipov, Volkov, et al., 2025). This contrasts with 2500-epoch schedule used in (Smets et al., 2023).

We report results for the setups listed in Table 3, where "best" refers to the hyperparameter value yielding the highest testing accuracy (for reference only), and "cv" denotes the value chosen by cross-validation. Here we present results for the CTG dataset; additional real datasets are reported in Supplementary Note 3.2.

Table 3: Classification setups evaluated in the experiments

| | |
|---|---|
| SVM | Support Vector Machine, one-vs-all (Matlab, default) |
| Ridge | Ridge Classifier, one-vs-all (sklearn, best $\lambda$) |
| Ridge cv | Ridge Classifier one-vs-all (sklearn, cross-validated $\lambda$) |
| Perc | Standard multiclass Perceptron |
| Perc bias | Standard multiclass Perceptron with bias |
| MMPerc | Multiclass Perceptron with the multiplicative margin, best $\alpha$ |
| MMPerc cv | Perceptron with the multiplicative margin, cross-validated $\alpha$ |
| MMPerc bias | Multiclass multiplicative margin Perceptron with bias, best $\alpha$ |
| MMPerc bias cv | Multiclass multiplicative margin Perceptron with bias, cross-validated $\alpha$ |

#### 4.4.1 Result Example: CTG

For both the original and binarized versions of the CTG dataset (Figure 4 and Figure 5), MMPerc with the bias term achieves the highest classification accuracy. Notably, the accuracy with cross-validated $\alpha$-values matches that of the best-performing $\alpha$. As expected, binarization reduces accuracy vs the original version. Classification accuracies under RP are averaged over 30 RP matrix realizations. As shown in Figure 6, MMPerc ($\alpha \in [0.9]$) with bias, along with its cross-validated variant, outperforms the baselines. SVM and Perc with bias perform comparably, while bias-free variants of Perc and MMPerc show a noticeable drop in accuracy. Accuracy under RP remains stable across *D*, attributable to the linearity of the RP transformation. An exception is SVM, with accuracy declining at higher *D*, possibly due to suboptimality of its own hyperparameters in overparameterized regimes.

In contrast, in the RP+bin setting (Figure 7), accuracy varies with *D* due to local nonlinear features introduced by binarization. Interestingly, at larger *D*, classifiers trained on binary hypervectors outperform those trained on original or binarized CTG data (Figure 4 and Figure 5). However, this lies outside our focus on comparing the relative accuracy of the evaluated classifiers at each *D*. To highlight this, we compute classifier ranks per *D* and report the average rank (Figure 8). MMPerc (α ∈ [0.1]) achieves the best (lowest) average rank, further improved by including the bias term (bias = $norm_{max}$).

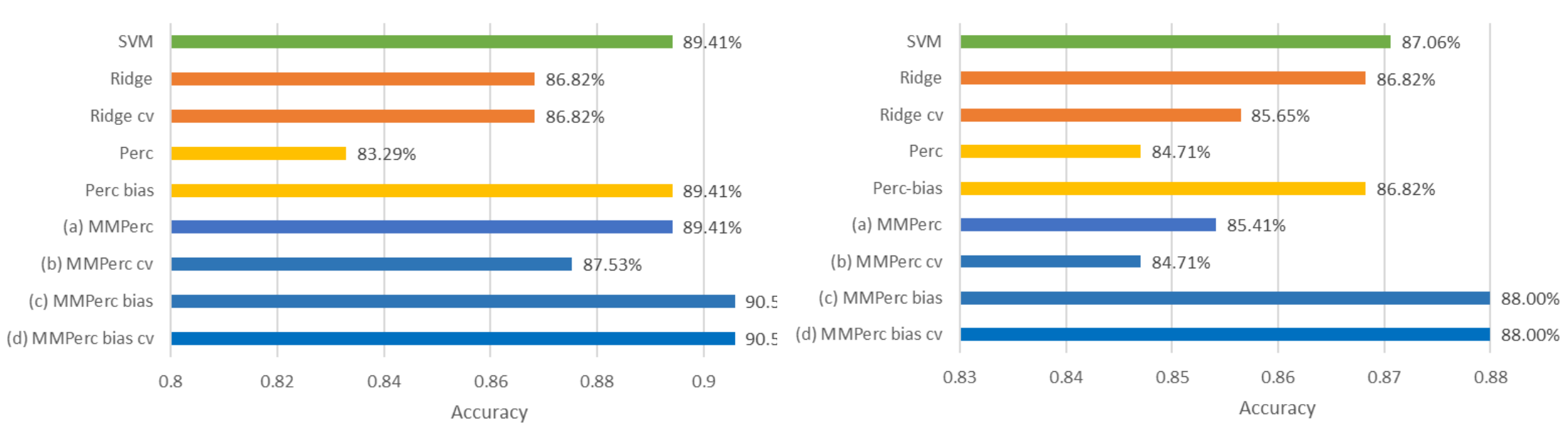

Figure 4: Classification accuracies for CTG orig. (a) α = 0.22; (b) 0.07; (c) 0.22; (d) 0.22. α ∈ [0.3].

Figure 5: Classification accuracies for CTG orig+bin. (a) α = 0.07; (b) 0.0; (c) 0.08; (d) 0.08. α ∈ [0.1].

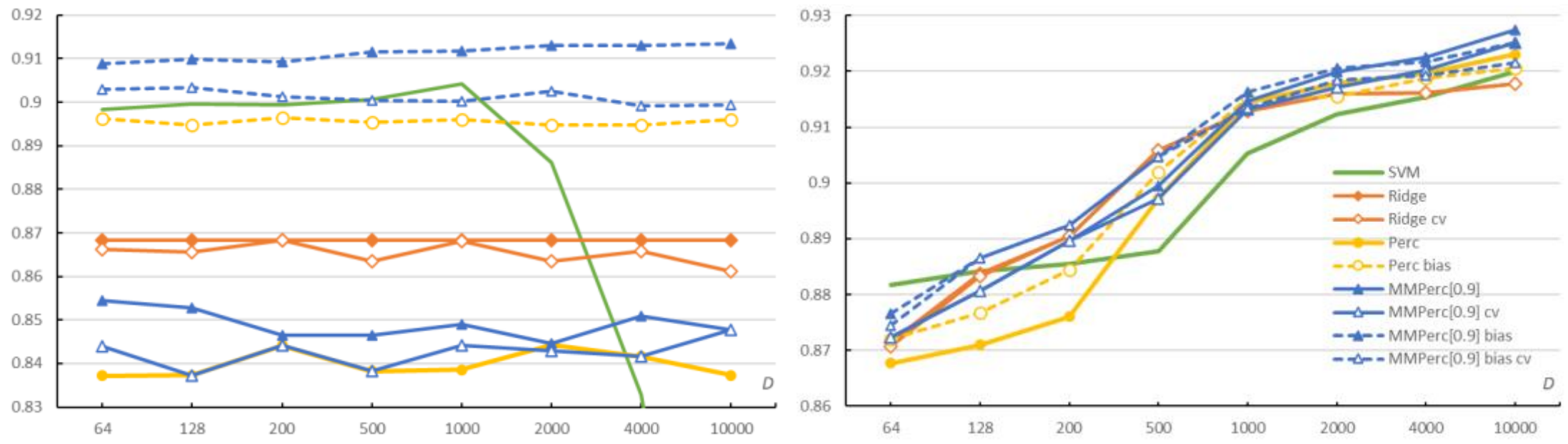

Figure 6: Mean classification accuracies for CTG RP. α ∈ [0.9].

Figure 7: Mean classification accuracies for CTG RP+bin. α ∈ [0.9].

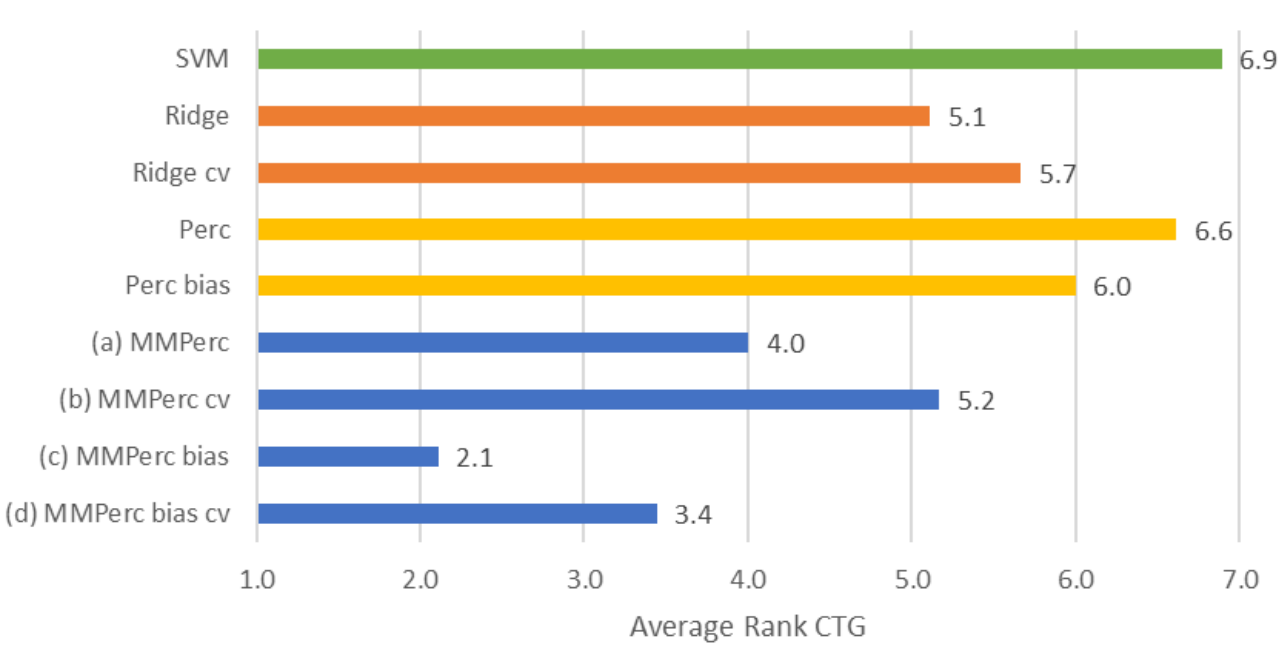

Figure 8: Average rank of classifiers operating on CTG RP+bin hypervectors, computed over $D \in \{64, 128, 200, 500, 1000, 2000, 4000, 10000\}$. A lower average rank indicates better relative performance.

### 4.4.2 Selected Results, Comparison, and Discussion

Table 4 summarizes selected results from our real dataset experiments, along with prior studies using multi-epoch training. Most of those works employ variants of linear Perceptrons within the HDC framework, differing mainly in their learning rules. Only our MMPerc (the MMPerc-Abs variant tested here) and the Confidence Centroid (CCentr) (Smets et al., 2023) use large-margin Perceptrons. Unlike CCentr, which uses binarized class weights, our models operate with real-valued or integer weights.

Direct comparison with prior studies is not straightforward due to differing data formats (original vs hypervectors), hypervector transformations, and evaluation protocols (e.g., best-case vs cross-validated results). Dataset splits and even dataset versions may differ across studies. Nonetheless, Table 4 reports results that appear to be based on equivalent dataset versions. In contrast, our experiments are internally consistent, using identical data vectors (or hypervectors) and dataset splits, so performance differences arise from specified hyperparameters and design choices.

MMPerc achieves classification accuracies close to those of SVM and higher than most other classifiers. Across real datasets and their transformations (binarized, RP, and RP+bin, across a wide range of dimensionalities), several observations emerge. When the margin threshold α is properly selected, MMPerc typically outperforms the standard Perceptron. On the HAND datasets with many training instances, both perform similarly, and cross-validation tends to select $\alpha$-values close to optimal.

Incorporating a bias term typically improves accuracy, particularly for non-binary data. This effect is weaker on binarized RP data, where class separation tends to align with hyperplanes passing through the origin. On UCIHAR, the bias-free variant even provides higher accuracy, though the difference is small.

Across the range of datasets and transformations tested, MMPerc with appropriately chosen $\alpha$-value and bias often outperforms both SVM and Ridge, all operating under the same memory budget of one prototype vector per class. These trends are consistent with the classifier rankings reported in Table 5 (excluding SVM's accuracy drop at higher RP dimensionalities).

In summary, our experiments on both synthetic and real datasets demonstrate that the multiclass Perceptron with a multiplicated margin is a compelling alternative to the standard Perceptron and to classic linear classifiers such as SVM and Ridge.

## 5 Discussion

In this work, we presented a family of multiclass linear Perceptron classifiers based on a multiplicative margin mechanism applied to class scores, as an alternative to traditional additive margin based approaches. We proposed several architectural and algorithmic variants, introducing the corresponding learning rules, and examined key design considerations relevant to their practical deployment. We also developed a basic theoretical analysis, including the formulation of associated loss functions and the derivation of mistake bounds for both linearly separable and non-separable data.

This approach enforces not only correct classification of training instances, but also confidence in those classifications: during training, an instance is considered correctly classified only if the score of the true class exceeds that of the closest competitor by a margin. Unlike traditional additive margins, the multiplicative margin is defined as a fixed proportion of the winning class score. This formulation makes the margin invariant to the absolute magnitude of class scores, eliminating the need to explicitly account for the norms of class prototypes and data vectors when setting the margin threshold value, which is a common issue with additive margins.

Table 4: Selected testing set accuracies (%) on real datasets.

The CTG dataset

| Study | Description | Transformation | Vectors | Weights | *d* / *D* | Acc % |
|---|---|---|---|---|---|---|
| (Smets et al., 2023) | CCentr | Compositional | {0,1} | {0,1} | 10000 | 86.89 |
| (Duan, Xu, et al., 2022) | TinyLDC | Trained Compositional | {0,1} | {0,1} | 64 | 90.50 |
| Ours | SVM | none | real | real | 21 | 89.41 |
| Ours | MMPerc | none | real | real | 21 | 90.59 |
| Ours | SVM | RP with Gauss + bin | {0,1} | integer | 10000 | 91.99 |
| Ours | MMPerc | RP with Gauss + bin | {0,1} | integer | 10000 | 92.51 |

The UCIHAR dataset

| Study | Description | Transformation | Vectors | Weights | *d* / *D* | Acc % |
|---|---|---|---|---|---|---|
| (Smets et al., 2023) | CCentr | Compositional | {0,1} | {0,1} | 10000 | 94.33 |
| (Vergés, Givargis, et al., 2023) | SparseHDr | Compositional | {0,1} | integer | 10000 | 86.98 |
| (Imani, Morris, et al., 2019) | AdaptHD | Compositional | {0,1} | integer | 10000 | 96.20 |
| (Hernández-Cano et al., 2021) | OnlineHD | RP with {-1,+1} | real | real | 10000 | 96.50 |
| (Ponzina & Rosing, 2024) | OnlineHD | Compositional | {–1,+1} | integer | 10000 | 90.40 |
| (Ponzina & Rosing, 2024) | OnlineHD | RP 16 bit | integer | integer | 10000 | 91.31 |
| (Ponzina & Rosing, 2024) | MicroHD | Compositional | {–1,+1} | 3bit | 2000 | 89.48 |
| (Ponzina & Rosing, 2024) | MicroHD | RP 16 bit | integer | 4bit | 200 | 90.33 |
| (Duan, Liu, et al., 2022) | LeHDC | Compositional | {–1,+1} | {–1,+1} | 10000 | 95.23 |
| (Hsiao et al., 2021) | L-HDC | Trained Compositional | {–1,+1} | integer | 3000 | 95.54 |
| (Duan, Xu, et al., 2022) | TinyLDC | Trained Compositional | {0,1} | {0,1} | 128 | 92.56 |
| (Z. Yan et al., 2023) | EffHDC | Trained FCNN | {0,1} | {0,1} | 64 | 94.20 |
| (Z. Yan et al., 2023) | EffHDC | Trained FCNN | {0,1} | {0,1} | 128 | 94.81 |
| Ours | SVM | none | real | real | 561 | 96.26 |
| Ours | MMPerc | none | real | real | 561 | 95.72 |
| Ours | SVM | RP with Gauss + bin | {0,1} | integer | 10000 | 96.12 |
| Ours | MMPerc | RP with Gauss + bin | {0,1} | integer | 10000 | 95.76 |

The ISOLET dataset

| Study | Description | Transformation | Vectors | Weights | *d* / *D* | Acc % |
|---|---|---|---|---|---|---|
| (Smets et al., 2023) | CCentr | Compositional | {0,1} | {0,1} | 10000 | 94.30 |
| (Vergés, Heddes, et al., 2023) | RefineHD | Compositional | {0,1} | real | 10000 | 94.00 |
| (Imani, Morris, et al., 2019) | AdaptHD | Compositional | {0,1} | integer | 10000 | 92.24 |
| (Hernández-Cano et al., 2021) | OnlineHD | RP with {-1,+1} | real | real | 10000 | 94.60 |
| (Ponzina & Rosing, 2024) | OnlineHD | Compositional | {–1,+1} | integer | 10000 | 91.41 |
| (Ponzina & Rosing, 2024) | OnlineHD | RP 16 bit | integer | integer | 10000 | 93.39 |
| (Ponzina & Rosing, 2024) | MicroHD | Compositional | {–1,+1} | 3/16bit | 2000 | 90.45 |
| (Ponzina & Rosing, 2024) | MicroHD | RP 16 bit | integer | 4/16 bit | 200 | 92.51 |
| (Duan, Liu, et al., 2022) | LeHDC | Compositional | {–1,+1} | {–1,+1} | 10000 | 94.89 |
| (Hsiao et al., 2021) | L-HDC | Trained Compositional | {–1,+1} | integer | 3000 | 92.44 |
| (Duan, Xu, et al., 2022) | TinyLDC | Trained Compositional | {0,1} | {0,1} | 128 | 91.33 |
| (Z. Yan et al., 2023) | EffHDC | Trained FCNN | {0,1} | {0,1} | 64 | 91.40 |
| (Z. Yan et al., 2023) | EffHDC | Trained FCNN | {0,1} | {0,1} | 128 | 93.20 |
| Ours | SVM | none | real | real | 617 | 94.36 |
| Ours | MMPerc | none | real | real | 617 | 95.77 |
| Ours | SVM | RP with Gauss + bin | {0,1} | integer | 10000 | 95.88 |
| Ours | MMPerc | RP with Gauss + bin | {0,1} | integer | 10000 | 96.04 |

The HAND datasets

| Study | Description | Transformation | Vectors | Weights | *d* / *D* | Acc % |
|---|---|---|---|---|---|---|
| (Smets et al., 2023) HAND | CCentr iter | Compositional | {0,1} | {0,1} | 10000 | 96.31 |
| Ours HAND1 | SVM | none | real | real | 4 | 97.60 |
| Ours HAND1 | MMPerc | none | real | real | 4 | 97.85 |
| Ours HAND1 | SVM | RP with Gauss + bin | {0,1} | integer | 10000 | 99.63 |
| Ours HAND1 | MMPerc | RP with Gauss + bin | {0,1} | integer | 10000 | 99.58 |
| Ours HAND5 | SVM | none | real | real | 4 | 94.41 |
| Ours HAND5 | MMPerc | none | real | real | 4 | 94.62 |
| Ours HAND5 | SVM | RP with Gauss + bin | {0,1} | integer | 10000 | 98.77 |
| Ours HAND5 | MMPerc | RP with Gauss + bin | {0,1} | integer | 10000 | 98.73 |

Table 5: Classifier ranking on real datasets. Lower ranks indicate better performance

| **orig** | CTG | UCIHAR | ISOLET | HAND1 | HAND5 | Mean |
|---|---|---|---|---|---|---|
| SVM | 3 | 2 | 5 | 1 | 1 | 2.4 |
| Ridge | 7 | 1 | 8 | 8 | 5 | 5.8 |
| Ridge cv | 7 | 2 | 9 | 8 | 5 | 6.2 |
| Perc | 9 | 7 | 5 | 5 | 7 | 6.6 |
| Perc bias | 3 | 9 | 6 | 1 | 1 | 4.0 |
| MMPerc | 3 | 4 | 1 | 5 | 8 | 4.2 |
| MMPerc cv | 6 | 4 | 1 | 5 | 8 | 4.8 |
| MMPerc bias | 1 | 6 | 1 | 1 | 1 | 2.0 |
| MMPerc bias cv | 1 | 8 | 1 | 1 | 1 | 2.4 |
| **orig+bin** | CTG | UCIHAR | ISOLET | HAND1 | HAND5 | |
| SVM | 3 | 6 | 7 | 6 | 9 | 6.2 |
| Ridge | 3 | 7 | 8 | 4 | 1 | 4.6 |
| Ridge cv | 6 | 7 | 8 | 4 | 1 | 5.2 |
| Perc | 8 | 6 | 5 | 6 | 6 | 6.2 |
| Perc bias | 3 | 3 | 6 | 1 | 1 | 2.8 |
| MMPerc | 6 | 1 | 1 | 6 | 6 | 4.0 |
| MMPerc cv | 8 | 3 | 3 | 6 | 6 | 5.2 |
| MMPerc bias | 1 | 1 | 1 | 1 | 1 | 1.0 |
| MMPerc bias cv | 1 | 3 | 3 | 1 | 1 | 1.8 |
| **RP** | CTG | UCIHAR | ISOLET | HAND1 | HAND5 | |
| SVM | 2 | 3 | 7 | 4 | 4 | 4.0 |
| Ridge | 5 | 1 | 7 | 8 | 5 | 5.2 |
| Ridge cv | 6 | 1 | 7 | 8 | 5 | 5.4 |
| Perc | 9 | 6 | 6 | 7 | 9 | 7.4 |
| Perc bias | 4 | 9 | 5 | 1 | 3 | 4.4 |
| MMPerc | 7 | 4 | 1 | 5 | 7 | 4.8 |
| MMPerc cv | 8 | 4 | 1 | 5 | 8 | 5.2 |
| MMPerc bias | 1 | 6 | 1 | 1 | 1 | 2.0 |
| MMPerc bias cv | 2 | 8 | 1 | 1 | 1 | 2.6 |
| **RP+bin** | CTG | UCIHAR | ISOLET | HAND1 | HAND5 | |
| SVM | 6.9 | 6.0 | 6.4 | 3.8 | 4.1 | 5.4 |
| Ridge | 5.1 | 4.5 | 7.4 | 8.2 | 7.8 | 6.6 |
| Ridge cv | 5.7 | 5.4 | 7.8 | 8.8 | 8.6 | 7.2 |
| Perc | 6.6 | 8.6 | 6.4 | 4.9 | 5.2 | 6.3 |
| Perc bias | 6.0 | 7.6 | 6.9 | 6.3 | 6.4 | 6.6 |
| MMPerc | 4.0 | 2.2 | 3.1 | 2.0 | 2.3 | 2.7 |
| MMPerc cv | 5.2 | 4.6 | 3.1 | 2.4 | 2.5 | 3.5 |
| MMPerc bias | 2.1 | 1.8 | 1.9 | 4.2 | 4.1 | 2.8 |
| MMPerc bias cv | 3.4 | 4.4 | 1.9 | 4.4 | 4.1 | 3.7 |
| **Overall mean** | CTG | UCIHAR | ISOLET | HAND1 | HAND5 | |
| SVM | 3.7 | 4.3 | 6.4 | 3.7 | 4.5 | **4.5** |
| Ridge | 5.0 | 3.4 | 7.6 | 7.1 | 4.7 | **5.6** |
| Ridge cv | 6.2 | 3.8 | 8.0 | 7.2 | 4.9 | **6.0** |
| Perc | 8.2 | 6.9 | 5.6 | 5.7 | 6.8 | **6.6** |
| Perc bias | 4.0 | 7.1 | 6.0 | 2.3 | 2.9 | **4.5** |
| MMPerc | 5.0 | 2.8 | 1.5 | 4.5 | 5.8 | **3.9** |
| MMPerc cv | 6.8 | 3.9 | 2.0 | 4.6 | 6.1 | **4.7** |
| MMPerc bias | 1.3 | 3.7 | 1.2 | 1.8 | 1.8 | **2.0** |
| MMPerc bias cv | 1.9 | 5.9 | 1.7 | 1.8 | 1.8 | **2.6** |

For a representative variant of our large-margin Perceptron, we conducted extensive experiments on both synthetic and real datasets, including real-valued and binary data vectors across a wide range of dimensionalities. We evaluated and compared classification performance against the standard Perceptron, Ridge classifier, and Support Vector Machine classifiers, all under the same memory budget. Across the datasets, the multiplicative margin Perceptron achieved competitive accuracy and frequently outperformed the baselines when the bias and margin threshold were properly chosen. A noted limitation shared with other large-margin classifiers, but absent in the standard Perceptron, is the need to select hyperparameters; in this case, the margin threshold, which we addressed through cross-validation.

Compared to other large-margin and margin-free Perceptrons, the variants considered in this work are distinguished by their minimalistic design and simplicity, both in terms of computational complexity and operational requirements.

From a computational perspective, our models avoid several complexities common in alternative methods. They do not employ variable learning rates, weight decay, weight normalization, or weight projection (whether after each mistake or even after correct classification), all of which add overhead. Instead, we adopt the fixed unit learning rate of the standard Perceptron, thereby avoiding floating-point operations when processing integer/binary data vectors. Weight projection or normalization is not employed. Furthermore, training follows the conservative or ultraconservative paradigm (Crammer & Singer, 2003), where weight updates occur only upon mistakes and are restricted to the true and returned class prototypes (see also “Multiclass weight update” in section 5.1.3).

From an operational perspective, our models are well suited not only to floating-point data but even more so to integer and binary vectors, further reducing implementation cost. For integer-valued data, the algorithms can be implemented primarily with integer arithmetic; while for binary data, vector-matrix operations reduce to simple counting or addition. The approach is likewise efficient for sparse data vectors.

## 5.1 Future Research Directions

As this is the first work to systematically investigate multiclass linear Perceptron classifiers with a multiplicative margin, numerous avenues for future research arise. Below we outline several promising directions.

#### 5.1.1 Theoretical Extensions

Further theoretical development of the proposed framework is a natural next step. This includes formal analysis of margin threshold selection, derivation of generalization bounds, and extension of the analysis to settings with quantized weights.

A particularly compelling line of work is to explore connections with recent theoretical advances that aim to explain surprising phenomena in modern machine learning with overparameterized models, including linear ones: implicit regularization bias (Soudry et al., 2018), (Schliserman & Koren, 2022), (Ravi et al., 2024); benign overfitting (K. Wang et al., 2023), (Wu & Sahai, 2024); double descent (Aubin et al., 2020), (Deng et al., 2022), (Lee & Cherkassky, 2024); and grokking (Levi et al., 2024), (Lyu et al., 2024). The HDC framework, with its inherently high-dimensional representations, naturally introduces overparameterization and provides a natural testbed for investigating these effects. Studying how large-margin Perceptrons interact with them could enhance our understanding of both linear models and DNNs.

#### 5.1.2 Empirical Comparison of Large-Margin Perceptron Variants

Another important direction is a comprehensive experimental comparison of different large-margin Perceptron variants. Such work should include evaluation on large and diverse datasets, as well as systematic testing of extensions involving quantized and binary class weights. Such studies would build on and extend previous work, e.g., (Smets et al., 2023; Smets, Rachkovskij, Osipov, Van Leekwijck, et al., 2025; Smets, Rachkovskij, Osipov, Volkov, et al., 2025).

A complementary line of experimentation is a systematic comparison with nonlinear baselines, such as Random Forests, evaluated on both original input features and hypervector representations, see also (Fernández-Delgado et al., 2014), (Pale et al., 2022), (Jeong et al., 2025).

#### 5.1.3 Key Design Considerations

**Two-class classification.** Although this study focuses on multiclass classification, extending the proposed methods to the two-class setting is important for practical applications.

**Margin threshold selection** remains a key challenge. We used cross-validation over a predefined candidate set of α-values, which may need to be tailored per dataset. For linearly separable datasets, the largest α that still ensures separability can be found via binary search over the candidate range. Another practical approach could be to estimate α using lower-dimensional hypervectors, where training is faster, and then refine the candidate range for higher dimensions. This is supported by the relative nature of the multiplicative margin and by our empirical observations that best-performing α-values tend to remain stable across nearby dimensionalities. Relatedly, α may be estimated from early-epoch performance. Additionally, training time could be reduced by initializing training at higher α-values with weights obtained at lower α-values, potentially reducing the number of required epochs. Furthermore, (Smets et al., 2023) proposed guiding α selection by analyzing the empirical margin distribution of a margin-free classifier; a practical implementation of this idea would be valuable.

The current formulation uses single α shared across all classes. A challenge is to support class-specific α-values without incurring a significant increase in computational complexity of margin threshold selection.

Unlike additive margins, the effective multiplicative margin threshold does not decrease during training (section 3.4.2), which may hinder decision boundary stabilization in later stages. One way to address this is to reduce α during training; however, this introduces additional hyperparameters to control the annealing schedule. On the other hand, the naturally decreasing effective learning rate (section 3.4.2) stabilizes training but also slows it down. If needed, scaling down the weights after a number of epochs could be used to restore a higher effective learning rate.

Finally, an implementation-related consideration is to increase the number of cross-validation folds to improve the stability and consistency of α-value selection.

**Multiclass weight update.** Another direction is exploring weight update strategies in which not only the competitor class but all classes violating the margin threshold condition are updated. For additive margins, this includes all classes *i* with $s_i > s_y - \beta$; for a multiplicative margins, those with $s_i > s_y - \alpha|s_y|$. Such updates were considered for additive margins in, e.g., (Duda & Fossum, 1966) and for multiplicative margins in (Rachkovskij, 2007); see also the ultraconservative setting in (Crammer & Singer, 2003). Even more aggressive updates occur in softmax-based classifiers with cross-entropy loss, where the gradient for each class weight vector is $\nabla \mathbf{w}_i L = -(1\{i = y\} - \mathrm{softmax}(s_i))\, \mathbf{x}$, so that all class weights are updated since $1\{i = y\} - \mathrm{softmax}(s_i)$ is generally nonzero. However, such multiclass updates are computationally expensive, especially with many classes, and typically require floating-point operations. In contrast, the two-class update used in this study is more efficient and better suited to resource-constrained and online settings. Still, further investigation of multiclass update rules remains promising, with potential to speed up training and improve generalization in specific scenarios.

**Weight decay** has been shown to approximate SVM solutions via stochastic subgradient descent (Shalev-Shwartz et al., 2007; Z. Wang et al., 2010), see also section 1.3. However, it introduces additional computational overhead: floating-point multiplications to compute the decay term, additions to apply it, and weight projections requiring norm computations. Research into lightweight integration of weight decay into large-margin Perceptrons is therefore a promising direction.

**Selecting weights for classifier evaluation.** In preliminary experiments, selecting Perceptron weights from the epoch with the highest training accuracy often yielded reasonable testing accuracy, but sometimes underperformed compared to the weights from the best epoch in hindsight. To address this gap (when it arises), it would be useful to identify dataset properties correlated with such behavior and to explore alternative strategies for weight selection. One direction is online-to-batch conversion techniques (Littlestone, 1989), (Cesa-Bianchi et al., 2004), (Shalev-Shwartz & Ben-David, 2014), (Mohri et al., 2018) which aggregate multiple hypotheses generated during training, e.g., via averaging or stability-based selection, see also (Gallant, 1990).

**Restricting weights to be non-negative**, as in the original multiplicative margin Perceptrons (Kussul et al., 2001; Kussul & Baidyk, 2004), helps mitigate issues with enforcing a margin between true and competitor class scores of small but opposite signs (see also section 3.4.2). These issues are fully avoided when input vector components are also non-negative. Moreover, non-negative weights support more interpretable models, especially when input vector components represent the presence or importance of features, where negative weights are ambiguous or counterintuitive, see also (Magri, 2015).

A further direction is exploring multiplicative margins together with **multiplicative weight updates**, exemplified by Cristianini et al. (1999) and Sha et al. (2003) in the dual/kernel setting for large-margin classification and by Littlestone (1987) in the primal and margin-free formulation. Unlike standard additive updates, which increment or decrement class weights, multiplicative updates rescale them. This naturally enforces non-negativity and can induce sparsity – properties desirable in certain applications. To our knowledge, such updates have not yet been explored in conjunction with multiplicative margin formulations. Investigating their interaction could lead to novel large-margin Perceptron variants.

Another promising avenue is to **combine large-margin Perceptron training with complementary enhancements** developed for Perceptron-like classifiers, particularly within the HDC framework, e.g., (Imani, Morris, et al., 2019), (Hernández-Cano et al., 2021), (Hsiao et al., 2021), (Ponzina & Rosing, 2024), (Vergés et al., 2025). Many of these enhancements are orthogonal to margin-related design considerations, making them natural candidates for integration. Such combinations may further improve classification accuracy, robustness, and computational efficiency.

#### 5.1.4 IoT, TinyML, Edge Computing, EdgeAI, AIoT

The large-margin Perceptron classifiers explored in this work are extremely simple yet show strong potential for deployment in resource-constrained settings such as IoT, TinyML, Edge Computing, EdgeAI and AIoT (Xu et al., 2024). Their minimalist design leads to low computational and memory requirements for both training and inference, making them well-suited for embedded and edge devices. The large-margin mechanism also improves robustness and generalization, supporting high classification accuracy even under limited-resource conditions. This approach can also help mitigate certain limitations of cloud-based architectures, which require data transmission to remote servers for processing, incurring latency, high communication energy costs, and privacy concerns (Basaklar et al., 2021).

Our methods also extend naturally to Perceptron models with quantized class weights, including binary weights, which is a configuration particularly relevant in the HDC context (Duan, Liu, et al., 2022), (Duan, Xu, et al., 2022), (Z. Yan et al., 2023), (Smets et al., 2023), (Smets, Rachkovskij, Osipov, Volkov, et al., 2025), (Smets, Rachkovskij, Osipov, Van Leekwijck, et al., 2025).

### 5.1.5 Other Tasks and Applications

**Alternative learning tasks and settings.** Future work could apply our large-margin Perceptrons to broader scenarios beyond standard multiclass classification, such as classification with partial or weak labels, multi-label classification, bandit feedback settings, selective classification or abstention-based scenarios, and integration with DNNs, among others.

**Confidence in prediction.** Another promising direction is to explicitly exploit classifier confidence at inference time, for example by abstaining (refusing to classify) when confidence is low. This approach has been studied, for example, in (Chuang et al., 2020), (Mao et al., 2024), see also (Chen et al., 2024), (Yi et al., 2025). Large-margin classifiers are naturally suited to such confidence-aware prediction, since the score margin provides a direct measure of prediction confidence. While in this work the margin is used during training, it could also guide abstention decisions at inference.

**Classification of other data types.** HDC provides principled methods for constructing similarity-preserving hypervectors from a variety of data types. By employing these representations within large-margin Perceptrons, classification can be naturally extended to complex structured data such as sequences, trees, and graphs.

# 6 Conclusion

This work presented and analyzed a family of multiclass linear Perceptron classifiers that incorporate a multiplicative margin mechanism. We developed their architectural formulations and associated learning algorithms, and provided a basic theoretical framework encompassing loss functions, weight update rules, and mistake bounds for both linearly separable and non-separable data. Compared to the existing large-margin methods, our models require no modifications or only minor ones to the standard Perceptron update rule with a fixed unit learning rate. This simplicity makes them computationally lightweight and adaptable to a variety of scenarios involving real-valued, quantized, or binary vector data, and applicable across diverse dimensionalities and dataset sizes.

We investigated key design considerations relevant to the practical deployment of our large-margin Perceptrons, including the role of the bias term, margin threshold selection, training regimes, epoch control, and the choice of classifier weights for inference. Extensive empirical evaluations were conducted on both synthetic and real datasets, using real-valued and binary input vector across a wide range of dimensionalities. Our results demonstrate that the MMPerc variant used in the experiments typically achieves higher classification accuracy than the standard Perceptron, as well as SVM and Ridge classifiers. This improvement, however, comes at the cost of selecting an appropriate margin threshold. In the absence of prior knowledge, this can be addressed via cross-validation, albeit at additional computational cost compared to the standard Perceptron. Nonetheless, the accuracy gains and implementation simplicity make multiplicative margin Perceptrons promising candidates for broader adoption in relevant resource-constrained and edge-oriented scenarios, in HDC contexts, and in linear evaluation of DNNs.

Finally, this study highlights promising directions for future research, including broader theoretical analysis and empirical comparisons, extended design considerations, and exploration of alternative tasks and applications.

## Acknowledgments

The work of D.R. was supported in part by the Swedish Foundation for Strategic Research (SSF, grant nos. UKR22-0024, UKR24-0014), the National Research Fund of Ukraine (NRFU, grant no. 2023.04/0082), and LTU support grant. The work of E.O. and D.R. was supported in part by the Swedish Research Council (VR grant no. 2022-04657). The works of E.O. and D.S. were supported in part by STINT, the Swedish Foundation for International Cooperation in Research and Education (STINT, grant no. MG2020-8842) and Intel Neuromorphic Research Community Project: Unsupervised learning in NLP tasks on using vector-symbolic representations on phasor-based associative memory.

The work of D.S. was supported in part by the Department of Climate Change, Energy, the Environment and Water of Australian Federal Government, as part of the International Clean Innovation Researcher Networks (ICIRN) Program, under grant ICIRN000077.

D.K. acknowledges funding from the Swedish Strategic Research Foundation under the Future Research Leaders program (grant no. FFL24-0111) and the Swedish

Research Council under the Starting Grant program (grant no. 2025-05421). This work was supported in part by the AFOSR under award number FA8655-25-1-7007.

E.O. and D.R. acknowledge the computational resources provided by the National Academic Infrastructure for Supercomputing in Sweden, partially funded by the Swedish Research Council through grant agreement no. 2022-06725, as well as help and support of Tim Ufer and Philip Gard.

# 1 Supplementary Note 1: Related work

## 1.1 The Perceptrons of Rosenblatt

One of the earliest yet still influential paradigms in machine learning and neural networks is the Perceptron, originally proposed and implemented both as computer programs and hardware by Frank Rosenblatt in the 1950s (Rosenblatt, 1957), (Hay et al., 1960), (Rosenblatt, 1962), (Mays, 1964), (Duda & Hart, 1973). The Perceptron was composed of neural-like elements inspired by the McCulloch-Pitts logical threshold units (McCulloch & Pitts, 1943). Each element received multiple inputs, weighted by connection strengths, and produced a single output, typically a binary value, by applying a threshold to the weighted sum of input signals.

Rosenblatt, together with his research group and followers, developed various Perceptron variants, primarily aimed at pattern classification and regression tasks. These efforts led to more advanced designs, including multi-layer Perceptrons (now commonly known as feedforward neural networks and a foundational component of modern DNNs) and Perceptrons with feedback connections, intended to handle more complex computational problems and dynamic systems.

In Rosenblatt's version of the multi-layer Perceptron, the connection weights of the nonlinear units (neuron-like elements) in the initial layers were non-adaptive and often randomly assigned. The function of these initial layers was to perform a fixed, randomized, nonlinear transformation of the input data vectors into a higher-dimensional secondary feature space. In this transformed space, the target task such as classification could then be effectively addressed by a model that is linear in its parameters. For instance, the alpha Perceptron, exemplified by the hardware-implemented Mark I (Hay et al., 1960), comprised three layers: sensory units, association units, and response units. The sensory units were randomly connected to the association units via fixed weights, while the association units were connected to the response units through adjustable weights.

This linear model, implemented as the single output layer of the multilayer Perceptron, consists of parameters corresponding to the weights of the neurons in the final layer. These weights are adjusted during training to solve the target problem. Thus, the fixed Perceptron layers that perform input transformations can be naturally separated from the final, trainable output linear layer, or omitted entirely in inherently linear problems where they act as identity transformations. This output layer alone forms the simplest variant of the Perceptron: the single-layer Perceptron, or Rosenblatt's Perceptron, which includes only one layer of trainable connection weights, forming a model that is linear in its parameters.

Various versions of simple, incremental online training procedures based on error correction were proposed for Rosenblatt's Perceptrons. Many of these methods were proven to converge to a set of weights that correctly classify all vector data instances from linearly separable classes.

Over time, Rosenblatt's original framework has been reinvented and extended in multiple directions. Within the scope of this paper, multiclass and large-margin Perceptron constructions are most relevant.

### 1.2 Multiclass Linear Perceptrons

The multiclass linear Perceptron classifier (without margin) was first introduced in a non-adaptive setting by (Eldredge et al., 1956), with adaptive versions appearing later in (Roberts, 1960), (Uhr & Vossler, 1961). This model was referred to as a "linear machine" in (Nilsson, 1965a), (Nilsson, 1965b), (Duda & Hart, 1973).

As noted in (Nilsson, 1965a), (Nilsson, 1965b), (Duda & Hart, 1973), Carl Kesler proposed an equivalent representation of this model as a two-class classifier by concatenating the class weight vectors, see also (Har-Peled et al., 2002), (Crammer & Singer, 2003). This "Kesler construction" enables the use of an error-correcting learning rule and allows the application of two-class Perceptron analysis. It was used establish convergence guarantees in the linear separable case in (Nilsson, 1965a).

The explicit (linear machine) multiclass linear Perceptron classifier is treated in, e.g., (Nilsson, 1965a), (Nilsson, 1965b), (Duda & Hart, 1973). Mistake bounds for the linearly separable case are presented in, e.g., (Crammer & Singer, 2003), (Beygelzimer et al., 2019). This multiclass construction remains widely used and continues to be analyzed under various loss functions, settings, and tasks. Recent examples include (Beygelzimer et al., 2017), (Soudry et al., 2018), (Beygelzimer et al., 2019), (K. Wang et al., 2023), (Ravi et al., 2024), (Wu & Sahai, 2024), among others.

## 1.3 Large Margin Perceptrons

Perceptrons with an additive margin were considered as early as (Mays, 1964) in the two-class case, and in the multiclass setting by (Nilsson, 1965b), (Duda & Fossum, 1966). Since then, many variants of Perceptron classifiers with additive margins have been proposed, exemplified by the following methods.

The two-class MinOver algorithm (Krauth & Mezard, 1987) updates only single instance with the minimum score in each epoch, provided that this score falls below the margin threshold β. PAUM (Li et al., 2002) is an online two-class classifier that introduces an asymmetric margin to handle class imbalance. ROMMA (Li & Long, 2002) is another online two-class algorithm that, upon a mistake, updates the classifier to find a minimal-norm weight vector satisfying two linear constraints: to classify previous instances well, and to correctly classify the current instance. ROMMA detects mistakes using a zero margin threshold, while aggressive ROMMA (Li & Long, 2002) employs a margin of one. A multiclass extension of ROMMA is presented in (Crammer & Singer, 2003).

The Passive-Aggressive algorithm (Crammer et al., 2006) is an online multiclass method that updates the model only when the margin falls below one. The update is computed by solving a constrained optimization problem that balances enforcing a margin of at least 1 on the current instance with remaining close to the current weights. MIRA (Crammer & Singer, 2003) is another online multiclass algorithm that a β-margin threshold to detect mistakes. Its update rule is based on a constrained optimization problem which coincides with Passive-Aggressive in the separable two-class case. Both ALMA (Gentile, 2001) and MICRA (Tsampouka & Shawe-Taylor, 2007) constrain the norm of the weight vector and adjust both the learning rate and the margin threshold as training proceeds.

The Support Vector Machine (SVM) (Vapnik & Chervonenkis, 1964), (Vapnik & Chervonenkis, 1974), (Cervantes et al., 2020) is a principled approach for large-margin classification. Its loss function combines the squared norm of the weight vector with the 1-hinge loss. Notably, replacing the latter with the squared error loss yields Tikhonov regularization (Ridge regression).

Pegasos (Shalev-Shwartz et al., 2007), (Shalev-Shwartz et al., 2010) aims to approximate SVM in the two-class offline and online settings. Stochastic subgradient descent is performed on randomly sampled training instance. If the margin is less than one (i.e., a mistake is made), a Perceptron-style update with weight decay is applied; otherwise, only weight decay is used. An optional step is to limit the norm of the weight vector. A multiclass extension of Pegasos was proposed in (Z. Wang et al., 2010).

Various incremental two-class SVMs are reviewed in (Lawal, 2019). A lesser-known geometric approach for iteratively finding the optimal separating hyperplane between two classes is that of Schlesinger-Kozinec (Kozinets, 1964), (Schlesinger & Hlaváč, 2002), (Franc & Hlaváč, 2003), (Fainzilberg & Matushevych, 2018), (Malozemov et al., 2025). The target hyperplane is defined to be perpendicular to the midpoint of the shortest line segment connecting two points, each belonging to the convex hull of the training instances of the respective classes. The approximate method described in (Schlesinger & Hlaváč, 2002), (Franc & Hlaváč, 2003) maintains a prototype weight vector for each of the two classes. At each epoch, it identifies the training instance closest to the current separating hyperplane and updates the corresponding class prototype using this instance along with weight decay. The learning rate is computed adaptively based on the current weight vector and the selected instance. Training proceeds until a well-defined, sophisticated stopping criterion is met.

In summary, all large-margin training methods described in this section require computations that are more complex than those used in the standard Perceptron and the large-margin Perceptron variants considered in this paper. Our models operate by computing the maximum of unnormalized class scores, obtained via dot products and margin-discounted for the true class during training. In the case of a mistake, the unmodified instance vector is added to the true class weight vector and subtracted from the weights of the incorrectly predicted class. Furthermore, some existing large-margin algorithms do not address extensions from the two-class case to the multiclass setting. Notably, multiplicative margin algorithms have not been explored in prior work of others.

### 1.4 Fixed Input Nonlinear Transformation

Linear classifiers built upon fixed, randomized nonlinear transformations have been widely studied and form an important class of neural network architectures. One well-known example is the Random Vector Functional-Link (RVFL) network (Pao &

Takefuji, 1992), (Malik et al., 2023), (M. Kim & Heo, 2024), as well as the related Extreme Learning Machine (ELM) (Huang et al., 2006), (Ortín et al., 2015), (J. Wang et al., 2022), both designed for vector-based input data. For data with sequential structure, similar ideas were developed under the framework of Reservoir Computing (RC) (Schrauwen et al., 2007), (Lukoševičius & Jaeger, 2009), (Tanaka et al., 2019) (M. Yan et al., 2024), which includes the Echo State Networks (ESN) (Jaeger, 2001) and Liquid State Machines (LSM) (Maass et al., 2002). These models often employ non-iterative, closed-form solutions based on the Least Squares Method, sometimes enhanced with Tikhonov regularization (i.e., Ridge regression). However, when such solutions rely on computationally intensive singular value decomposition (SVD), they do not natively support online or incremental training.

It is also worth noting the early research from the 1960s on *potential functions* (Aizerman et al., 1964, 1970). This work operated with the concept of a "linearization space" (also referred to as a "secondary feature" Hilbert space), which anticipates the kernel-based formulations later developed in SVM frameworks.

Biological perspective of Perceptron and expansion into high dimensional space is considered in (Babadi & Sompolinsky, 2014), (Brunel et al., 2025). Complementing analysis of (Babadi & Sompolinsky, 2014), (Frady et al., 2018) developed a theory of sequence indexing and working memory in recurrent networks that directly links echo-state networks to HDC superposition and retrieval. The later extensions in (Kleyko et al., 2024) linked the theory to linear Perceptrons.

## 1.5 HDC Classifiers

As noted in Introduction, HDC applications employ a variety of hypervector formats and dimensionalities. In classification tasks, this necessitates the use of appropriate efficient classifiers – typically linear, since the transformations used to construct hypervectors are already nonlinear.

The simplest linear multiclass HDC classifier is the prototype/centroid classifier, as described in (Kleyko et al., 2015), (Rahimi, Kanerva, et al., 2016). This method builds class prototypes by averaging the training hypervectors per class, with an option to quantize those prototypes. Although computationally efficient, this approach typically yields modest classification accuracy because it lacks a mechanism to correct misclassifications by updating the class prototypes.

Several "prototype refining" methods in the HDC literature effectively implement variants of the Perceptron algorithm. In VoiceHD (Imani et al., 2017), error correction is applied in a single pass over the training data, yielding integer-valued weights. In contrast, (Y. Kim et al., 2018) investigates models where weights are binarized after multiple training epochs. BinHD (Imani, Messerly, et al., 2019) uses binary prototypes for inference but updates underlying non-binary prototypes upon mistakes, with re-binarization after each update. QuantHD (Imani et al., 2020) similarly maintains both quantized and unquantized prototypes and applies quantization at the end of each epoch. This strategy is employed in other HDC studies, such as (Hsiao et al., 2021).

AdaptHD (Imani, Morris, et al., 2019) dynamically adjusts the learning rate based on the score difference between the true and predicted classes; binarization is performed after training. OnlineHD (Hernández-Cano et al., 2021) improves classification accuracy by using non-binary hypervectors and also adapts the learning rate based on prediction confidence. TD-HDC (Chuang et al., 2020) maintains integer prototypes that are updated upon misclassification by either the binary or integer classifier. Final binarized prototypes are obtained after training, and a "classification confidence" is used at inference to determine which classifier to apply.

Notably, none of these HDC classifiers employ a margin during training.

RefineHD (Vergés, Givargis, et al., 2023) introduces a heuristic that uses the score difference between the top two predicted classes to guide updates. The most relevant HDC approaches in the context of our work, beyond those covered in Section 2.4, are (Smets et al., 2023), (Smets, Rachkovskij, Osipov, Van Leekwijck, et al., 2025), and (Smets, Rachkovskij, Osipov, Volkov, et al., 2025). These models do consider margins between class scores during training, but maintain two sets of weights and use the binarized set to compute class predictions.

# 2 Supplementary Note 2: Proofs of Mistake Bounds

## 2.1 Proof of Theorem 1 (Mistake bounds for large margin Perceptrons in the linearly separable case)

The formulation of Theorem 1 is given in section 3.2.1.

**Proof.** Consider the update of the Perceptron weight matrix ${}^{(t)}\mathbf{W}$ (Eq. W), starting from ${}^{(1)}\mathbf{W} = \mathbf{0}$, in the case of a classification mistake at round $t$, i.e., when ${}^{(t)}y^* \neq {}^{(t)}y$. We analyze this update for the various versions of the multiclass Perceptron introduced above.

### 2.1.1 Lower Bound on $||\mathbf{W}||$

Let us obtain a lower bound on $||\mathbf{W}||$ in terms of the number $M$ of mistakes. Consider the Frobenius dot product $\langle \mathbf{U}, \mathbf{W} \rangle$, where $\mathbf{U}$ satisfies (3.11) and (3.12). Suppose a mistake occurs in round $t$.

(1)-(3) For AMPerc, MMPerc, and MMPerc-Abs, using the Perceptron update rule (2.2)-(2.3) and taking γ-margin condition (3.11) into account, we obtain the following inequality:

$\langle \mathbf{U}, {}^{(t+1)}\mathbf{W} \rangle = \langle \mathbf{U}_{\setminus\{y,y^*\}}, {}^{(t)}\mathbf{W}_{\setminus\{y,y^*\}} \rangle + \langle \mathbf{u}_y, {}^{(t)}\mathbf{w}_y + {}^{(t)}\mathbf{x} \rangle + \langle \mathbf{u}_{y^*}, {}^{(t)}\mathbf{w}_{y^*} - {}^{(t)}\mathbf{x} \rangle =$

$\langle \mathbf{U}, {}^{(t)}\mathbf{W} \rangle + \langle \mathbf{u}_y - \mathbf{u}_{y^*}, {}^{(t)}\mathbf{x} \rangle \geq \langle \mathbf{U}, {}^{(t)}\mathbf{W} \rangle + \gamma.$

Since we start from ${}^{(1)}\mathbf{W} = \mathbf{0}$, it follows that after $M$ mistakes $\langle \mathbf{U}, {}^{(M+1)}\mathbf{W} \rangle \geq \gamma M$.

(4) For MMPerc-Asm, using the weight update rule (3.5), we obtain:

$\langle \mathbf{U}, {}^{(t+1)}\mathbf{W} \rangle = \langle \mathbf{U}_{\setminus\{y,y^*\}}, {}^{(t)}\mathbf{W}_{\setminus\{y,y^*\}} \rangle + \langle \mathbf{u}_y, {}^{(t)}\mathbf{w}_y + {}^{(t)}\mathbf{x}\,(1-\alpha) \rangle + \langle \mathbf{u}_{y^*}, {}^{(t)}\mathbf{w}_{y^*} - {}^{(t)}\mathbf{x} \rangle =$

$\langle \mathbf{U}, {}^{(t)}\mathbf{W} \rangle + \langle \mathbf{u}_y - \mathbf{u}_{y^*}, {}^{(t)}\mathbf{x} \rangle - \alpha \langle \mathbf{u}_y, {}^{(t)}\mathbf{x} \rangle.$

Applying the Cauchy–Schwarz inequality and (3.10): $\langle \mathbf{u}_y, {}^{(t)}\mathbf{x} \rangle \leq ||\mathbf{u}_y||\, ||{}^{(t)}\mathbf{x}|| \leq R$. Substituting this bound and using (3.11), we get $\langle \mathbf{U}, {}^{(t+1)}\mathbf{W} \rangle \geq \langle \mathbf{U}, {}^{(t)}\mathbf{W} \rangle + \gamma - \alpha R$. Thus, after $M$ mistakes: $\langle \mathbf{U}, {}^{(M+1)}\mathbf{W} \rangle \geq (\gamma - \alpha R) M$.

(5) For MMPerc-AbsAsm, using the weight update rule (3.9), we obtain:

$\langle \mathbf{U}, {}^{(t+1)}\mathbf{W} \rangle = \langle \mathbf{U}_{\setminus\{y,y^*\}}, {}^{(t)}\mathbf{W}_{\setminus\{y,y^*\}} \rangle + \langle \mathbf{u}_y, {}^{(t)}\mathbf{w}_y + {}^{(t)}\mathbf{x}\,(1-\alpha \text{ sign } \langle {}^{(t)}\mathbf{w}_y, {}^{(t)}\mathbf{x} \rangle) \rangle + \langle \mathbf{u}_{y^*}, {}^{(t)}\mathbf{w}_{y^*} - {}^{(t)}\mathbf{x} \rangle = \langle \mathbf{U}, {}^{(t)}\mathbf{W} \rangle + \langle \mathbf{u}_y - \mathbf{u}_{y^*}, {}^{(t)}\mathbf{x} \rangle - \alpha\, (\text{sign } \langle {}^{(t)}\mathbf{w}_y, {}^{(t)}\mathbf{x} \rangle)\, \langle \mathbf{u}_y, \mathbf{x} \rangle.$

Since $(\text{sign } \langle \mathbf{w}_y, \mathbf{x} \rangle) \langle \mathbf{u}_y, \mathbf{x} \rangle \leq |\langle \mathbf{u}_y, \mathbf{x} \rangle| \leq ||\mathbf{u}_y||\, ||\mathbf{x}|| \leq R$, we get $\langle \mathbf{U}, {}^{(t+1)}\mathbf{W} \rangle \geq \langle \mathbf{U}, {}^{(t)}\mathbf{W} \rangle + \gamma - \alpha R$. Thus, after $M$ mistakes: $\langle \mathbf{U}, {}^{(M+1)}\mathbf{W} \rangle \geq (\gamma - \alpha R) M$, which matches the bound for MMPerc-Asm.

Using the Cauchy–Schwarz inequality and (3.12), $\langle \mathbf{U}, {}^{(M+1)}\mathbf{W}\rangle^2 \leq ||\mathbf{U}||^2\, ||{}^{(M+1)}\mathbf{W}||^2 = ||{}^{(M+1)}\mathbf{W}||^2$. Thus, we get:

(1)-(3) For AMPerc, MMPerc, and MMPerc-Abs: $(\gamma M)^2 \leq ||{}^{(M+1)}\mathbf{W}||^2$.

(4)-(5) For MMPerc-Asm and MMPerc-AbsAsm: $((\gamma - \alpha R)\, M)^2 \leq ||{}^{(M+1)}\mathbf{W}||^2$, with $\gamma - \alpha R \geq 0$.

2.1.2 Upper Bound on $||\mathbf{W}||$

Let us now derive an upper bound on $||\mathbf{W}||$ in terms of the number $M$ of mistakes. In the case of a mistake, using the Perceptron update rule (2.2)-(2.3), which is employed in AMPerc, MMPerc, and MMPerc-Abs, we have:

$||{}^{(t+1)}\mathbf{W}||^2 = ||{}^{(t)}\mathbf{W}_{\backslash\{y,y^*\}}||^2 + ||{}^{(t)}\mathbf{w}_y + {}^{(t)}\mathbf{x}||^2 + ||{}^{(t)}\mathbf{w}_c - {}^{(t)}\mathbf{x}||^2 =$

$||{}^{(t)}\mathbf{W}_{\backslash\{y,y^*\}}||^2 + ||{}^{(t)}\mathbf{w}_y||^2 + ||{}^{(t)}\mathbf{x}||^2 + 2\,\langle {}^{(t)}\mathbf{w}_y, {}^{(t)}\mathbf{x}\rangle + ||{}^{(t)}\mathbf{w}_{y^*}||^2 + ||{}^{(t)}\mathbf{x}||^2 - 2\,\langle {}^{(t)}\mathbf{w}_{y^*}, {}^{(t)}\mathbf{x}\rangle =$

$||{}^{(t)}\mathbf{W}||^2 + 2||{}^{(t)}\mathbf{x}||^2 + 2\,\langle {}^{(t)}\mathbf{w}_y - {}^{(t)}\mathbf{w}_{y^*}, {}^{(t)}\mathbf{x}\rangle$.

Let us note that when $2||{}^{(t)}\mathbf{x}||^2 + 2\,\langle {}^{(t)}\mathbf{w}_y - {}^{(t)}\mathbf{w}_{y^*}, {}^{(t)}\mathbf{x}\rangle < 0$, the norm of the weight matrix decreases after the update: $||{}^{(M+1)}\mathbf{W}|| < ||{}^{(M)}\mathbf{W}||$. However, due to the lower bound of the previous section, we still have $||{}^{(M+1)}\mathbf{W}|| \geq \gamma M$.

Now, we have:

(1) For AMPerc, a mistake occurs when $\langle {}^{(t)}\mathbf{w}_y - {}^{(t)}\mathbf{w}_{y^*}, {}^{(t)}\mathbf{x}\rangle < \beta$. Taking into account that $||{}^{(t)}\mathbf{x}||^2 \leq R^2$ from (3.10), we obtain: $||{}^{(t+1)}\mathbf{W}||^2 < ||{}^{(t)}\mathbf{W}||^2 + 2R^2 + 2\beta$, and after $M$ mistakes, we get: $||{}^{(M+1)}\mathbf{W}||^2 < 2(R^2 + \beta)\, M$.

(2) For MMPerc, a mistake occurs when $\langle {}^{(t)}\mathbf{w}_y - {}^{(t)}\mathbf{w}_{y^*}, {}^{(t)}\mathbf{x}\rangle < \alpha\,\langle {}^{(t)}\mathbf{w}_y, {}^{(t)}\mathbf{x}\rangle$. This gives: $||{}^{(t+1)}\mathbf{W}||^2 < ||{}^{(t)}\mathbf{W}||^2 + 2R^2 + 2\,\alpha\,\langle {}^{(t)}\mathbf{w}_y, {}^{(t)}\mathbf{x}\rangle$. Using the Cauchy-Shwartz inequality: $|\langle {}^{(t)}\mathbf{w}_y, {}^{(t)}\mathbf{x}\rangle| \leq ||{}^{(t)}\mathbf{w}_y||\; ||{}^{(t)}\mathbf{x}||$. By applying a (rather loose) upper bound $||{}^{(t)}\mathbf{w}_y|| \leq R\,(t-1)$, we get $|\langle {}^{(t)}\mathbf{w}_y, {}^{(t)}\mathbf{x}\rangle| \leq ||{}^{(t)}\mathbf{w}_y||\; ||{}^{(t)}\mathbf{x}|| \leq (t-1)R^2$ and $||{}^{(t+1)}\mathbf{W}||^2 < ||{}^{(t)}\mathbf{W}||^2 + 2R^2 + 2\,\alpha\,(t-1)\, R^2$. Thus, after $M$ mistakes, we obtain: $||{}^{(M+1)}\mathbf{W}||^2 < \sum_{t=1,M} 2R^2 + 2\alpha R^2 \sum_{t=1,M} (t-1) = 2R^2 M + 2\alpha R^2 M\,(M-1)/2 = R^2\,(2\,M + \alpha\, M^2 - \alpha\, M)$.

(3) For MMPerc-Abs, a mistake occurs when $\langle {}^{(t)}\mathbf{w}_y - {}^{(t)}\mathbf{w}_{y^*}, {}^{(t)}\mathbf{x}\rangle < \alpha\,|\langle {}^{(t)}\mathbf{w}_y, {}^{(t)}\mathbf{x}\rangle|$. This gives: $||{}^{(t+1)}\mathbf{W}||^2 < ||{}^{(t)}\mathbf{W}||^2 + 2R^2 + 2\,\alpha\,|\langle {}^{(t)}\mathbf{w}_y, {}^{(t)}\mathbf{x}\rangle|$. After $M$ mistakes, similarly to MMPerc, we obtain: $||{}^{(M+1)}\mathbf{W}||^2 < R^2\,(2\,M + \alpha\, M^2 - \alpha\, M)$.

(4) For MMPerc-Asm, using the weight update rule (3.5), we get:

$||{}^{(t+1)}\mathbf{W}||^2 = ||{}^{(t)}\mathbf{W}_{\backslash\{y,y^*\}}||^2 + ||{}^{(t)}\mathbf{w}_y + {}^{(t)}\mathbf{x}(1-\alpha)||^2 + ||{}^{(t)}\mathbf{w}_{y^*} - {}^{(t)}\mathbf{x}||^2 =$

$||^{(t)}\mathbf{W}_{\backslash\{y,y^*\}}||^2 + ||^{(t)}\mathbf{w}_y||^2 + ||^{(t)}\mathbf{x}(1-\alpha)||^2 + 2\ \langle^{(t)}\mathbf{w}_y,^{(t)}\mathbf{x}(1-\alpha)\rangle + ||^{(t)}\mathbf{w}_{y^*}||^2 + ||^{(t)}\mathbf{x}||^2 - 2\ \langle^{(t)}\mathbf{w}_{y^*},^{(t)}\mathbf{x}\rangle$

$= ||^{(t)}\mathbf{W}||^2 + (2 - 2\alpha + \alpha^2)||^{(t)}\mathbf{x}||^2 + 2\ \langle^{(t)}\mathbf{w}_y\ (1 - \alpha) - {}^{(t)}\mathbf{w}_{y^*},^{(t)}\mathbf{x}\rangle =$

$||^{(t)}\mathbf{W}||^2 + (2 - 2\alpha + \alpha^2)||^{(t)}\mathbf{x}||^2 + 2(\ \langle^{(t)}\mathbf{w}_y - {}^{(t)}\mathbf{w}_{y^*},^{(t)}\mathbf{x}\rangle - \alpha\ \langle^{(t)}\mathbf{w}_y\ ,^{(t)}\mathbf{x}\rangle).$

Since a mistake occurs when $\langle^{(t)}\mathbf{w}_y - {}^{(t)}\mathbf{w}_{y^*},^{(t)}\mathbf{x}\rangle < \alpha\ \langle^{(t)}\ \mathbf{w}_y,\ ^{(t)}\ \mathbf{x}\rangle$, we have: $2(\ \langle^{(t)}\mathbf{w}_y\ - {}^{(t)}\mathbf{w}_{y^*},^{(t)}\mathbf{x}\rangle - \alpha\ \langle^{(t)}\mathbf{w}_y\ ,^{(t)}\mathbf{x}\rangle) < 0$. Thus, $||^{(t+1)}\mathbf{W}||^2 < ||^{(t)}\mathbf{W}||^2 + (2 - 2\alpha + \alpha^2)\ R^2$. After $M$ mistakes, we get: $||^{(M+1)}\mathbf{W}||^2 < (2 - 2\alpha + \alpha^2)\ R^2\ M = (1+(1-\alpha)^2)\ R^2\ M$.

(5) For MMPerc-AbsAsm, using the weight update rule (3.9), we get:

$||^{(t+1)}\mathbf{W}||^2 = ||^{(t)}\mathbf{W}_{\backslash\{y,y^*\}}||^2 + ||^{(t)}\mathbf{w}_y + {}^{(t)}\mathbf{x}(1-\alpha\ \text{sign}\ \langle^{(t)}\mathbf{w}_y,^{(t)}\mathbf{x}\rangle)||^2 + ||^{(t)}\mathbf{w}_{y^*} - {}^{(t)}\mathbf{x}||^2 =$

$||^{(t)}\mathbf{W}_{\backslash\{y,y^*\}}||^2 + ||^{(t)}\mathbf{w}_y||^2 + ||^{(t)}\mathbf{x}(1-\alpha\ \text{sign}\ \langle^{(t)}\mathbf{w}_y,^{(t)}\mathbf{x}\rangle)||^2 + 2\ \langle^{(t)}\mathbf{w}_y,^{(t)}\mathbf{x}\ (1-\alpha\ \text{sign}\langle^{(t)}\mathbf{w}_y,^{(t)}\mathbf{x}\rangle)\rangle$

$+ ||^{(t)}\mathbf{w}_{y^*}||^2 + ||^{(t)}\mathbf{x}||^2 - 2\ \langle^{(t)}\mathbf{w}_{y^*},^{(t)}\mathbf{x}\rangle =$

$||^{(t)}\mathbf{W}||^2 + (2 - 2\alpha\ \text{sign}\ \langle^{(t)}\mathbf{w}_y,^{(t)}\mathbf{x}\rangle + \alpha^2)||^{(t)}\mathbf{x}||^2 + 2(\ \langle^{(t)}\mathbf{w}_y\ - {}^{(t)}\mathbf{w}_{y^*},^{(t)}\mathbf{x}\rangle - \alpha\ |\langle^{(t)}\mathbf{w}_y\ ,^{(t)}\mathbf{x}\rangle|).$

Since $2(\ \langle^{(t)}\mathbf{w}_y\ - {}^{(t)}\mathbf{w}_c,^{(t)}\mathbf{x}\rangle - \alpha\ |\langle^{(t)}\mathbf{w}_y\ ,^{(t)}\mathbf{x}\rangle|) < 0$ and $-\ 2\alpha\ \text{sign}\ \langle^{(t)}\mathbf{w}_y,\ ^{(t)}\mathbf{x}\rangle \leq 2\alpha$, we get $||^{(t+1)}\mathbf{W}||^2 < ||^{(t)}\mathbf{W}||^2 + (2 + 2\alpha + \alpha^2)\ ||^{(t)}\mathbf{x}||^2 = ||^{(t)}\mathbf{W}||^2 + ||^{(t)}\mathbf{x}||^2\ (1+(1+\alpha)^2)$. Thus, after $M$ mistakes, we get $||^{(M+1)}\mathbf{W}||^2 < (2 + 2\alpha + \alpha^2)\ R^2\ M = (1+(1+\alpha)^2)\ R^2\ M$.

#### 2.1.3 Combining the Lower and Upper Bounds on $||\mathbf{W}||$

By combining the derived lower and upper bounds on $||^{(M+1)}\mathbf{W}||^2$, we obtain the corresponding mistake bounds.

(1) AMPerc: $(\gamma M)^2 \leq ||^{(M+1)}\mathbf{W}||^2 < 2(R^2 + \beta)\ M$. Solving for $M$, we get: $M < 2(R^2+\beta)/\gamma^2 = 2\ (R^2/\gamma^2)\ (1 + \beta/R^2) = 2\ (R/\gamma)^2\ (1 + \beta/R^2)$.

(2)-(3) MMPerc and MMPerc-Abs: $(\gamma M)^2 \leq ||^{(M+1)}\mathbf{W}||^2 < R^2\ (2\ M + \alpha\ M^2 - \alpha\ M)$. Thus, $M < R^2\ (2 - \alpha)\ /\ (\gamma^2 - \alpha\ R^2) = (R^2\ /\ \gamma^2)\ (2 - \alpha)\ /\ (1 - \alpha\ R^2\ /\ \gamma^2)$, where $\gamma^2 - \alpha\ R^2 > 0$.

(4) MMPerc-Asm: $((\gamma - \alpha\ R)\ M)^2 \leq ||^{(M+1)}\mathbf{W}||^2 < (2 - 2\alpha + \alpha^2)\ R^2\ M$. Thus, $M < (2 - 2\alpha + \alpha^2)\ R^2\ /\ (\gamma - \alpha\ R)^2 = (1 + (1 - \alpha)^2)\ R^2\ /\ (\gamma - \alpha\ R)^2$, where $\gamma - \alpha\ R > 0$.

(5) MMPerc-AbsAsm: $((\gamma - \alpha\ R)\ M)^2 \leq ||^{(M+1)}\mathbf{W}||^2 < (2 + 2\alpha + \alpha^2)\ R^2\ M$. Thus, $M < (2 + 2\alpha + \alpha^2)\ R^2\ /\ (\gamma - \alpha\ R)^2 = (1 + (1 + \alpha)^2)\ R^2\ /\ (\gamma - \alpha\ R)^2$, where $\gamma - \alpha\ R > 0$.

**Q.E.D.**

### 2.2 Proof of Theorem 2 (Mistake bounds for large margin Perceptrons in the linearly non-separable case)

The formulation of Theorem 2 is given in section 3.3.1.

**Proof.** We initialize all the Perceptron variants considered with ${}^{(1)}\mathbf{W} = \mathbf{0}$. To derive the upper bound on the number of their mistakes, we analyze both a lower and an upper bound on the Frobenius dot product $\langle \mathbf{U}, {}^{(T+1)}\mathbf{W} \rangle$.

2.2.1 Lower Bound on $\langle \mathbf{U},\mathbf{W} \rangle$

Let us derive a lower bound on $\langle \mathbf{U}, {}^{(T+1)}\mathbf{W} \rangle$. At round $t$, when a mistake occurs on ${}^{(t)}\mathbf{x}$, a particular Perceptron variant updates its weight matrix $\mathbf{W}$ using ${}^{(t)}\mathbf{x}$. This update changes $\langle \mathbf{U}, \mathbf{W} \rangle$ as follows (where $y^*$ is the incorrectly predicted class):

(1)-(3) For the learning rule (2.2)-(2.3) used in AMPerc, MMPerc, and MMPerc-Abs:

$\langle \mathbf{U}, {}^{(t+1)}\mathbf{W} \rangle - \langle \mathbf{U}, {}^{(t)}\mathbf{W} \rangle = \langle \mathbf{U}, {}^{(t)}\Delta\mathbf{W} \rangle = \langle \mathbf{u}_y, {}^{(t)}\mathbf{x} \rangle + \langle \mathbf{u}_{y^*}, -{}^{(t)}\mathbf{x} \rangle =$

$b - b + \langle \mathbf{u}_y, {}^{(t)}\mathbf{x} \rangle - \langle \mathbf{u}_{y^*}, {}^{(t)}\mathbf{x} \rangle = b - [b - (\langle \mathbf{u}_y, {}^{(t)}\mathbf{x} \rangle - \langle \mathbf{u}_{y^*}, {}^{(t)}\mathbf{x} \rangle)]$.

Consider the term inside the square brackets. Since $z \le [z]_+$, we have: $[b - (\langle \mathbf{u}_y, {}^{(t)}\mathbf{x} \rangle - \langle \mathbf{u}_{y^*}, {}^{(t)}\mathbf{x} \rangle)] \le [b - (\langle \mathbf{u}_y, {}^{(t)}\mathbf{x} \rangle - \langle \mathbf{u}_{y^*}, {}^{(t)}\mathbf{x} \rangle)]_+$. Also, since $\max_{i \ne y} \langle \mathbf{u}_i, {}^{(t)}\mathbf{x} \rangle \ge \langle \mathbf{u}_{y^*}, {}^{(t)}\mathbf{x} \rangle$, we get:

$[\, b - (\langle \mathbf{u}_y, {}^{(t)}\mathbf{x} \rangle - \langle \mathbf{u}_{y^*}, {}^{(t)}\mathbf{x} \rangle)]_+ \le [b - (\langle \mathbf{u}_y, \mathbf{x} \rangle - \max_{i \ne y} \langle \mathbf{u}_i, \mathbf{x} \rangle)]_+ = L_b(\mathbf{U},({}^{(t)}\mathbf{x},{}^{(t)}y))$. Therefore, in the case of a mistake $\langle \mathbf{U}, {}^{(t)}\Delta\mathbf{W} \rangle \ge b - L_b(\mathbf{U}, ({}^{(t)}\mathbf{x},{}^{(t)}y))$. In the case of the correct classification (no mistake), ${}^{(t)}\Delta\mathbf{W}=0$, and therefore $\langle \mathbf{U}, {}^{(t)}\Delta\mathbf{W} \rangle = 0$.

Since we initialize the weights as ${}^{(1)}\mathbf{W} = \mathbf{0}$, then $\langle \mathbf{U}, {}^{(T+1)}\mathbf{W} \rangle$ is equal to the sum of $\langle \mathbf{U}, {}^{(t)}\Delta\mathbf{W} \rangle$ at mistakes. Thus, after $M$ mistakes in $T$ rounds, we have: $\langle \mathbf{U}, {}^{(T+1)}\mathbf{W} \rangle = \sum_{m=1,M} \langle \mathbf{U}, {}^{(m)}\Delta\mathbf{W} \rangle \ge b\, M - L_b(\mathbf{U})$, where $L_b(\mathbf{U}) = \sum_{m=1,M} L_b(\mathbf{U}, ({}^{(m)}\mathbf{x}, {}^{(m)}y)) \equiv L_b$ and $m$ indexes the rounds on which the particular Perceptron variant made a mistake.

(4) For the learning rule of MMPerc-Asm (3.5):

$\langle \mathbf{U}, {}^{(t)}\Delta\mathbf{W} \rangle = b - b + \langle \mathbf{u}_y, {}^{(t)}\mathbf{x} \rangle + \langle \mathbf{u}_{y^*}, -{}^{(t)}\mathbf{x} \rangle - \langle \mathbf{u}_y, \alpha\, {}^{(t)}\mathbf{x} \rangle =$

$b - [\, b - (\langle \mathbf{u}_y, {}^{(t)}\mathbf{x} \rangle - \langle \mathbf{u}_{y^*}, {}^{(t)}\mathbf{x} \rangle)] - \langle \mathbf{u}_y, \alpha\, {}^{(t)}\mathbf{x} \rangle$.

Therefore, $\langle \mathbf{U}, {}^{(t)}\Delta\mathbf{W} \rangle \ge b - L_b(\mathbf{U},({}^{(t)}\mathbf{x},{}^{(t)}y)) - \alpha \langle \mathbf{u}_y, {}^{(t)}\mathbf{x} \rangle \ge b - L_b(\mathbf{U},({}^{(t)}\mathbf{x},{}^{(t)}y)) - \alpha\, ||\mathbf{U}||\, R$, taking into account $\langle \mathbf{u}_y,{}^{(t)}\mathbf{x} \rangle \le ||\mathbf{U}||\, R$. Thus, after $M$ mistakes:

$\langle \mathbf{U}, {}^{(T+1)}\mathbf{W} \rangle = \sum_{m=1,M} \langle \mathbf{U}, {}^{(m)}\Delta\mathbf{W} \rangle \ge (b - \alpha\, ||\mathbf{U}||\, R)M - L_b(\mathbf{U})$.

(5) For the learning rule of MMPerc-AbsAsm (3.9):

$\langle \mathbf{U}, {}^{(t)}\Delta\mathbf{W} \rangle \ge b - L_b(\mathbf{U}, (\mathbf{x}, y)) - \alpha \langle \mathbf{u}_y, {}^{(t)}\mathbf{x} \rangle \operatorname{sign} \langle {}^{(t)}\mathbf{w}_y, {}^{(t)}\mathbf{x} \rangle$.

Since $\langle \mathbf{u}_y, {}^{(t)}\mathbf{x} \rangle$ sign $\langle {}^{(t)}\mathbf{w}_y, {}^{(t)}\mathbf{x} \rangle \leq ||\mathbf{U}||\ R$, we obtain: $\langle \mathbf{U}, {}^{(t)}\Delta\mathbf{W} \rangle \geq b - L_b (\mathbf{U},({}^{(t)}\mathbf{x},{}^{(t)}y)) - \alpha\ ||\mathbf{U}||\ R$. Thus, after $M$ mistakes, $\langle \mathbf{U}, {}^{(T+1)}\mathbf{W} \rangle = \sum_{m=1,M} \langle \mathbf{U}, {}^{(m)}\Delta\mathbf{W} \rangle \geq (b - \alpha\ ||\mathbf{U}||\ R)M - L_b\ (\mathbf{U})$.

### 2.2.2 Upper Bound on $\langle \mathbf{U},\mathbf{W} \rangle$

Using the Cauchy-Shwartz inequality: $\langle \mathbf{U}, \mathbf{W} \rangle^2 \leq ||\mathbf{U}||^2\ ||\mathbf{W}||^2$, and applying the upper bounds on $||{}^{(T+1)}\mathbf{W}||^2$ after $M$ mistakes in $T$ rounds (as derived in section 2.1.2 for various large-margin Perceptron variants), we obtain the following bounds:

(1) AMPerc: $\langle \mathbf{U}, {}^{(T+1)}\mathbf{W} \rangle^2 < 2(R^2 + \beta)\ M\ ||\mathbf{U}||^2$.

(2) MMPerc: $\langle \mathbf{U}, {}^{(T+1)}\mathbf{W} \rangle^2 < R^2\ (\alpha\ M^2 + (2 - \alpha)\ M)\ ||\mathbf{U}||^2$.

(3) MMPerc-Abs: $\langle \mathbf{U}, {}^{(T+1)}\mathbf{W} \rangle^2 < R^2\ (\alpha\ M^2 + (2 - \alpha)\ M)\ ||\mathbf{U}||^2$.

(4) MMPerc-Asm: $\langle \mathbf{U}, {}^{(T+1)}\mathbf{W} \rangle^2 < (2 - 2\alpha + \alpha^2)\ R^2\ M\ ||\mathbf{U}||^2$.

(5) MMPerc-AbsAsm: $\langle \mathbf{U}, {}^{(T+1)}\mathbf{W} \rangle^2 < (2 + 2\alpha + \alpha^2)\ R^2\ M\ ||\mathbf{U}||^2$.

### 2.2.3 Combining the Lower and Upper Bounds on $\langle \mathbf{U},\mathbf{W} \rangle$

By combining the derived lower and upper bounds on $\langle \mathbf{U}, {}^{(T+1)}\mathbf{W} \rangle$, we obtain the corresponding mistake bounds for the various large-margin multiclass Perceptron variants.

(1) For AMPerc with the additive margin $\beta$ (and therefore using the $L_{\beta\text{-hinge}}$ loss), and a competitor using the $L_{b\text{-hinge}}$ loss, we get: $(b\ M - L_b)^2 = b^2 M^2 - 2\ bML_b + L_b^2 < 2\ M\ ||\mathbf{U}||^2\ (R^2 + \beta)$ and $b^2 M^2 - 2\ [bL_b + ||\ \mathbf{U}\ ||^2\ (R^2 + \beta)]\ M + L_b^2 < 0$.

Solving the quadratic inequality for $M$, and focusing on the upper bound, we obtain: $M < (1/2)\ (1/b^2)\ \{2\ [b\ L_b + ||\mathbf{U}||^2\ (R^2 + \beta)] + \Delta^{1/2}\}$, where the discriminant $\Delta$ is given by:

$$\Delta = 4\ (b\ L_b + ||\mathbf{U}||^2\ (R^2 + \beta))^2 - 4\ b^2\ L_b^2 = 4\ \{2\ b\ L_b\ ||\mathbf{U}||^2\ (R^2 + \beta)^2 + (||\mathbf{U}||^2\ (R^2 + \beta))^2\}.$$

To obtain a simpler upper bound on $\Delta^{1/2}$, we observe that the summands inside the square root are non-negative. Thus, we apply the subadditivity property of the square root function (i.e., $(a + b)^{1/2} \leq a^{1/2} + b^{1/2}$) to get:

$$\Delta^{1/2} = 2\ [2\ b\ L_b\ ||\mathbf{U}||^2\ (R^2 + \beta) + (||\mathbf{U}||^2\ (R^2 + \beta))^2]^{1/2} \leq$$
$$2\ [2\ b\ L_b\ ||\mathbf{U}||^2\ (R^2 + \beta)]^{1/2} + ||\mathbf{U}||^2\ (R^2 + \beta).$$

Substituting into the earlier expression yields the final mistake bound:

$M < (1/b^2)\ \{b\ L_b + 2\ ||\mathbf{U}||^2\ (R^2 + \beta) + [2\ b\ L_b\ ||\mathbf{U}||^2\ (R^2 + \beta)]^{1/2}\}$.

(2)-(3) For MMPerc and MMPerc-Abs, we begin with the inequality: $(bM - L_b)^2 < R^2\ (\alpha\ M^2 + (2-\alpha)\ M)\ ||\mathbf{U}||^2$, $0 \leq \alpha < 1$, which leads to the following quadratic inequality in $M$: $(b^2 - \alpha R^2\ ||\mathbf{U}||^2)\ M^2 - (2b\ L_b + (2-\alpha)\ R^2\ ||\mathbf{U}||^2)\ M + L_b^2 < 0$.

To find the upper bound, we solve this inequality using the root with the positive sign: $M < \{2(b^2 - \alpha R^2\ ||\mathbf{U}||^2)\}^{-1}\ \{(2b\ L_b + (2-\alpha)\ R^2||\mathbf{U}||^2) + \Delta^{1/2}\}$, where the discriminant $\Delta$ is: $\Delta = [(2bL_b + (2-\alpha)\ R^2\ ||\mathbf{U}||^2)^2 - 4(b^2 - \alpha R^2\ ||\mathbf{U}||^2)\ L_b^2] =$
$4\ b^2L_b^2 + 4\ bL_b\ (2-\alpha)\ R^2\ ||\mathbf{U}||^2 + ((2+\alpha)\ R^2\ ||\mathbf{U}||^2)^2 - 4\ b^2L_b^2 + 4\ \alpha R^2\ ||\mathbf{U}||^2\ L_b^2 =$
$+4\ bL_b\ (2-\alpha)\ R^2\ ||\mathbf{U}||^2 + ((2-\alpha)\ R^2\ ||\mathbf{U}||^2)^2 + 4\ \alpha R^2\ ||\mathbf{U}||^2\ L_b^2$.

Taking the square root of $\Delta$ and applying the subadditivity property of square roots: $\Delta^{1/2} \leq (2-\alpha)\ R^2\ ||\mathbf{U}||^2 + 2\ [bL_b\ (2-\alpha)\ R^2\ ||\mathbf{U}||^2 + \alpha\ L_b^2\ R^2\ ||\mathbf{U}||^2]^{1/2}$.

Finally, we obtain the mistake bound: $M < \{2(b^2 - \alpha R^2\ ||\mathbf{U}||^2)\}^{-1}$
$\{2b\ L_b + 2\ (2-\alpha)\ R^2||\mathbf{U}||^2 + 2\ [bL_b\ (2-\alpha)\ R^2\ ||\mathbf{U}||^2 + \alpha R^2\ ||\mathbf{U}||^2\ L_b^2]^{1/2} =$
$\{b\ L_b + (2-\alpha)\ R^2||\mathbf{U}||^2 + [bL_b\ (2-\alpha)\ R^2\ ||\mathbf{U}||^2 + \alpha R^2\ ||\mathbf{U}||^2\ L_b^2]^{1/2}\ \}\ /\ (b^2 - \alpha R^2\ ||\mathbf{U}||^2)$.

(4) For MMPerc-Asm, we begin with the inequality: $((b - \alpha||\mathbf{U}||R)\ M - L_b)^2 < (2 - 2\alpha + \alpha^2)\ R^2\ M\ ||\mathbf{U}||$, which expands to:
$(b - \alpha||\mathbf{U}||R)^2\ M^2 - \{2\ (b - \alpha||\mathbf{U}||R)\ L_b + (2 - 2\alpha + \alpha^2)\ R^2\ ||\mathbf{U}||^2\}M + L_b^2 < 0$.

Solving the quadratic inequality for $M$, we obtain:
$M < \{2\ (b - \alpha||\mathbf{U}||R)^2\}^{-1}\ \{2\ (b - \alpha||\mathbf{U}||R)\ L_b + (2 - 2\alpha + \alpha^2)\ R^2\ ||\mathbf{U}||^2 + \Delta^{1/2}\}$,
where the discriminant $\Delta$ is given by:
$\Delta = \{-2\ (b - \alpha||\mathbf{U}||R)\ L_b - (2 - 2\alpha + \alpha^2)\ R^2\ ||\mathbf{U}||^2\}^2 - 4\ (b - \alpha||\mathbf{U}||R)\ L_b^2 =$
$4(b - \alpha||\mathbf{U}||R)\ L_b\ (2 - 2\alpha + \alpha^2)\ R^2\ ||\mathbf{U}||^2 + \{(2 - 2\alpha + \alpha^2)\ R^2\ ||\mathbf{U}||^2\}^2$.

Using the subadditivity property of the square root function, the mistake bound becomes: $M < \{(b - \alpha||\mathbf{U}||R)\ L_b + (2 - 2\alpha + \alpha^2)\ R^2\ ||\mathbf{U}||^2 + [(b - \alpha||\mathbf{U}||R)\ L_b\ (2 - 2\alpha + \alpha^2)\ R^2\ ||\mathbf{U}||^2]^{1/2}\}\ /\ (b - \alpha||\mathbf{U}||R)^2$.

(5) For MMPerc-AbsAsm, following the same derivation steps, we get:
$M < \{(b - \alpha||\mathbf{U}||R)\ L_b + (2 + 2\alpha + \alpha^2)\ R^2\ ||\mathbf{U}||^2 + [(b - \alpha||\mathbf{U}||R)\ L_b\ (2 + 2\alpha + \alpha^2)\ R^2\ ||\mathbf{U}||^2]^{1/2}\}\ /\ (b - \alpha||\mathbf{U}||R)^2$.

**Q.E.D.**

# 3 Supplementary Note 3: Extended Experimental Results

## 3.1 Results on Synthetic DataGen Datasets

In addition to the results on the original DataGen data with configurations $C \in \{3,10\}$ and $d \in \{2,10\}$, reported in section 4.3, here we present classification results after increasing the dimensionality of data vectors using random projection (RP). We further examine the effects of binarizing the RP vectors and varying the bias term.

### 3.1.1 Real-Valued Vectors after RP

In this setting, we apply a linear Gaussian RP to expand the original vectors with $d$=2 to $D$=1000 (prior to appending the bias component). Note that RP also transforms the original unipolar input vectors into bipolar hypervectors.

Overall, the trends in classification accuracy remain similar to those without RP, although some differences emerge. For example, in the configuration $C$ =3, $d$ =2, $S$=500, Perc achieves nearly the same accuracy as the best-performing MMPerc (0.9900 vs 0.9927). Also, SVM accuracy improves: 0.9500 vs 0.8620 without RP.

For the case $C$ =3, $d$ =10, $S$=5, SVM outperforms the best-performing MMPerc (0.8253 vs 0.8113). This may be attributed to the one-vs-all strategy of SVM, which can offer better separability in low-sample regimes.

Table 6: Classification accuracy of SVM, Ridge, Perc, MMPerc (for $\alpha$-values indicated in the top row of each table section). Settings: $C \in \{3,10\}$, $d \in \{2,10\}$ transformed to $D$ = 1000 via RP. Corresponding class regions and decision boundaries are shown in Figure 9 and Figure 10.

| *d* | *C* | *S* | SVM | Ridge | Perc | 0.1 | 0.2 | 0.3 | 0.4 | 0.5 | 0.6 | 0.7 | 0.8 | 0.9 | 0.95 |
|---|---|---|---|---|---|---|---|---|---|---|---|---|---|---|---|
| 2 | 3 | 5 | 0.8420 | 0.8647 | 0.8340 | 0.8687 | 0.8767 | 0.8953 | 0.8953 | **0.9447** | 0.9380 | 0.9180 | 0.9300 | 0.9267 | 0.9167 |
| 2 | 3 | 50 | 0.8987 | 0.8493 | 0.874 | 0.8507 | 0.9327 | 0.9193 | 0.9233 | 0.938 | **0.9607** | 0.95 | 0.94 | 0.9293 | 0.9247 |
| 2 | 3 | 500 | 0.9500 | 0.8347 | 0.9900 | 0.9720 | **0.9927** | 0.9887 | 0.9887 | 0.9793 | 0.9713 | 0.9653 | 0.9567 | 0.948 | 0.944 |

| *d* | *C* | *S* | SVM | Ridge | Perc | 0.1 | 0.2 | 0.3 | 0.4 | 0.5 | 0.6 | 0.7 | 0.8 | 0.9 | 0.95 |
|---|---|---|---|---|---|---|---|---|---|---|---|---|---|---|---|
| 2 | 10 | 5 | 0.6052 | 0.5178 | 0.7658 | **0.8292** | 0.7762 | 0.7454 | 0.7016 | 0.685 | 0.6752 | 0.634 | 0.6324 | 0.6044 | 0.5934 |
| 2 | 10 | 50 | 0.5426 | 0.5590 | 0.9406 | **0.9516** | 0.8934 | 0.7954 | 0.7480 | 0.7072 | 0.6764 | 0.6474 | 0.6104 | 0.6200 | 0.6078 |

| *d* | *C* | *S* | SVM | Ridge | Perc | 0.02 | 0.04 | 0.06 | 0.08 | 0.1 | 0.12 | 0.14 | 0.16 | 0.18 | 0.2 |
|---|---|---|---|---|---|---|---|---|---|---|---|---|---|---|---|
| 2 | 10 | 5 | 0.6052 | 0.5178 | 0.7658 | **0.8700** | 0.8576 | 0.7766 | 0.8146 | 0.8292 | 0.7818 | 0.784 | 0.801 | 0.8218 | 0.7762 |
| 2 | 10 | 50 | 0.5426 | 0.5590 | 0.9406 | 0.942 | 0.9214 | 0.9494 | 0.9466 | **0.9516** | 0.9244 | 0.9336 | 0.9164 | 0.901 | 0.8934 |

| *d* | *C* | *S* | SVM | Ridge | Perc | 0.1 | 0.2 | 0.3 | 0.4 | 0.5 | 0.6 | 0.7 | 0.8 | 0.9 | 0.95 |
|---|---|---|---|---|---|---|---|---|---|---|---|---|---|---|---|
| 10 | 3 | 5 | **0.8253** | 0.7393 | 0.5220 | 0.5220 | 0.5220 | 0.7053 | 0.7453 | 0.7787 | 0.7853 | 0.7940 | 0.8000 | 0.8040 | 0.8113 |
| 10 | 3 | 50 | 0.7707 | 0.7933 | 0.6853 | 0.7073 | 0.7040 | 0.7120 | 0.8293 | 0.8527 | **0.8660** | 0.8453 | 0.8220 | 0.7880 | 0.7787 |
| 10 | 3 | 500 | 0.7653 | 0.7987 | 0.8060 | 0.8173 | 0.8160 | 0.7513 | 0.8720 | 0.8540 | 0.8713 | **0.9220** | 0.8520 | 0.8213 | 0.8047 |

| *d* | *C* | *S* | SVM | Ridge | Perc | 0.1 | 0.2 | 0.3 | 0.4 | 0.5 | 0.6 | 0.7 | 0.8 | 0.9 | 0.95 |
|---|---|---|---|---|---|---|---|---|---|---|---|---|---|---|---|
| 10 | 10 | 500 | 0.6266 | 0.5868 | 0.7486 | 0.7492 | 0.7612 | 0.7606 | 0.7434 | 0.7922 | **0.8082** | 0.7916 | 0.7642 | 0.693 | 0.6512 |
| 10 | 10 | 5000 | 0.6828 | 0.5972 | 0.7638 | 0.8086 | 0.7860 | 0.8262 | **0.8388** | 0.8230 | 0.8138 | 0.8052 | 0.7784 | 0.7128 | 0.6684 |

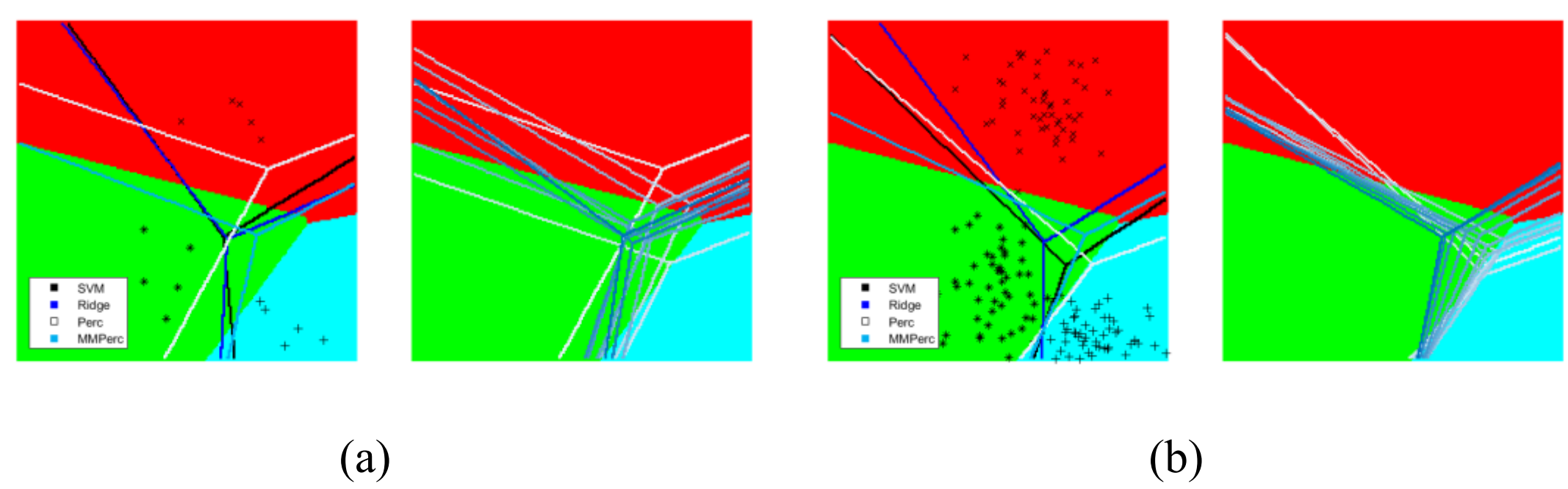

(a) (b)

Figure 9: Class regions and decision boundaries for classification on synthetic data. $C$=3, $d$=2 data transformed to $D$ = 1000 via RP. $\alpha \in [0.95]$. (a) $S$=5; (b) $S$=50.

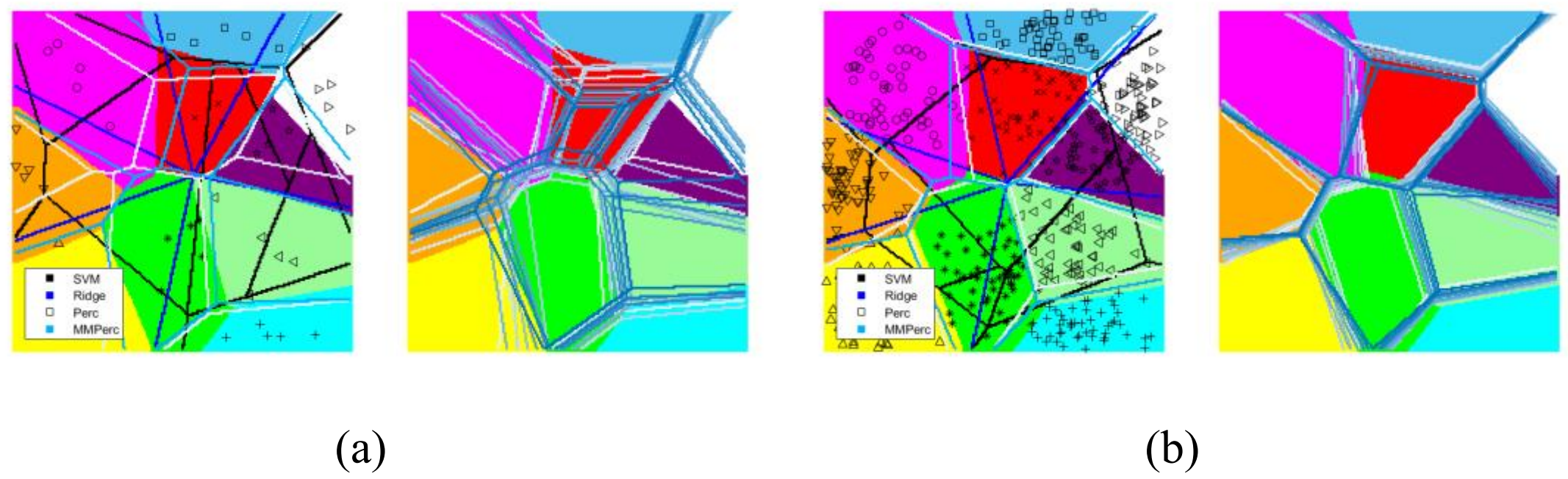

(a) (b)

Figure 10: Class regions and decision boundaries for classification on synthetic data. $C$=10, $d$=2 data transformed to $D$ = 1000 via RP. $\alpha \in [0.2]$. (a) $S$=5; (b) $S$=50.

### 3.1.2 Binarized Data after RP

This experiment demonstrates that, in contrast to standard RP, the binarized RP transformation (RP+bin) alters the structure of the data and disrupts the original class regions generated by DataGen. While RP approximately preserves vector norms, dot products, and Euclidean distances, RP+bin retains only vector directions (angles), while discurding norms and distances. As a consequence, vectors that originally belonged to different DataGen class regions due to their norm differences may be mapped to similar hypervectors after RP+bin. This is visible in the decision boundaries learned by linear classifiers trained on RP+bin data: the boundaries radiate from the origin, mimicking bias-free classification (see Figure 11 and Table 7).

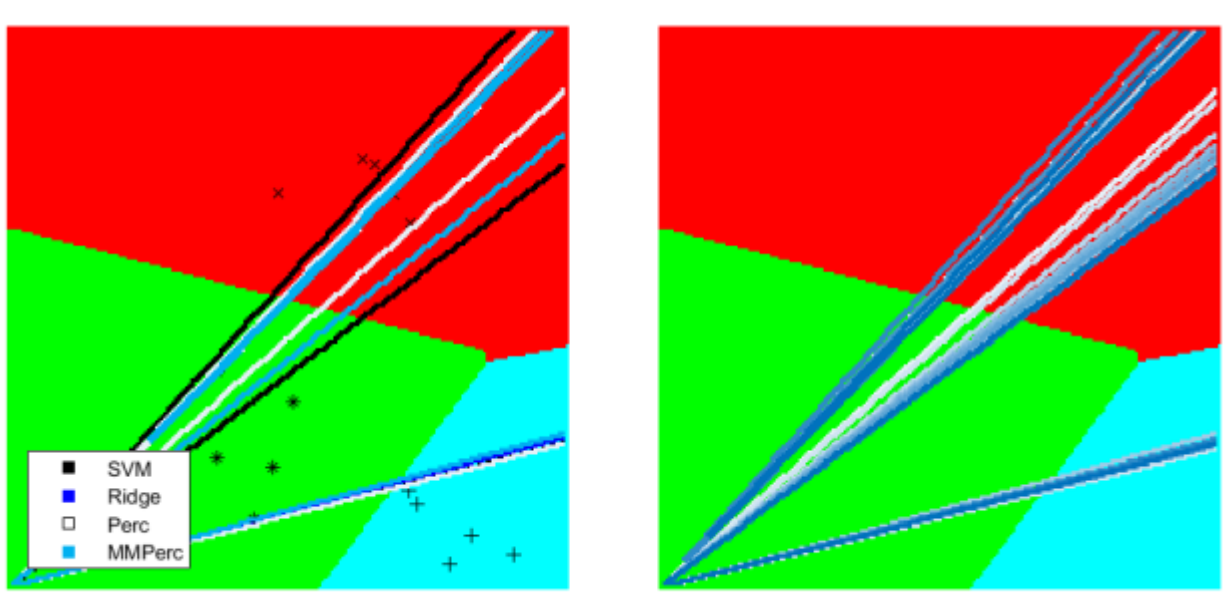

Figure 11: Class regions and decision boundaries for classification on synthetic data. $C$=3, $d$=2, transformed to $D$ = 1000 via RP followed by binarization. $\alpha \in [0.95]$. $S$=5.

Table 7: Classification results for data transformed via RP followed by binarization (RP+bin). $C$=3, $d$=2 transformed to $D$=1000.

| *d* | *C* | *S* | SVM | Ridge | Perc | 0.1 | 0.2 | 0.3 | 0.4 | 0.5 | 0.6 | 0.7 | 0.8 | 0.9 | 0.95 |
|---|---|---|---|---|---|---|---|---|---|---|---|---|---|---|---|
| 2 | 3 | 5 | 0.6253 | **0.6433** | 0.6293 | 0.6213 | **0.6413** | 0.6393 | 0.6340 | 0.6327 | 0.6340 | 0.6307 | 0.6287 | 0.6380 | 0.6387 |

**bias:**

| *d* | *C* | *S* | SVM | Ridge | Perc | 0.1 | 0.2 | 0.3 | 0.4 | 0.5 | 0.6 | 0.7 | 0.8 | 0.9 | 0.95 |
|---|---|---|---|---|---|---|---|---|---|---|---|---|---|---|---|
| 10 | 3 | 5000 | 0.8180 | 0.7327 | 0.7960 | 0.7920 | 0.7833 | 0.7793 | 0.7640 | 0.7980 | 0.8107 | 0.8220 | **0.8367** | 0.8300 | 0.8067 |

**no bias:**

| *d* | *C* | *S* | SVM | Ridge | Perc | 0.1 | 0.2 | 0.3 | 0.4 | 0.5 | 0.6 | 0.7 | 0.8 | 0.9 | 0.95 |
|---|---|---|---|---|---|---|---|---|---|---|---|---|---|---|---|
| 10 | 3 | 5000 | 0.8180 | 0.7327 | 0.7853 | 0.7800 | 0.7393 | 0.7367 | 0.7647 | 0.8107 | 0.8193 | 0.8320 | **0.8340** | 0.8140 | 0.8167 |

### 3.1.3 Bias Term after RP

This experiment investigates the role of the bias term. While omitting the bias may be acceptable for certain vector datasets (such as hypervectors that are implicitly normalized, see section 4.1.1), it proves inadequate for other types of vector data. We also examine how the magnitude of the bias term affects performance of classifiers, in line with the discussion in section 4.1.1.

We revisit the benchmark configuration with $d$=2, $C$=3 after RP into $D$ = 1000. This high-dimensional case yields substantial bias component magnitudes if set to bias = $R$ = $\text{norm}_{\max}$ (the maximum norm of vectors in the dataset). Specifically, we obtain: $\text{bias}_{S=5}$ = 32; $\text{bias}_{S=50}$ = 38; $\text{bias}_{S=500}$ = 41. We also test with bias = 0 and bias = 1. The resulting classification accuracies and decision boundaries are given in Table 8 and Figure 12.

As expected, the case bias = 0 performs poorly, with the decision boundaries radiating from the coordinate origin, clearly illustrating the impact of the missing bias term. Setting bias = 1 improves accuracy relative to the zero-bias case, but remains inferior to bias = $norm_{max}$. Although setting bias = 1 shifts MMPerc decision boundaries in the "proper" direction (away from the origin), the shift proceeds much more slowly compared to bias = $norm_{max}$.

To better illustrate this effect, we increased $nepoch_{max}$ from 1000 to 10000 and recorded the epoch at which training converged (i.e., when training accuracy reached 1). For example, with $S$ = 5 and bias = $norm_{max}$, convergence occurred between epochs 2 and 38 as $\alpha$ increased. In contrast, with bias = 1, convergence was delayed (e.g., epoch 465 for $\alpha$ = 0.0), and did not occur within 10000 epochs for $\alpha \geq 0.4$.

Accross the tested setups, bias = $norm_{max}$ consistently outperformed both bias = 0 and bias = 1. We therefore recommend using bias = $norm_{max}$ as a default choice for new datasets, especially when the data vectors are known to be unnormalized.

Table 8: Classification accuracies for $d$=2, $C$=3, transformed to $D$=1000 by RP. Varied values of the bias component. The decision boundaries are shown in Figure 12.

| $S$ | bias | $n_{ep}$ | Perc | 0.1 | 0.2 | 0.3 | 0.4 | 0.5 | 0.6 | 0.7 | 0.8 | 0.9 | 0.95 |
|---|---|---|---|---|---|---|---|---|---|---|---|---|---|
| 5 | 32 | 100 | 0.8340 | 0.8687 | 0.8767 | 0.8953 | 0.8953 | **0.9447** | 0.9380 | 0.9180 | 0.9300 | 0.9267 | 0.9167 |
| Convergence epoch | | 100 | 2 | 6 | 8 | 8 | 8 | 4 | 10 | 8 | 8 | 26 | 38 |
| 50 | 38 | $10^3$ | 0.8740 | 0.8507 | 0.9327 | 0.9193 | 0.9233 | 0.9380 | **0.9607** | 0.9500 | 0.9400 | 0.9293 | 0.9247 |
| 50 | 38 | $10^4$ | 0.8740 | 0.8507 | 0.9527 | 0.9527 | 0.9607 | 0.9527 | **0.9647** | 0.9527 | 0.9433 | 0.9300 | 0.9260 |
| Convergence epoch | | $10^4$ | 190 | 246 | no cvg | no cvg | no cvg | no cvg | no cvg | no cvg | no cvg | no cvg | no cvg |
| 500 | 41 | $10^3$ | 0.9900 | 0.9720 | **0.9927** | 0.9887 | 0.9887 | 0.9793 | 0.9713 | 0.9653 | 0.9567 | 0.9480 | 0.9440 |
| 500 | 41 | $10^4$ | 0.9900 | 0.9913 | **0.9947** | 0.9840 | 0.9833 | 0.9807 | 0.9667 | 0.9667 | 0.9567 | 0.9480 | 0.9433 |
| Convergence epoch | | $10^4$ | 314 | no cvg | no cvg | no cvg | no cvg | no cvg | no cvg | no cvg | no cvg | no cvg | no cvg |

| $S$ | bias | nep | Perc | 0.1 | 0.2 | 0.3 | 0.4 | 0.5 | 0.6 | 0.7 | 0.8 | 0.9 | 0.95 |
|---|---|---|---|---|---|---|---|---|---|---|---|---|---|
| 5 | 0 | $10^3$ | 0.6540 | 0.6607 | 0.6440 | 0.6547 | 0.6733 | 0.6847 | 0.6847 | 0.6913 | 0.6840 | 0.6733 | 0.6580 |
| 5 | 1 | $10^3$ | 0.7007 | **0.7547** | 0.7407 | 0.7300 | 0.7093 | 0.7307 | 0.7200 | 0.7160 | 0.7100 | 0.6867 | 0.6747 |
| 5 | 1 | $10^4$ | 0.7007 | 0.7320 | 0.7920 | **0.8407** | 0.8213 | 0.7387 | 0.7553 | 0.7187 | 0.7107 | 0.6893 | 0.682 |
| Convergence epoch | | $10^4$ | 465 | 1154 | 2567 | 5808 | no cvg | no cvg | no cvg | no cvg | no cvg | no cvg | no cvg |
| 50 | 0 | $10^3$ | 0.6740 | 0.6173 | 0.6487 | 0.6387 | 0.6293 | 0.6380 | 0.6560 | 0.6747 | 0.6840 | 0.6820 | 0.6713 |
| 50 | 1 | $10^3$ | 0.8387 | **0.9140** | 0.9120 | 0.8867 | 0.7680 | 0.6867 | 0.6827 | 0.6980 | 0.7060 | 0.6953 | 0.6827 |
| 50 | 1 | $10^4$ | 0.8387 | 0.9140 | 0.8940 | **0.9333** | 0.9267 | 0.7480 | 0.6847 | 0.6967 | 0.7080 | 0.6967 | 0.6833 |
| Convergence epoch | | $10^4$ | 750 | 2251 | no cvg | no cvg | no cvg | no cvg | no cvg | no cvg | no cvg | no cvg | no cvg |
| 500 | 0 | $10^3$ | 0.5760 | 0.5633 | 0.5920 | 0.6027 | 0.6180 | 0.6273 | 0.6413 | 0.6540 | 0.6660 | 0.6827 | 0.6800 |
| 500 | 1 | $10^3$ | **0.9627** | 0.9300 | 0.9593 | 0.9713 | 0.9333 | 0.8893 | 0.6767 | 0.6727 | 0.6853 | 0.6920 | 0.6847 |
| 500 | 1 | $10^4$ | 0.9627 | **0.9633** | 0.9540 | 0.9613 | 0.9413 | 0.9007 | 0.6813 | 0.6733 | 0.6847 | 0.6927 | 0.6853 |
| Convergence epoch | | $10^4$ | 918 | no cvg | no cvg | no cvg | no cvg | no cvg | no cvg | no cvg | no cvg | no cvg | no cvg |

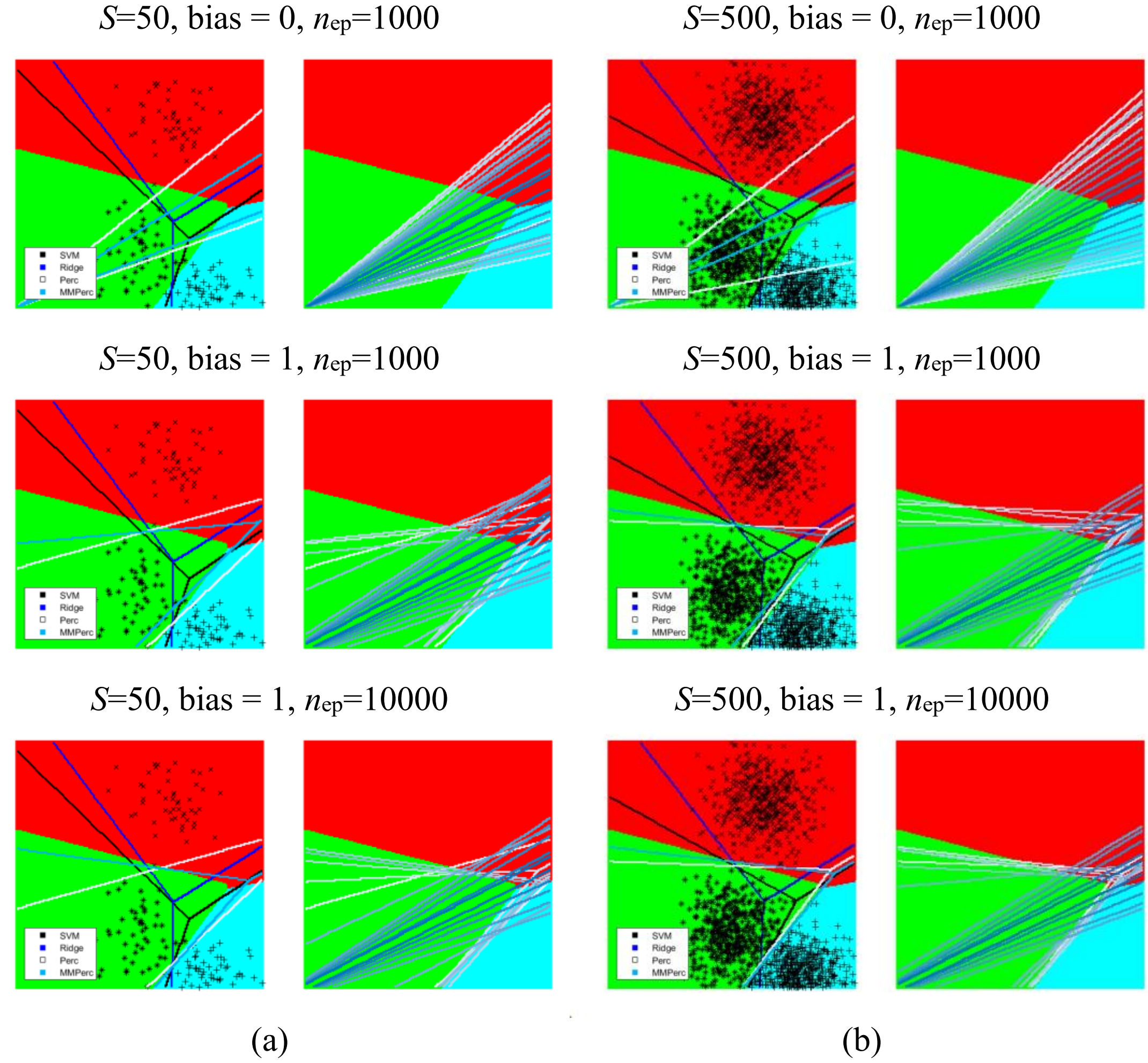

Figure 12: Class regions (red, green, and blue for the three classes) and decision boundaries of the evaluated classifiers. Bias $\in \{0,1\}$, max number of epochs $n_{ep} \in \{1000, 10000\}$. Left: boundaries produced by SVM, Ridge, Perc, and the best MMPerc. Right: MMPerc boundaries for varying $\alpha$-values, with darker colors indicating larger $\alpha$. The corresponding $\alpha$ values and accuracies are reported in Table 8. $C$=3, $d$=2. (a) $S$=50; (b) $S$=500.

## 3.2 Results on Real Datasets

### 3.2.1 UCIHAR

For the original UCIHAR dataset (Figure 13), both MMPerc and Perc with a bias term achieve slightly lower classification accuracy than their bias-free counterparts. Still, MMPerc outperforms Perc. SVM and Ridge achieve slightly higher accuracy than MMPerc.

On the binarized original UCIHAR data (Figure 14), MMPerc surpasses the accuracy of Perc, SVM, and Ridge. The inclusion of a bias term has little impact on performance. While the accuracy obtained with the cross-validated α-value is slightly lower than that achieved with the best-performing α, MMPerc still maintains its advantage over Perc.

The RP-based results (averaged over 10 RP realizations; Figure 15) follow the pattern of the original data. Accuracy is stable across classifiers for $D > 500$, which is expected given the linearity of RP, and consistent with the CTG RP results. However, accuracy declines noticeably for $D \leq 500$. We attribute this to the original data dimensionality being $d$=561, so that RP to lower dimensions distorts the data's similarity structure. Again, MMPerc (α ∈ [0.9]) outperforms Perc. However, Ridge achieves the highest accuracy, followed by SVM.

With RP+bin data (Figure 16), classification accuracy improves as $D$ increases, similar to CTG. The classifier ranks averaged across $D$ values (Figure 17) shows that MMPerc (α ∈ [0.3]) achieves the lowest (best) average rank. The use of the bias term (bias = $norm_{max}$) leads to modest accuracy improvements. Using cross-validated α-values yields slightly lower accuracy than using the best-performing α-values.

Overall, UCIHAR displays a somewhat different pattern compared to other datasets: the bias term does not improve results (however, this difference is small), and for non-binary data, both Ridge and SVM outperform MMPerc.

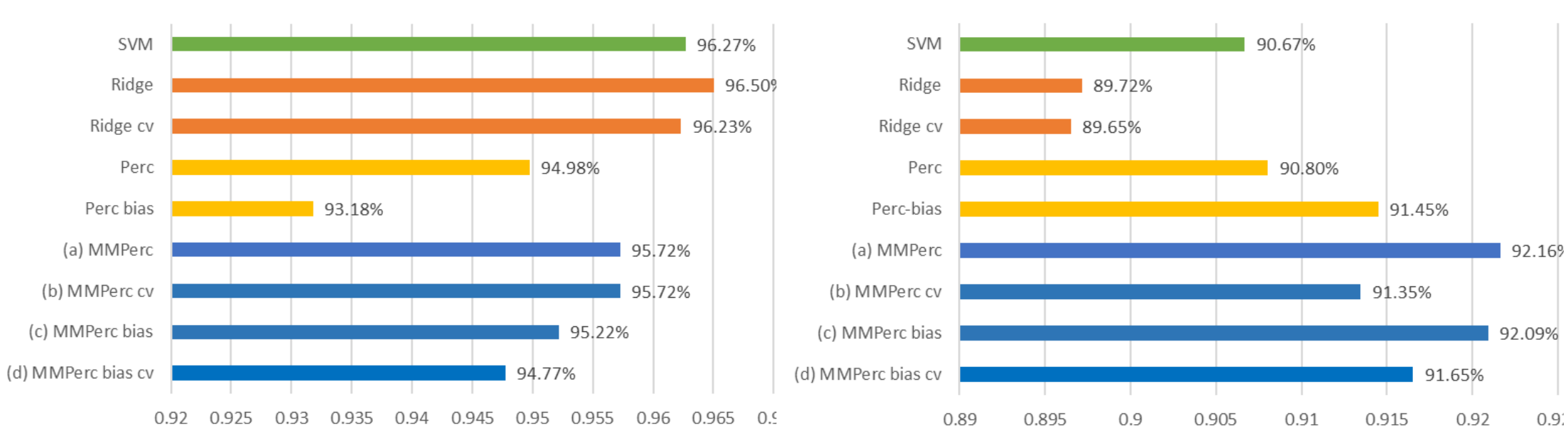

Figure 13: UCIHAR orig. α ∈ [0.3]. (a) α = 0.25; (b) 0.25; (c) 0.17; (d) 0.27

Figure 14: UCIHAR orig+bin. α ∈ [0.9]. (a) α = 0.5; (b) 0.4; (c) 0.6; (d) 0.4

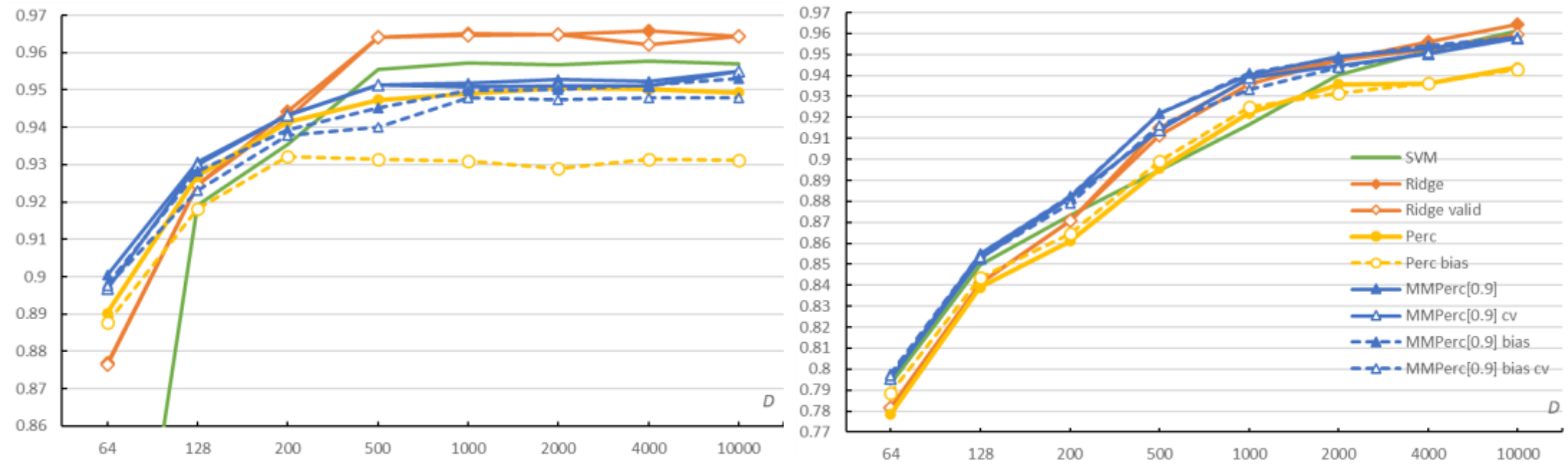

Figure 15: UCIHAR RP. α ∊ [0.9]    Figure 16: UCIHAR RP+bin. α ∊ [0.9]

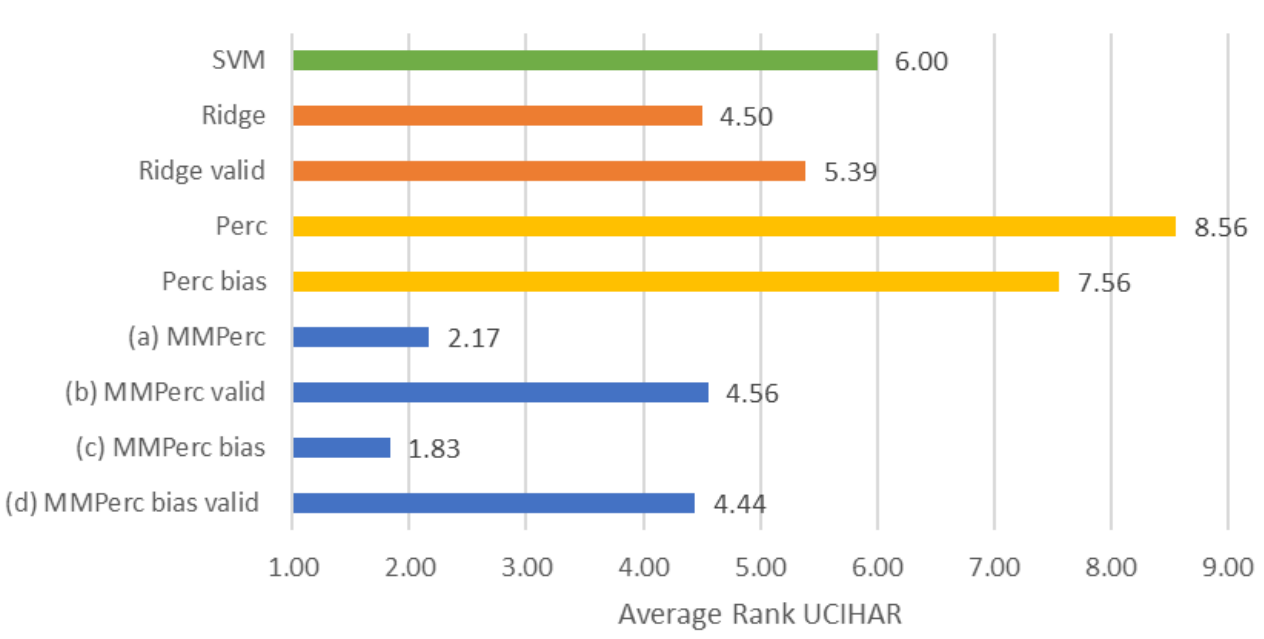

Figure 17: Average rank (across values of $D$) of classifiers operating on the RP+bin hypervectors of UCIHAR. α ∊ [0.9]. A lower average rank corresponds to better relative classification performance

### 3.2.2 ISOLET

For the original ISOLET dataset (Figure 18), both MMPerc and Perc with a bias term perform comparably to their bias-free counterparts. However, MMPerc outperforms Perc, SVM, and Ridge. A similar pattern of classification accuracy is observed on the binarized ISOLET (Figure 19), albeit with lower classification accuracy due to binarization.

The accuracies with RP (averaged over 10 RP realizations; Figure 20) follow the pattern of ISOLET orig. As with the CTG RP and UCIHAR RP cases, classification accuracy remains approximately constant across classifiers when $D > 500$, which is expected given the linearity of RP. For $D \leq 500$, accuracy declines due to the original data dimensionality being $d$=617. Again, MMPerc (α ∊ [0.9]) outperforms Perc, as well as SVM and Ridge. For RP+bin, classification accuracy increases with $D$ (Figure 21), as observed for CTG and UCIHAR. Figure 22 shows that MMPerc ($\alpha \in [0.9]$) achieves the lowest average rank across *D*, outperforming Perc, SV, Ridge.

Across all ISOLET configurations, the inclusion of the bias term (bias = $norm_{max}$) has little effect on accuracy Furthermore, the accuracy obtained using cross-validation remains close to that achieved with the best-performing $\alpha$-values, indicating that cross-validation is effective for ISOLET.

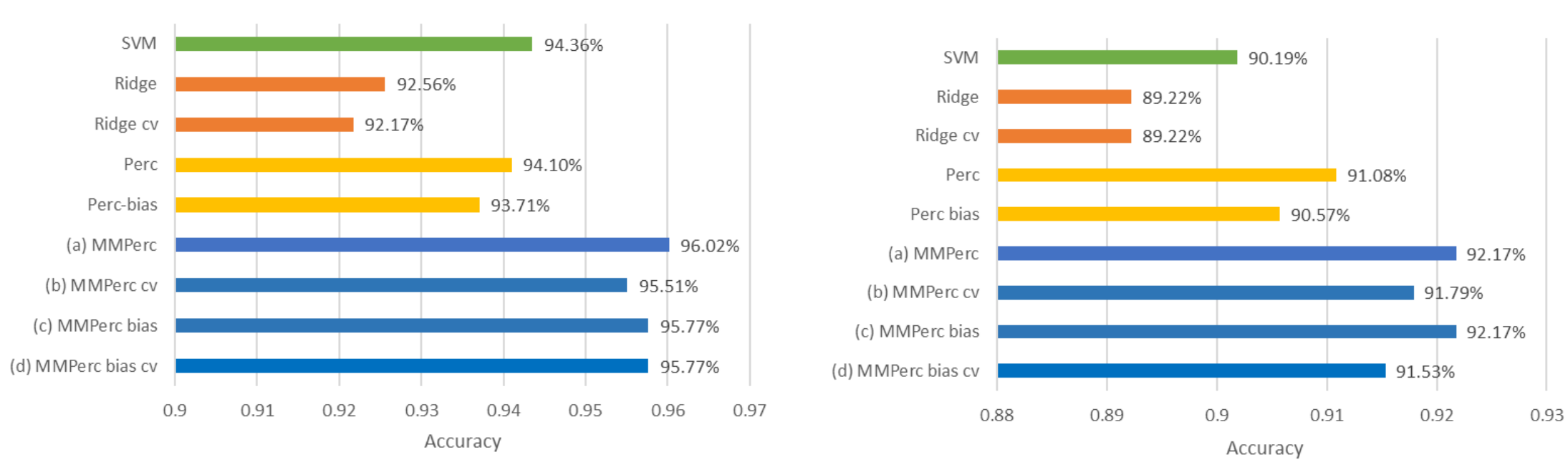

Figure 18: ISOLET orig. $\alpha \in [0.9]$.
(a) $\alpha = 0.4$; (b) 0.5; (c) 0.4; (d) 0.4

Figure 19: ISOLET orig+bin. $\alpha \in [0.9]$.
(a) $\alpha = 0.4$; (b) 0.5; (c) 0.3; (d) 0.5

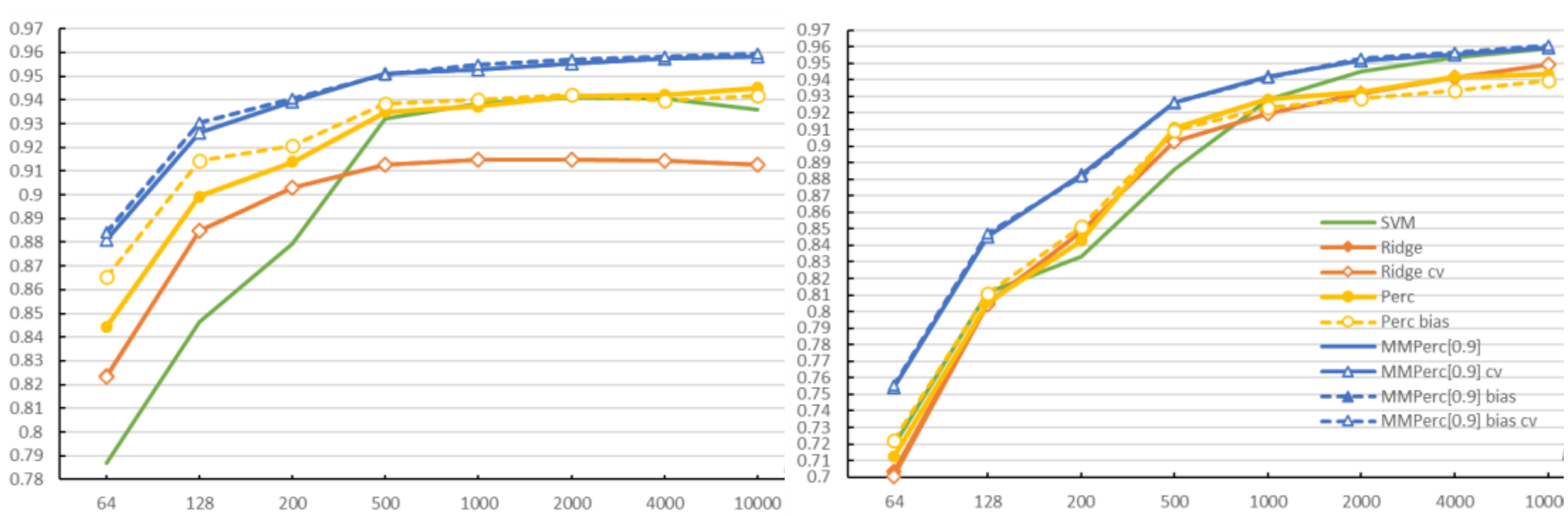

Figure 20: ISOLET RP. $\alpha \in [0.9]$

Figure 21: ISOLET RP bin. $\alpha \in [0.9]$

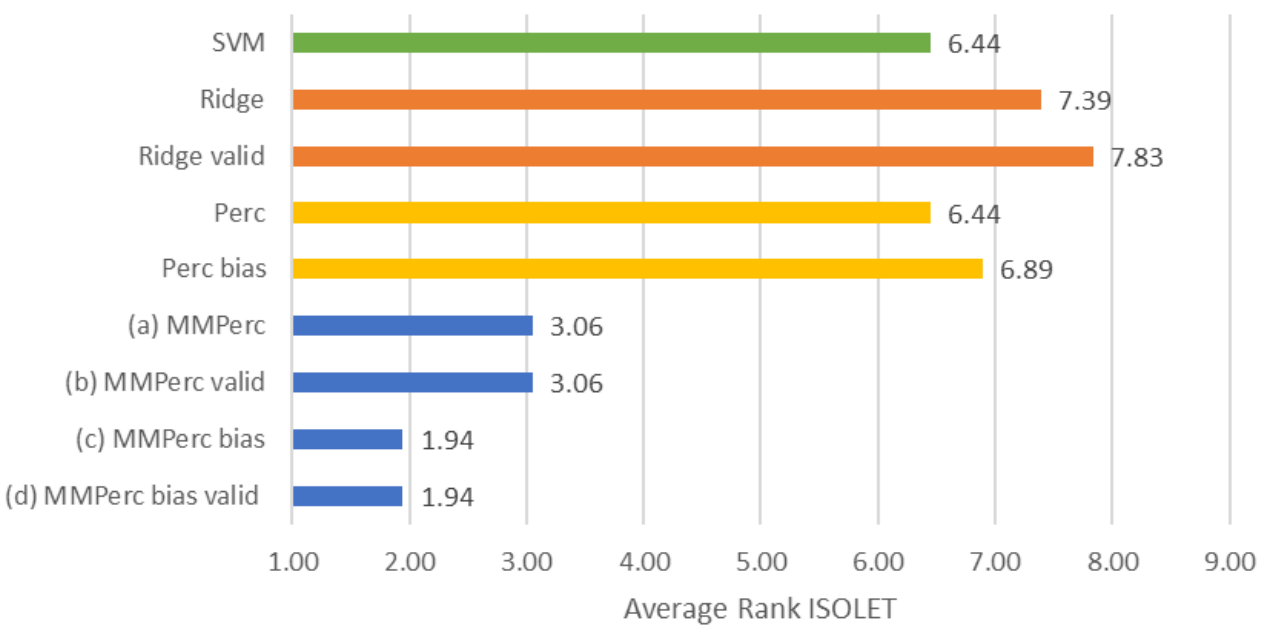

Figure 22: Average (across *D*) rank of the classifiers operating on RP+bin hypervectors of ISOLET. $\alpha \in [0.9]$

### 3.2.3 HAND1

For both the HAND1 orig (Figure 23) and HAND1 orig+bin (Figure 24) datasets, MMPerc achieves classification accuracy that is equal to or very close to that of Perc. In both cases, incorporating a bias term improves accuracy relative to the bias-free variant. On HAND1 orig, both MMPerc and Perc perform on par with SVM and outperform Ridge. On HAND1 orig+bin, MMPerc and Perc surpass both Ridge and SVM. As expected, classification accuracies are substantially lower on the binarized data compared to the original. Notably, the accuracy obtained using cross-validated α-values is very close to that achieved with the best-performing α.

On the HAND1 RP dataset (Figure 25), the accuracy values and trends mirror those observed for HAND1 orig across all tested *D* values. While SVM performs comparably to MMPerc and Perc at lower *D*, its accuracy degrades at higher *D*, which can be attributed to suboptimal SVM hyperparameter settings. Ridge shows the lowest accuracy in this setup. For HAND1 RP+bin (Figure 26), Ridge again underperforms, while the other classifiers achieve similar accuracy. Interestingly, the bias-free variant outperforms the one with a bias term in this case. Accuracy improves with increasing *D*; and RP+bin achieves higher accuracy than both HAND1 orig and all values of *D* tested in the HAND1 RP setting, including the lowest *D* values. Figure 27 shows that MMPerc (α ∈ [0.9]) without bias achieves the lowest average rank across *D*, followed by SVM and MMPerc with bias.

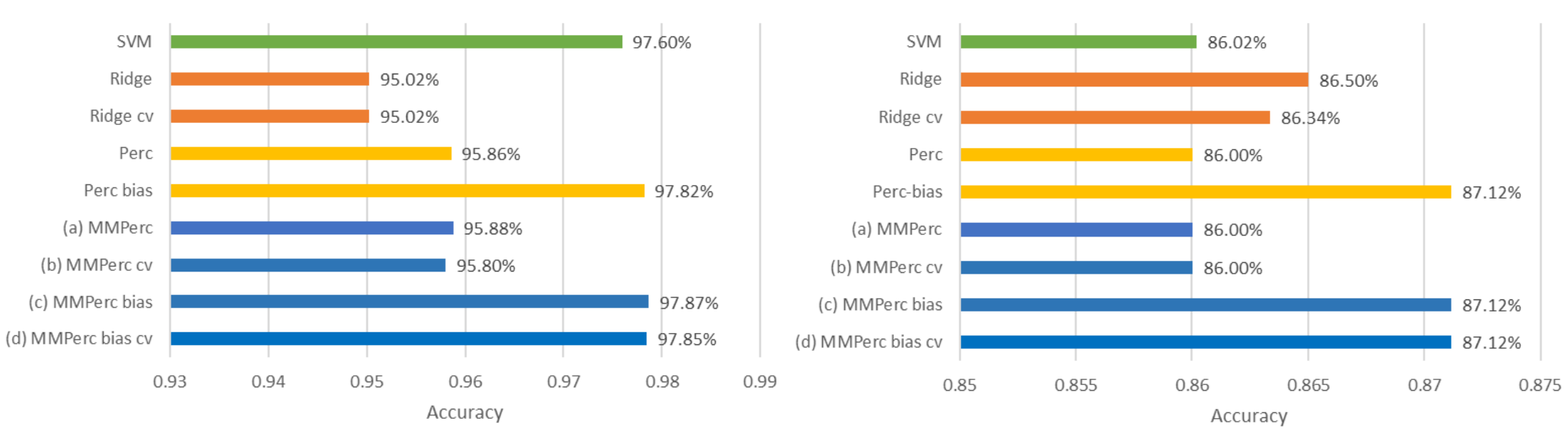

Figure 23: HAND1 orig. α ∈ [0.1]. (a) α = 0.06; (b) 0.10; (c) 0.07; (d) 0.01

Figure 24: HAND1 orig+bin. α ∈ [0.1]. (a) α = 0.0; (b) 0.01; (c) 0.0; (d) 0.0

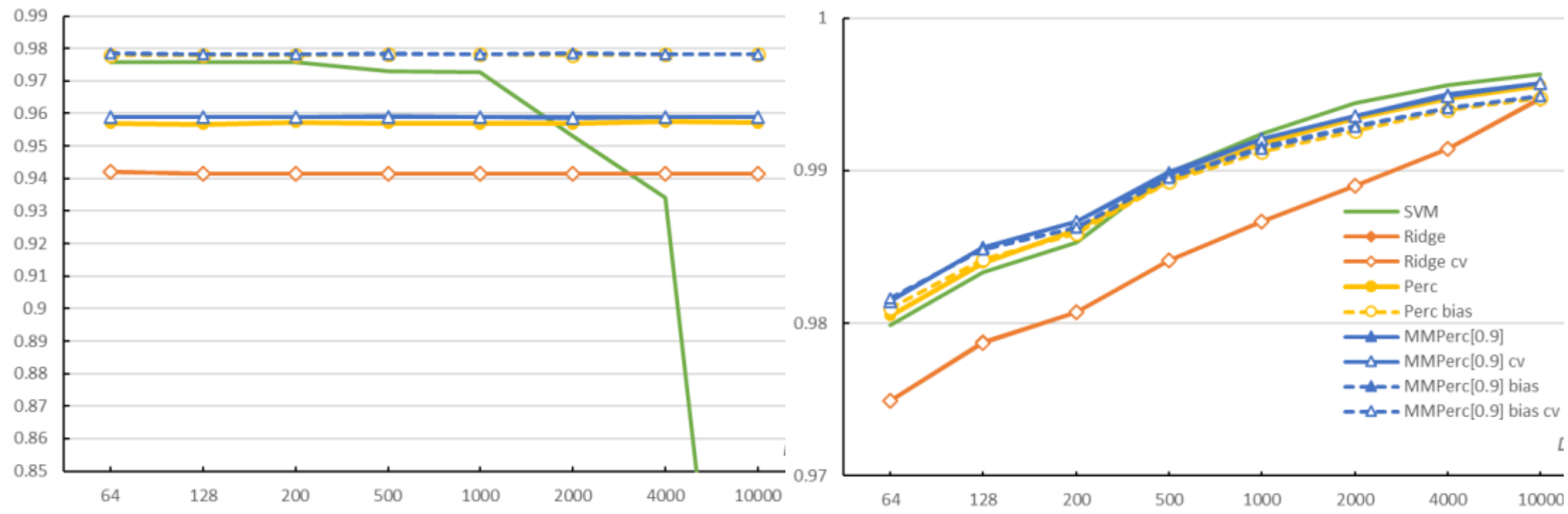

Figure 25: HAND1 RP. α ∈ [0.9]    Figure 26: HAND1 RP+bin. α ∈ [0.9]

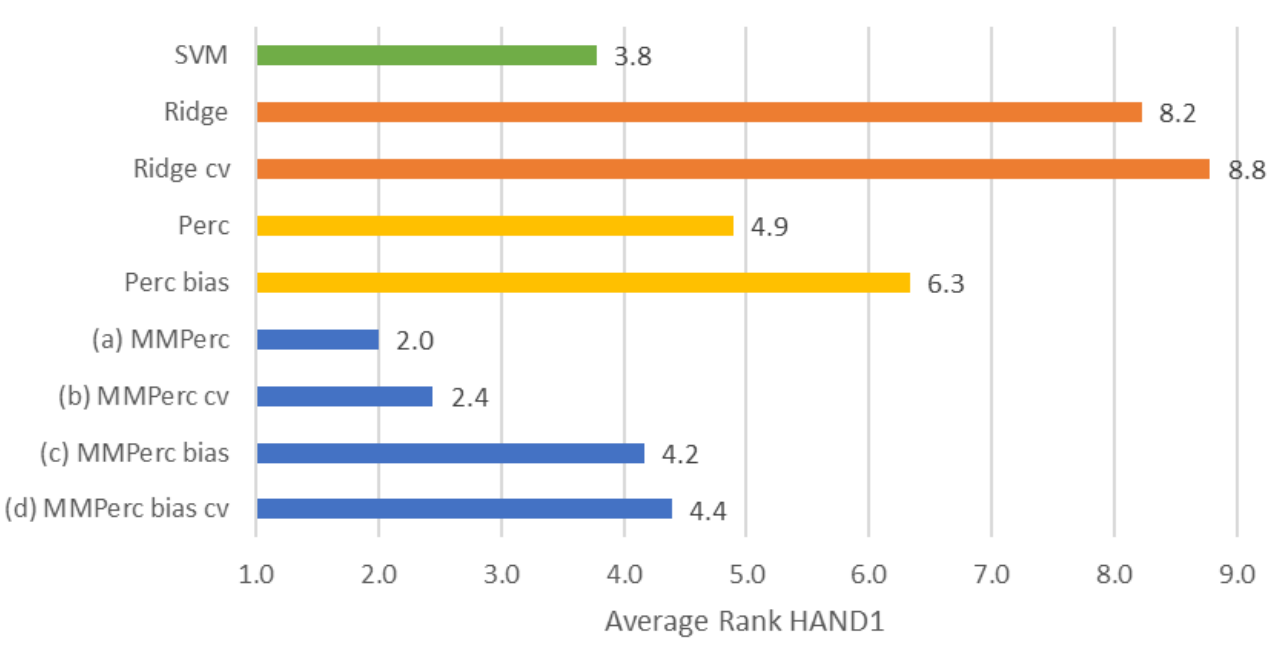

Figure 27: Average (across *D*) rank of the classifiers operating on RP+bin hypervectors of HAND1. α ∈ [0.9]

### 3.2.4 HAND2

For both the HAND2 orig (Figure 28) and HAND2 orig+bin (Figure 29) datasets, MMPerc achieves classification accuracy comparable to that of Perc. The accuracy obtained using cross-validated α-values is close to that achieved with the best-performing α. On HAND2 orig, incorporating the bias term into MMPerc and Perc has little effect. In contrast, on HAND2 orig+bin, the inclusion of a bias term yields a noticeable improvement in classification accuracy.

On HAND2 orig, SVM achieves a slightly lower accuracy than both MMPerc and Perc, but performs noticeably better than Ridge. For HAND2 orig+bin, SVM performs slightly better than Ridge; however, both fall well below MMPerc and Perc.

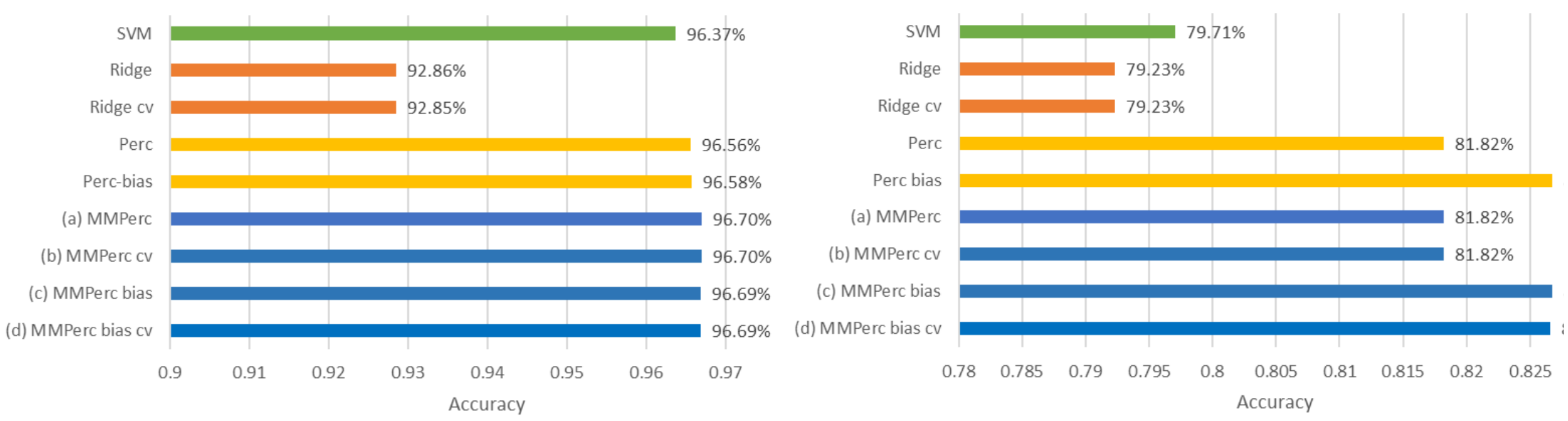

Figure 28: HAND2 orig. α ∈ [0.1].
(a) α = 0.09; (b) 0.09; (c) 0.10; (d) 0.10

Figure 29: HAND2 orig+bin. α ∈ [0.1].
(a) α = 0.0; (b) 0.0; (c) 0.0; (d) 0.09

### 3.2.5 HAND3

For both HAND3 orig (Figure 30) and HAND3 orig+bin (Figure 31) datasets, the accuracy of MMPerc is comparable to that of Perc. The MMPerc accuracy obtained using cross-validated α-values is close to that with the best-performing α (though slightly worse in the case of MMPerc valid). For HAND3 orig, incorporating the bias term into MMPerc and Perc results in substantially improved performance, whereas for HAND3 orig+bin, including the bias term results in approximately the same accuracy as without bias.

For HAND3 orig, SVM achieves slightly lower accuracy than MMPerc and Perc with bias, and slightly outperforms Ridge. In contrast, for HAND3 orig+bin, SVM performs substantially worse than Ridge, while Ridge itself remains somewhat below MMPerc and Perc.

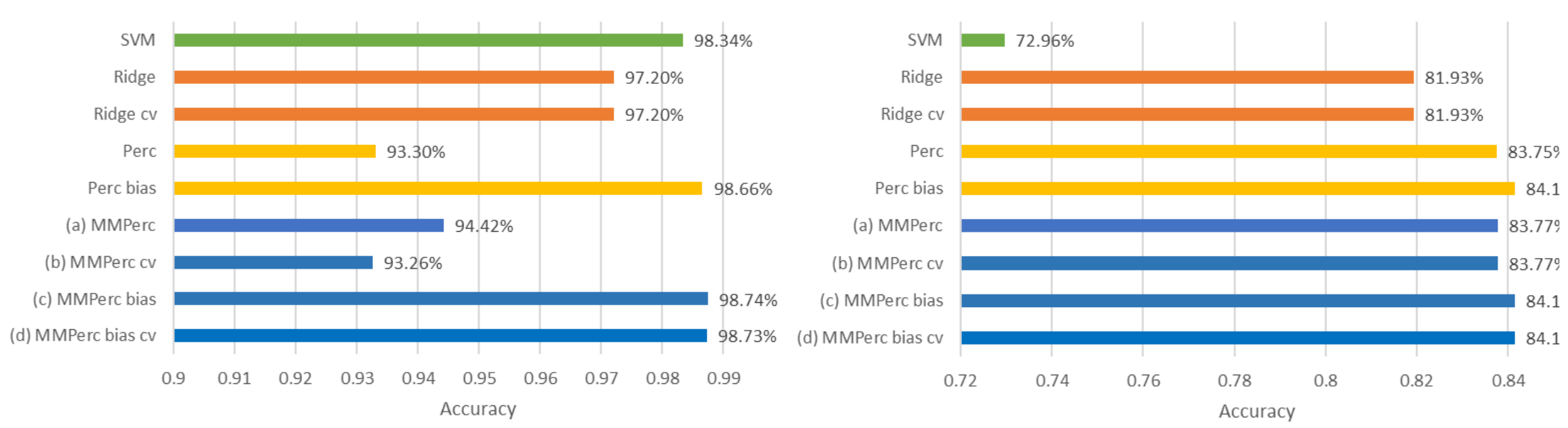

Figure 30: HAND3 orig. α ∈ [0.1].
(a) α = 0.05; (b) 0.03; (c) 0.08; (d) 0.07

Figure 31: HAND3 orig+bin. α ∈ [0.1].
(a) α = 0.01; (b) 0.01; (c) 0.0; (d) 0.09

### 3.2.6 HAND4

For both HAND4 orig (Figure 32) and HAND4 orig+bin (Figure 33) datasets, the accuracy of MMPerc is comparable to that of Perc. The MMPerc accuracy obtained using cross-validated α values is close to that with the best-performing α when using the bias term, and is slightly lower in the case without bias. For both HAND4 orig and for HAND4 orig+bin, incorporating the bias term into MMPerc and Perc leads to substantially improved performance.

For HAND4 orig, SVM achieves slightly lower accuracy than MMPerc and Perc (both with bias), but outperforms Ridge. In contrast, for HAND4 orig+bin, SVM performs worse than Ridge, while Ridge remains slightly below MMPerc and Perc in classification accuracy.

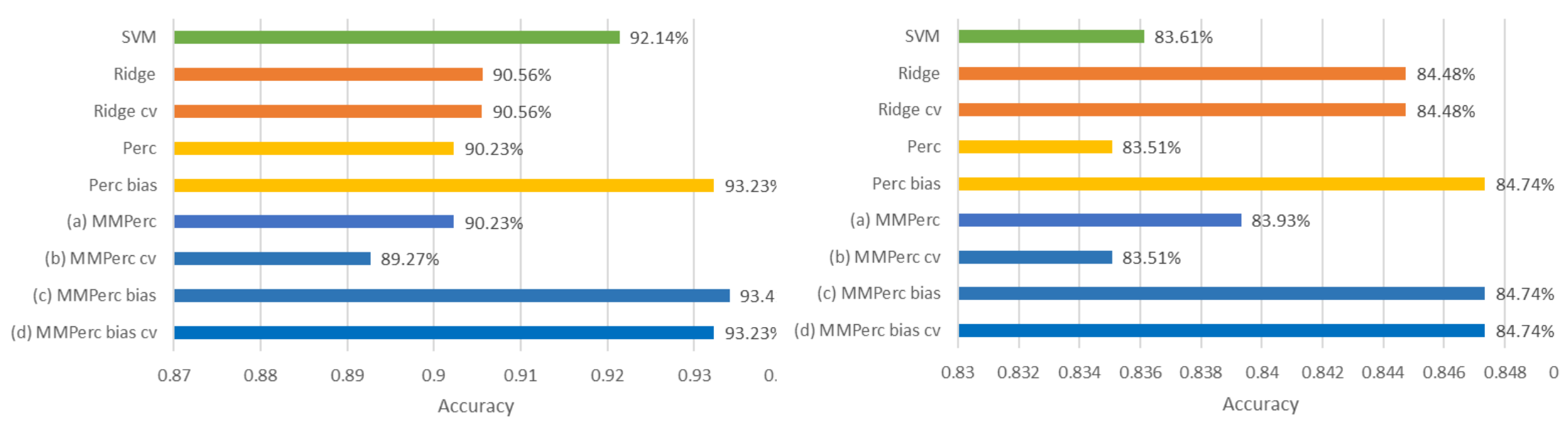

Figure 32: HAND4 orig. α ∈ [0.1]. (a) α = 0.0; (b) 0.06; (c) 0.09; (d) 0.0

Figure 33: HAND4 orig+bin. α ∈ [0.1]. (a) α = 0.01; (b) 0.0; (c) 0.0; (d) 0.0

### 3.2.7 HAND5

For both HAND5 orig (Figure 34) and HAND5 (Figure 35) orig+bin datasets, the accuracy of MMPerc is again very close to that of Perc. The accuracy obtained by MMPerc using cross-validated α-values closely matches that achieved with the best-performing α when a bias term is used, and is slightly lower in the case without bias. In both datasets, incorporating a bias term into MMPerc and Perc substantially improves performance. On HAND5 orig, SVM performs slightly worse than MMPerc and Perc (both with bias), but still outperforms Ridge. In contrast, for HAND5 orig+bin, SVM underperforms compared to Ridge, which performs on par with MMPerc and Perc with bias.

For HAND5 RP (Figure 36), the accuracy values and their pattern mirror those observed for HAND5 orig across all values of *D*. While the SVM accuracy is close to that of MMPerc and Perc with bias at lower $D$, it declines at higher $D$, which can be attributed to suboptimal SVM hyperparemeter settings. MMPerc and Perc without bias perform substantially worse than their biased counterparts, while Ridge accuracy lies between the two.

For HAND5 RP+bin (Figure 37), Ridge shows the lowest performance, while the other classifiers achieve rather close accuracies. Accuracy consistently increases with $D$; notably, RP+bin outperformes both HAND5 orig and HAND5 RP, starting with $D = 200$. Figure 38 shows that MMPerc ($\alpha \in [0.9]$) without bias attains the lowest average rank across $D$, followed by SVM and MMPerc with bias, Perc, and Ridge.

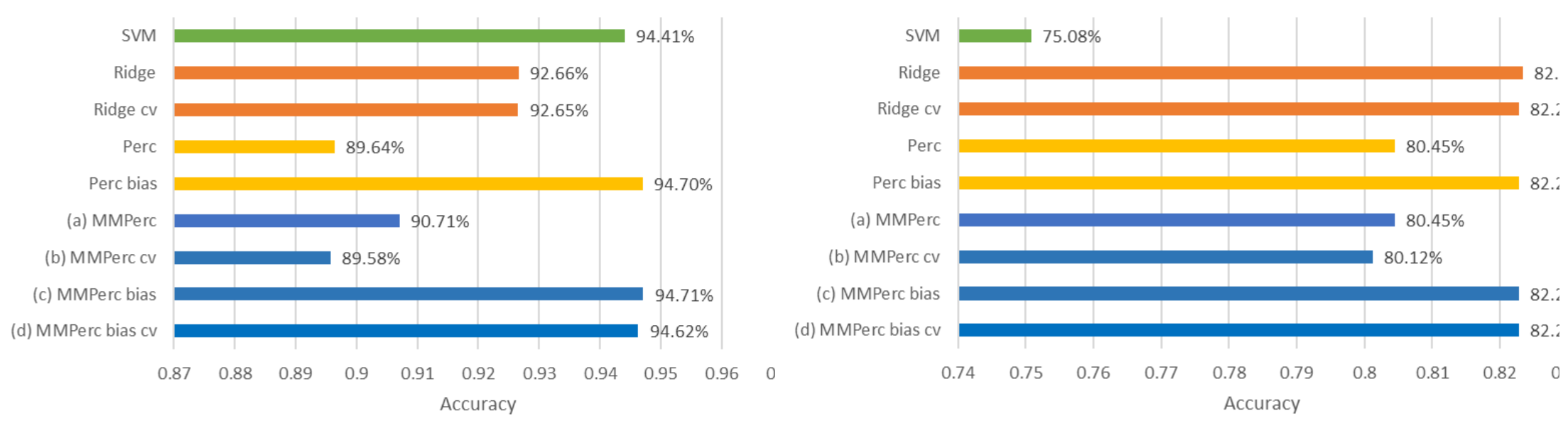

Figure 34: HAND5 orig. $\alpha \in [0.1]$. (a) $\alpha = 0.06$; (b) 0.09; (c) 0.01; (d) 0.1

Figure 35: HAND5 orig+bin. $\alpha \in [0.1]$. (a) $\alpha = 0.0$; (b) 0.01; (c) 0.0; (d) 0.0

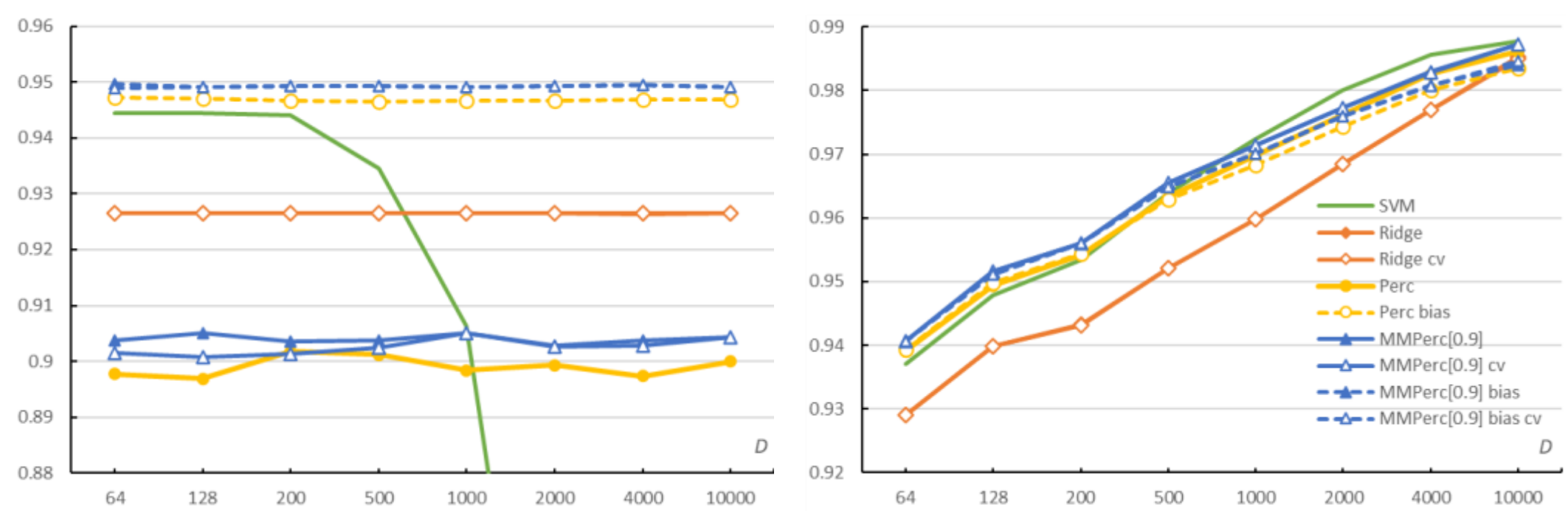

Figure 36: HAND5 RP. $\alpha \in [0.9]$

Figure 37: HAND5 RP+bin. $\alpha \in [0.9]$

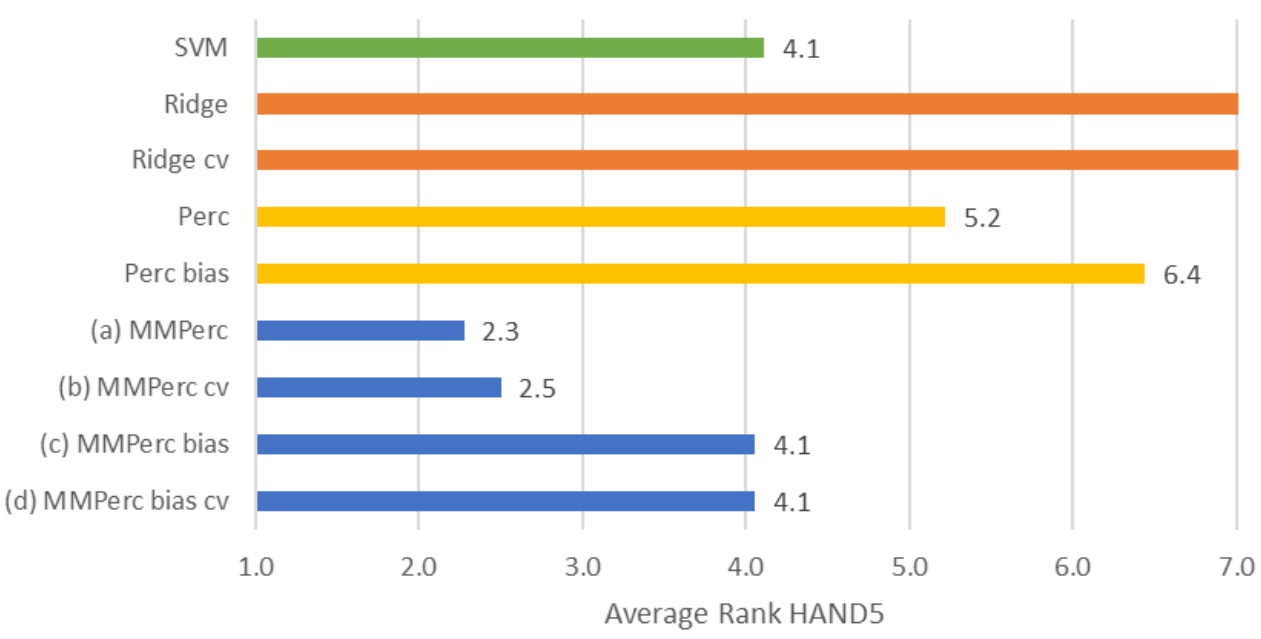

Figure 38: Average (across *D*) rank of the classifiers operating on RP+bin hypervectors of HAND5. $\alpha \in [0.9]$

**Overall**, across the HAND1–HAND5 datasets (both original and original with binarization), MMPerc consistently achieves classification accuracy comparable to that of the standard Perceptron (Perc), generally showing a slight advantage. The inclusion of a bias term (bias = $\text{norm}_{max}$) leads to substantial performance improvements in most cases, particularly for the original datasets, except for HAND2 orig and HAND3 orig+bin, where the results are on par. Cross-validated $\alpha$-values for MMPerc yield accuracies close to those achieved with the best-performing $\alpha$, especially when the bias term is used.

SVM typically performs slightly below MMPerc and Perc with bias on the original datasets but often outperforms Ridge. However, on the binarized original datasets, SVM's performance declines noticeably, falling below that of Ridge (except for HAND2, where Ridge performs slightly worse than SVM), which in turn is generally less accurate than MMPerc and Perc (both with bias).

Notably, for HAND1–HAND5, classification accuracies on the original data are substantially higher than those on the binarized original data. This is due to the very low dimensionality ($d = 4$) of the original data in all five datasets, where binarization severely limits representational capacity and reduces class separability.

For HAND1 and HAND5, MMPerc without bias attains the lowest average rank across *D* (and so is the best), followed by SVM and MMPerc with bias, Perc, and Ridge.